\documentclass{article}
\usepackage{iclr2027_conference,times}
\iclrfinalcopy
\usepackage[T1]{fontenc}
\usepackage[utf8]{inputenc}
\usepackage{amsmath,amssymb,mathtools,amsthm}
\usepackage{booktabs,tabularx,array,multirow}
\usepackage{graphicx,xcolor,microtype}
\usepackage{url}
\usepackage{float}
\usepackage{placeins}
\usepackage[hidelinks]{hyperref}
\newcommand{\E}{\mathbb{E}}

\newcommand{\Corr}{\operatorname{Corr}}

\newcommand{\arm}[1]{\textsf{#1}}

\title{Organizing LLM Inference for Aggregate\\ Predictions from Behavioral Data}
\author{Varun Gupta\\
Dept. of Operations and Information Systems\\
University of Utah\\
\url{varun.gupta@eccles.utah.edu}
\And
Vijay Kamble\\
Dept. of Information and Decision Sciences\\
University of Illinois Chicago\\
\url{kamble@uic.edu}}
\hypersetup{pdftitle={Organizing LLM Inference for Aggregate Predictions from Behavioral Data},pdfauthor={Varun Gupta, Vijay Kamble}}

\usepackage{listings}
\lstdefinestyle{prompt}{basicstyle=\ttfamily\scriptsize,breaklines=true,breakatwhitespace=false,
  columns=fullflexible,keepspaces=true,showstringspaces=false,xleftmargin=6pt,
  breakindent=0pt,postbreak=\mbox{\textcolor{gray}{$\hookrightarrow$}\space},
  aboveskip=4pt,belowskip=4pt,frame=none}
\usepackage{xcolor}
\usepackage{enumitem}

\begin{document}
\lhead{\small Organizing LLM Inference for Aggregate Predictions}
\rhead{\small Gupta and Kamble}
\maketitle
\thispagestyle{plain}
\vspace{-0.5em}
\begin{center}
{\large October 7, 2026}
\end{center}
\vspace{0.3em}

\begin{abstract}
When using a large language model (LLM) to simulate a user population, how should we combine individual behavioral histories to predict the fraction {of users} who take a particular action? We study this question in e-commerce, {predicting observed} product-level conversion rates from browsing sessions without {task-specific model training}. We vary two design choices, each involving a bias--variance tradeoff. Summarizing a product's sessions before inference {reduces cost} but can discard predictive detail, adding bias, whereas {predicting outcomes for individual sessions} and averaging afterward {preserves that detail} but requires many calls to reduce variance. Evaluating products together in one prompt can {help the model infer} what is typical in the setting, reducing bias, but also makes each {prediction} depend on which other products {appear alongside it}, adding noise. We evaluate both choices across five LLMs and two datasets. Our main finding is that {predictions from individual sessions outperform predictions from summaries when products are evaluated separately, but summaries lead when products are evaluated together.} Comparative summaries achieve the highest Spearman correlation with observed rates in 12 of 15 settings, with no alternative significantly better. Comparison thus helps summaries far more than session {predictions}, overcoming the advantage of behavioral detail. We interpret this {difference} through a Bayesian model of the second tradeoff, which suggests that comparison helps more when the inputs in a prompt {reveal more about differences} across products, consistent with {our results for four types of} comparative context. Comparative summaries also {incur} a small fraction of the inference cost of the session methods. {With the best model for each outcome, they} attain 66--89\% of the correlation of the best supervised benchmark trained on thousands of labeled products.
\end{abstract}

\section{Introduction}
\label{sec:introduction}
When a population-level prediction can draw on many individual behavioral observations, how should a large language model (LLM) leverage the evidence? E-commerce offers a natural setting for this question. Firms making product-level pricing, promotion, and inventory decisions often lack timely and relevant outcome data, especially when strategies are new or market conditions change quickly. At the same time, they may observe rich, recent customer behavior, such as browsing sequences, search activity, dwell time, and product consideration, that reflects customer intent and frictions and so could help predict how these decisions might fare if implemented. Conventional supervised models typically need many labeled outcomes from past interventions to exploit such data, and obtaining those labels through live experimentation can be slow, traffic-intensive, or even infeasible for decisions such as pricing.

Recent work on LLM-based human simulation and market research raises a natural alternative: can pretrained models extract predictive signal from behavioral context without extensive application-specific training \citep{aher2023,brand2024}? Doing so creates a basic inference-design problem with a natural bias-variance tradeoff. One approach is to prompt an LLM to predict outcomes for individual sessions and then aggregate those judgments. This preserves fine-grained behavioral detail, but can require many model calls per product and incurs estimation variance from finite sampling. Another approach is to aggregate the behavioral evidence first and prompt the model to predict directly from a product-level summary. This greatly reduces inference cost and can incorporate much more historical evidence, but compression may discard predictive information and introduce bias. A second design choice is whether {to show the model one product at a time or several products in the same prompt. For example, seeing other products' average viewing times may help the model judge whether a particular product's average viewing time is unusually high or low for that store. Such comparisons can reduce errors caused by assumptions that do not fit the store, but they also make each prediction depend on which other products happen to be included. This creates a second bias--variance tradeoff.} This paper studies how evidence representation, comparative context, and inference budget jointly determine aggregate prediction quality.
\begin{figure}[t]
\centering
\includegraphics[width=\linewidth]{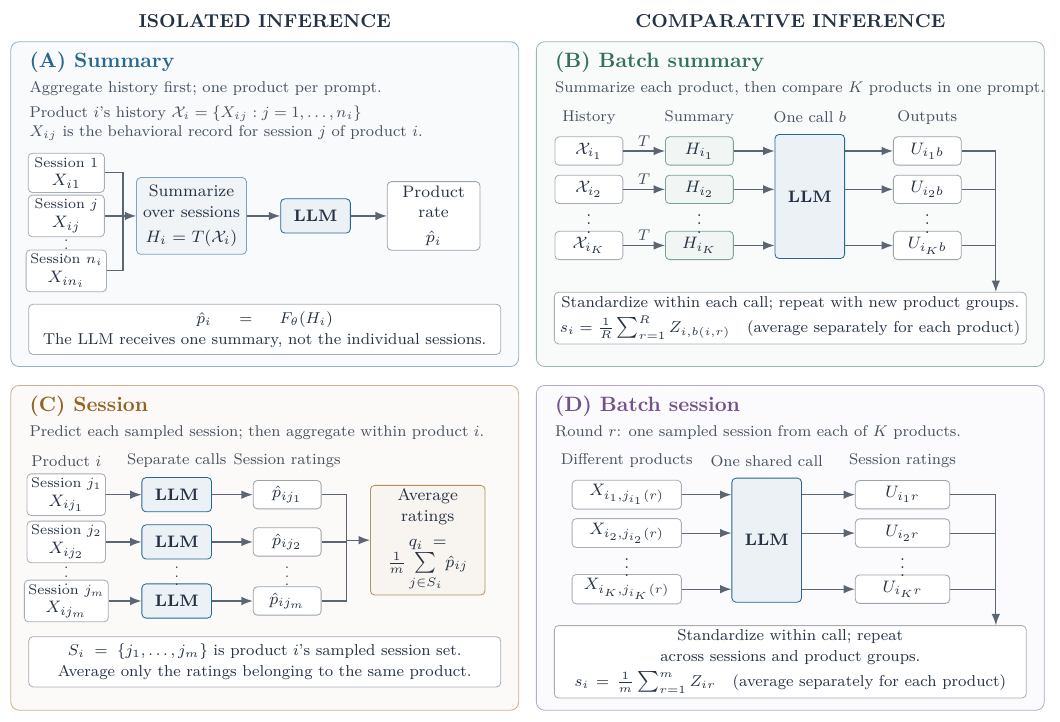}
\caption{The four main methods, by evidence representation (rows) and prompting context (columns). \arm{Summary} and \arm{Batch summary} aggregate sessions before inference; \arm{Session} and \arm{Batch session} aggregate session-level judgments afterwards.}
\label{fig:designs}
\end{figure}

We study these choices using two independent e-commerce datasets: Coveo, with approximately 4.93 million anonymized sessions \citep{tagliabue2021,coveo2021}, and YooChoose, with approximately 9.25 million sessions \citep{Y}. We mask outcome fields and predict product-level add-to-cart and purchase rates on Coveo and purchase rates on YooChoose (where add-to-cart is not available). Inputs contain behavioral signals but exclude product titles, descriptions, images, and semantic embeddings. Prompts contain no labeled examples, and the models receive no task-specific tuning. We develop the procedures using GPT-5.6 on a separate Coveo discovery pool, then evaluate five LLMs on $1{,}000$ held-out Coveo products and replicate the study on YooChoose. The task is retrospective: we predict rates observed under the existing offer. Predicting responses to new prices, promotions, or other interventions, which is our main motivation, raises the same design choices, but their effects can be measured only in experiments. Retrospective prediction provides observed targets for every product, so these choices can be compared directly at scale. Such on-policy prediction under an existing offer is also useful in itself, e.g., for inventory planning when outcomes are sparse, delayed, or unlogged. 

The design space has two axes, shown in Figure~\ref{fig:designs}. First, aggregation can occur before or after model inference. Summary methods compute product statistics from the complete history of a product's recorded sessions, and request one product-level prediction; session methods {make a separate prediction for each of 80 sampled sessions per product and average these predictions} (sample size was set on the discovery pool). Second, prompts can be isolated or comparative. Isolated prompts contain one product, whereas comparative prompts present several products jointly. Comparative prompting is known to improve LLM ranking in information retrieval relative to pointwise scoring \citep{sun2023,qin2024,zhuang2024setwise}, while batch prompting can also reduce inference cost \citep{cheng2023}. Applying comparison to both representations lets us separate the effect of prompting context from the choice of where aggregation occurs.

The zero-shot procedures we consider are not expected to be calibrated to the task, but they may still use the evidence well enough to produce useful predictions that can be calibrated downstream. A single population-level prediction gives one error, which mixes a miscalibrated overall level with how well the procedure uses the evidence, and offers no replication for comparing procedures. The multi-product setting of e-commerce is thus central to our study: we predict rates for many products, each with its own shoppers, and measure the Spearman rank correlation between predicted and observed rates across products. We leave the calibration of predicted levels to future work. The gains from comparison hold under Pearson correlation (Appendix~\ref{app:sensitivity}).

Across five models and three dataset--outcome pairs (Coveo add-to-cart, Coveo purchase, and YooChoose purchase), we find three consistent patterns.

\begin{enumerate}[leftmargin=*,itemsep=0.2pt,parsep=0.1pt,topsep=2pt,partopsep=0.1pt]
\item \textbf{Session-level inference is stronger in isolation.} When products are evaluated independently, session-level prediction outperforms summary prediction in 14 of 15 settings, with one statistically indistinguishable result and no reversals. %
\item \textbf{Comparative context disproportionately benefits summaries.} Evaluating 40 product summaries jointly outperforms evaluating them one at a time in all 15 settings. The same intervention has mixed effects on session inference: comparative session prediction improves in 3 settings, all on YooChoose, worsens in 5, all on Coveo, and is indistinguishable in 7. The gain from comparison is larger for summaries than for sessions in 14 of 15 settings.
\item \textbf{Comparative summaries provide the strongest accuracy--cost tradeoff.} Comparative summaries outperform isolated session inference in 10 of 15 settings and comparative session inference in 11, with no statistically significant reversal against either. Comparative summaries also need only 125 model calls per model--dataset pair, versus 2{,}000 for comparative and 80{,}000 for isolated session inference. 
\end{enumerate}
 
To summarize, our main finding is that comparison helps prediction with product summaries much more than prediction with individual sessions, to the extent that it overcomes the advantage of behavioral detail. To {interpret this result}, we propose a Bayesian model in which the other summaries or sessions in a prompt provide evidence about {what browsing behavior is typical for the store} (Section~\ref{sec:mechanism}). {This information can help the model interpret a product summary or an individual session} and reduce bias, but it also makes the prediction depend on which {other summaries or sessions are shown}, adding noise. The balance depends on the {kind of input}: interpreting a summary statistic may require knowing what is typical across products, which comparison can supply, whereas {an individual session may contain signals of customer intent that have similar meanings across stores}. We use this distinction to interpret the different gains from comparing summaries and sessions.

Finally, we compare the zero-shot LLM procedures with three supervised benchmarks (ridge, random forest and gradient boosting) trained on thousands of labeled products, and with a single summary statistic selected on the discovery pool and used directly to rank products. While the best supervised benchmark significantly outperforms the best LLM method for every outcome, the best LLM method attains 66\% of its Spearman correlation on Coveo add-to-cart, 89\% on Coveo purchase, and 85\% on YooChoose purchase. The single-feature rule outperforms every LLM method on Coveo add-to-cart, whereas the best comparative summaries exceed it on both purchase outcomes.

The broader lesson, within the procedures and budgets we test, is that preserving more behavioral detail is not sufficient for strong aggregate LLM prediction. Under finite inference budgets, comparative judgment based on summaries can outweigh the benefits of finer-grained inputs. Aggregate prediction with LLMs thus involves two bias--variance tradeoffs: where to aggregate the evidence and what context accompanies each judgment. Evidence representation, prompting context, and inference cost should therefore be studied jointly.

\section{Related work}
\label{sec:related}

Our study of aggregate prediction from behavioral records connects work
on the organization of LLM inference with research on predicting human
behavior. In this section, we discuss how our work relates to the most
relevant literature on batch and comparative prompting; inference from
unlabeled context; calibration and aggregation methods; and aggregate
prediction and LLM-based behavioral simulation.

\noindent\textbf{Batched and comparative LLM inference.}
Batch prompting places multiple inputs in a single call and requests a
prediction for each, reducing the repeated cost of shared instructions
and demonstrations.
\citet{cheng2023} find that this can preserve or improve accuracy while
substantially reducing token and time costs. \citet{lin2024batchprompt}
show that batching can also degrade accuracy and introduce sensitivity to
input order. Their method combines predictions across input permutations
by majority vote, recovering accuracy while retaining inference savings.
We similarly combine predictions across repeated calls, using random
groupings of products.

Comparative prompting also underlies LLM methods for ranking documents
by relevance to a query. Pairwise ranking compares two documents
\citep{qin2024}, listwise ranking orders a set of candidates
\citep{sun2023}, and setwise ranking selects among several documents
within a sorting procedure to reduce inference cost
\citep{zhuang2024setwise}. \citet{long2025precise} study zero-shot document
ranking, where an LLM assigns each candidate a numerical relevance score
for a given query. They construct a common reference by summarizing
candidate documents and compare each document against it, improving
score comparability and ranking accuracy. We apply comparative
prompting to both product summaries and individual sessions, eliciting
numerical conversion judgments that are aggregated into product scores.
Our experiments examine how the benefit of comparison depends on
representation, with ablations separating joint presentation from
standardization, repeated groupings, and the amount of behavioral evidence.

\noindent\textbf{Unlabeled context and in-context inference.}
Unlabeled observations can inform a prediction by providing information
about the environment in which it is made. \citet{gupta2024context}
study image classification across environments, such as handwritten
characters from different writers or medical images from different
hospitals. They train predictors to classify an image using other
unlabeled images from the same environment as context, improving both
average accuracy and accuracy in the worst-performing test environment.
\citet{gadetsky2025unsupervised} instead adapt
predictions through model-generated labels: their in-context procedure
iteratively refines these labels and uses them as demonstrations,
improving accuracy without observed task labels. Evidence on the role of
input information also comes from \citet{min2022rethinking}, who find that
randomly replacing demonstration labels often has little effect on
accuracy in the classification tasks they study, while the input
distribution and prompt format remain important. Our procedures examine
the use of unlabeled behavioral context by pretrained LLMs, with no
task-specific training or model-generated demonstrations.

\citet{xie2022explanation} provide a Bayesian account in which prompt
examples update beliefs about the process generating the data. For a
model of pretraining based on a mixture of hidden Markov models, they
derive conditions under which an ideal predictor approaches optimal
classification performance as the number of demonstrations grows, and
support this account with experiments on synthetic data. We propose a
similar Bayesian model to interpret how other products in a prompt might
affect a prediction. Our findings are consistent with this model:
comparison benefits product summaries more than individual sessions,
while batching sessions from the same product reduces ranking accuracy
on average.

Context also affects the interpretation of behavioral prediction tasks.
\citet{gui2025challenge} show that disclosing the price-randomization
procedure improves LLM predictions of responses in a demand experiment.
The experimental description clarifies how price should enter the
prediction. This connects to our use of context to inform the
interpretation of a focal product's behavioral evidence: their context
specifies the experimental design, while ours consists of observations
from other products in the store.

\noindent\textbf{Calibration and aggregation of LLM predictions.}
For classification with prompted language models, calibration methods
adjust class scores to reduce sensitivity to the prompt.
\citet{zhao2021} first estimate the model's baseline preference for each
class by prompting it with a content-free test input such as \texttt{N/A}.
\citet{zhou2024} instead estimate these preferences by averaging class
scores over unlabeled inputs. Both then use these estimates to adjust
class scores, improving classification accuracy and robustness to prompt
choices. Our standardization similarly addresses variation in outputs
across prompts, but operates on numerical conversion judgments: it centers
and scales predictions within each call before averaging across calls.
The resulting scores are evaluated by their ability to rank products.

A separate question is whether predictions are consistent across
levels of aggregation.
\citet{wolf2026partition} compare
a probability elicited for a population with the weighted average of
probabilities elicited for its subgroups. They document discrepancies
between these estimates and find that subgroup aggregation can improve
agreement with human data. Our study examines a related aggregation choice
using observed behavioral records: summarizing sessions
before inference or aggregating session-level predictions afterward.
We compare the accuracy and cost of these alternatives under isolated
and comparative prompting.

\noindent\textbf{Aggregate prediction and behavioral simulation.}
Predicting an outcome rate connects our task to quantification, which
estimates class proportions rather than individual labels
\citep{forman2008quantifying}. HistNetQ predicts these proportions using
learned histograms of collections, with collection-level proportions as
training labels \citep{perezmon2025histnetq}. Distribution regression,
Deep Sets, and Set Transformer more generally learn to predict a response
from a collection of observations
\citep{szabo2016distribution,zaheer2017deepsets,lee2019settransformer}.
Our question is how to represent and present such collections to a
pretrained LLM. TabLLM represents individual table rows as text for classification
\citep{hegselmann2023}. Our summary inputs use a similar representation,
with each row describing the browsing history of one product.

LLM-based behavioral simulation studies both individual responses and
aggregate patterns. \citet{aher2023} simulate participants in experiments
on bargaining and obedience, reproducing established human findings.
Their wisdom-of-crowds experiment also reveals that simulated participants
can be unrealistically accurate. \citet{brand2024} elicit product choices
under varying prices and attributes to estimate demand and willingness
to pay. Shopping benchmarks evaluate predictions against observed
individual behavior: \citet{lu2026} study customers' next actions from
shopping histories, while OPeRA evaluates predictions of a shopper's next
action and stated reason, using the shopper's profile and recorded
browsing history \citep{wang2026}. Our evaluation instead uses product rankings
by aggregate conversion rates, with browsing histories providing the
evidence for prediction.

Other approaches model aggregate responses to changes in prices or
storefronts. \citet{huang2026virtual} combine purchase predictions for
shopper personas using population weights fitted to observed demand.
SimGym uses simulated shoppers interacting with alternative storefronts
and validates their responses against real A/B tests
\citep{li2026simgym}. Our experiments assess product conversion rankings
under the recorded prices and offers.

\section{Data and evaluation}
\label{sec:design}

We evaluate the different inference methods on 1,000 products from each of two public e-commerce datasets. {We construct the model inputs using the information available in each dataset, so the input features differ between the two datasets.}

\noindent{\bf Coveo.}
We use the browsing logs released for the Coveo Data Challenge at the SIGIR 2021 eCom workshop \citep{tagliabue2021,coveo2021}. We select 1,000 products with at least 100 observed sessions. {We group eligible products by their number of sessions and price bucket, then sample products from these groups. The selected products do not overlap with the 280-product discovery pool used to develop and fix the inference procedures} (Section~\ref{sec:architectures}). We predict both add-to-cart and purchase.

\noindent{\bf YooChoose.}
We use the browsing and purchase logs released for the RecSys Challenge 2015 \citep{Y}, which record purchases but no add-to-cart events. The logs record a price only when a product is bought, so we take prices from purchases before August 1, 2014, and behavioral records and target rates from the window afterward; otherwise, prices would reveal which products were purchased. We select 1,000 products with at least 300 sessions in the later window and a priced purchase in the earlier one. We predict the purchase rate. 

In both datasets, we {remove the outcome labels from the inputs: the binary indicators of whether the product being evaluated was added to a cart or purchased during the session. In Coveo, some retained input features, including session length and viewing time, can include time after the outcome event of interest. In YooChoose, we truncate each session at its first purchase, if any. We use these records to reconstruct the conversion rates observed during the same period} (Appendix~\ref{app:protocol}).

\noindent{\bf Models and coverage.}
We use GPT-5.6, Qwen3.8, Gemini3.8 Flash, GLM5.3 and Claude Sonnet5, at temperature zero and low reasoning effort. We evaluate each model on the full pool of 1,000 products with every method (Table~\ref{tab:full-coverage}; Appendix~\ref{app:protocol}).

\noindent{\bf Ranking and uncertainty.}
We measure the association between predicted and observed product ranks by Spearman's rank correlation $\rho$, with tied values assigned their average rank. Ties are common, since 30\% of Coveo purchase rates are exactly zero, but breaking them by sort order instead changes little (Appendix~\ref{app:sensitivity}). {For each dataset (Coveo and YooChoose), we create 2,000 bootstrap resamples of the product pool by sampling products with replacement. For each resample, we compute rank correlations using the saved predictions for every method, model, and outcome, and calculate differences between these correlations. Using the same resampled products makes these differences paired. We use the resulting distributions to estimate 95\% percentile confidence intervals for the correlations and their differences. Because we reuse the saved predictions, these intervals do not capture variability from rerunning model calls, sampling different sessions, or changing prompt groupings. We treat products as independent observations and do not adjust the intervals for the number of comparisons made (Appendix~\ref{app:protocol}). Intervals for averages across models describe the mean performance of the five models evaluated here.}

\section{Organizing product-level inference}
\label{sec:architectures}
We define the four methods of Figure~\ref{fig:designs}, labeled A (\arm{Summary}), B (\arm{Batch summary}), C (\arm{Session}) and D (\arm{Batch session}) in figures and tables.

\noindent{\bf The prediction target.}
Let $i$ index products and $j$ index the browsing sessions in which product $i$ was viewed, and write $\mathcal X_i=\{X_{ij}\}_{j=1}^{n_i}$ for the product's full history, its $n_i$ recorded sessions. The record $X_{ij}$ describes one such session, for example how long the shopper viewed the product, what else they considered, and whether they arrived by search. Each record has a binary label $Y_{ij}$ indicating whether the shopper added product $i$ to the cart (or, for the purchase outcome, bought it) in that session; the label is never shown to the models. We treat the sessions that view product $i$ as draws from a product-specific population. The target is the product's conversion rate $p_i=\E_j[Y_{ij}]$, where the expectation is over a session $j$ drawn from that population, and we evaluate predictions against its sample estimate, the observed fraction $\bar Y_i=n_i^{-1}\sum_{j=1}^{n_i} Y_{ij}$. We thus evaluate ranking by observed fractions in the recorded window, not by population rates or later outcomes. Every prompt asks for both an add-to-cart (where applicable) and a purchase rate, and we evaluate each against the outcomes the dataset records. %

\noindent{\bf Averaging before the model.}
For \arm{Summary} we reduce a product's sessions to a fixed vector of statistics
$H_i=T(\mathcal X_i)$: means, medians, quartiles and proportions of viewing time and search activity, together with the viewing times, prices and counts of the products considered alongside it. The summary omits the product's traffic counts (its number of sessions and, on Coveo, of search sessions) for two reasons: the session methods do not see them, and in the matched variants that summarize the same 80 sessions the session methods use (Section~\ref{sec:isolated}), they are fixed by construction, so including them would make the gain from the full history partly reflect traffic volume. Our main findings hold even when we include them (Appendix~\ref{app:counts}). Writing $F_\theta$ for the model asked to produce a product rate, the prediction is $\hat p_i=F_\theta(H_i)$.

\noindent{\bf Averaging after the model.}
For \arm{Session} we judge each of $m=80$ sampled sessions of one product in its own call, ask for the probability that the session converts, and average the $m$ probabilities:
\begin{equation}
 q_i=\frac{1}{m}\sum_{j\in S_i}\hat p_{ij},\qquad \hat p_{ij}=f_\theta(X_{ij}),
 \label{eq:session}
\end{equation}
where $f_\theta$ is the model asked to judge a single session and $S_i=\{j_1,\dots,j_m\}$ is the sampled set. %

\noindent{\bf Putting several products in one prompt.}
In \arm{Batch summary}, we present $K=40$ product summaries in one prompt and ask for all $K$ rates, repeating over $R=5$ random groupings of the 1,000 evaluation products. In \arm{Batch session}, we present one sampled session from each of $K$ products and ask for all $K$ probabilities, repeating over $m=80$ rounds so that every product contributes each of its sampled sessions once.

\noindent{\bf Standardization within calls.}
The outputs of one call can share a level and scale specific to that call. We hence standardize within each call before combining calls, which removes such call-level shifts. Let $\mu_b$ and $\sigma_b$ be the mean and standard deviation of the outputs of call $b$. The score $s_i$ averages
\begin{equation}
 Z_{ib}=\frac{U_{ib}-\mu_b}{\sigma_b},\qquad
 s_i=\frac1R\sum_{r=1}^{R} Z_{i,b(i,r)}
 \label{eq:z}
\end{equation}
over the calls containing $i$, where $b(i,r)$ is the call holding $i$ in grouping $r$ and $\sigma_b$ is replaced by one when it falls below $10^{-12}$; \arm{Batch session} averages the same way over its $m$ rounds. We leave \arm{Session} probabilities unstandardized, since each of its calls returns a single probability.

\noindent{\bf Judgments, calls, and the discovery pool.}
A product's prediction rests on one judgment for \arm{Summary}, five for \arm{Batch summary}, and 80 for both session methods. The design choices (which statistics the summary uses, how many sessions it is computed from, how many products share a prompt, and how the request is worded) were fixed on a separate discovery pool of 280 Coveo products with GPT-5.6, before any evaluation product was used; we report ablations on that pool in Appendix~\ref{app:ablations}.

\section{Results}
\label{sec:results}

\begin{figure}[tb]
\centering
\includegraphics[width=\linewidth]{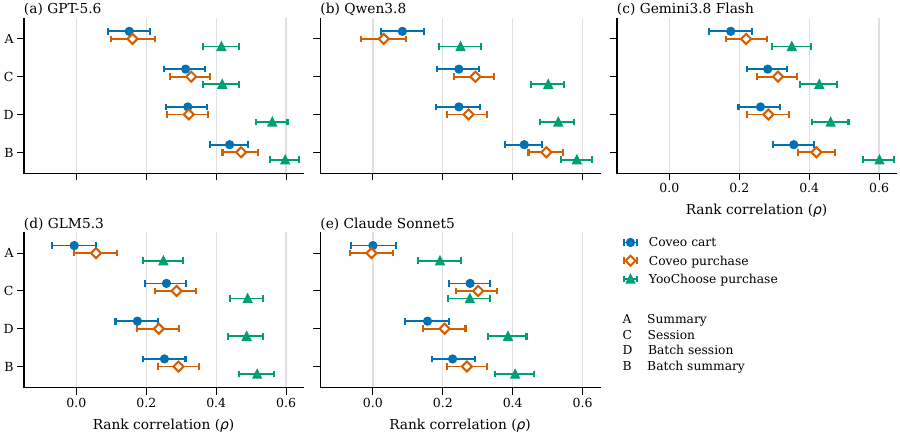}
\caption{Rank correlations of the four main methods for five LLMs, with 95\% product-bootstrap intervals. Each model uses its common support in a 1,000-product pool (Table~\ref{tab:full-coverage}); comparisons in the text use paired intervals.}
\label{fig:models}
\end{figure}

Figure~\ref{fig:models} shows the four main methods for all five models; we use the labels A, B, C and D of Figure~\ref{fig:designs} when naming a paired comparison. All paired differences across settings, with confidence intervals, are in Appendix~\ref{app:contrasts}.

\subsection{Comparing the designs}
\label{sec:isolated}\label{sec:comparative}\label{sec:cost}

\begin{figure}[tb]
\centering
\includegraphics[width=\linewidth]{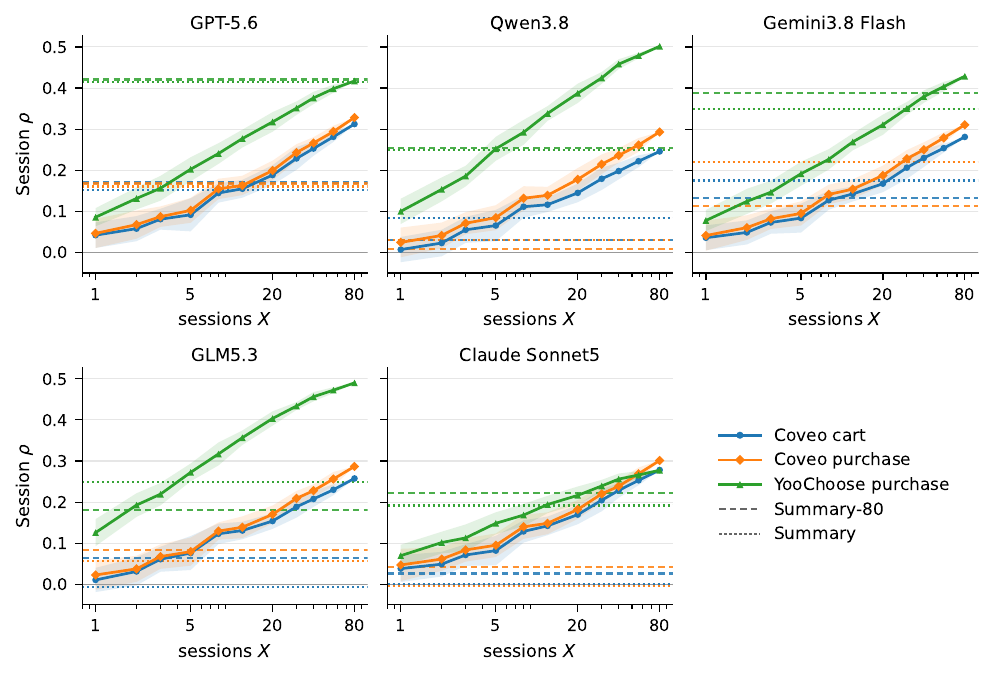}
\small\caption{\arm{Session} using \(X\) of each product's 80 sessions. Each point averages random draws of \(X\) stored judgments (band \(\pm1\) sd). Dashed: \arm{Summary-80}; dotted: \arm{Summary}.}
\label{fig:depth}
\end{figure}

\noindent{\bf Session judgments outrank summaries.}
\arm{Session} averages judgments of 80 sampled sessions per product, whereas \arm{Summary} is computed from the product's full history. To compare the two representations on the same evidence, we recompute the summary statistics from exactly those 80 sessions (\arm{Summary-80}, A80; Appendix~\ref{app:diagnostics}). \arm{Session} outperforms it on both datasets: the mean \arm{Session}$-$\arm{Summary-80} difference is \(0.190\;[0.1467,0.2288]\) for Coveo cart, \(0.221\;[0.1784,0.2610]\) for Coveo purchase and \(0.130\;[0.0864,0.1710]\) for YooChoose. The model-specific intervals lie above zero in 12 of the 15 settings, and none lies below zero. Adding the rest of each product's history to \arm{Summary-80}, which gives \arm{Summary}, does not close the gap: \arm{Session}$-$\arm{Summary} is \(0.194\), \(0.211\) and \(0.132\). However, the session advantage requires many judgments. When we recompute \arm{Session} from \(X\leq 80\) of its stored judgments per product, averaging over random draws so that no new model calls are needed (Figure~\ref{fig:depth}), accuracy rises steadily and reaches \arm{Summary-80} within 1 to 20 sessions on Coveo and, on YooChoose, within 2 to 56 sessions for four models; GPT's curve ends just below it, at \(0.417\) against \(0.421\). %

\noindent{\bf Comparison greatly improves summaries.}
Batching and standardizing product-summary predictions improves ranking for every model and outcome on both datasets. The mean \arm{Batch summary}$-$\arm{Summary} gain is \(0.260\;[0.2104,0.3082]\) for Coveo cart, \(0.297\;[0.2520,0.3387]\) for Coveo purchase, and \(0.250\;[0.2139,0.2843]\) for YooChoose purchase, and all 15 model-specific paired intervals are above zero. The gain does not depend on standardization or averaging: a single comparative call per product, neither standardized nor averaged, obtains exactly one judgment per product, as \arm{Summary} does, and already improves on \arm{Summary} by \(0.149\), \(0.179\) and \(0.127\), above zero in all 15 settings. Standardization and averaging then add gains that largely overlap: each adds \(0.06\) to \(0.09\) on its own, and together they add \(0.11\) to \(0.12\) (Appendix~\ref{app:normavg}). The overall gain from batching is largest where isolated summaries work poorly: on Coveo, Sonnet rises from \(0.000\) and \(-0.004\), no better than chance, to \(0.228\) and \(0.269\) on identical product statistics.

\noindent{\bf Comparison helps sessions less consistently.}
\arm{Batch session} applies the same design to single sessions, one from each of 40 products per call, and its effect depends on the dataset. The mean \arm{Batch session}$-$\arm{Session} difference is \(-0.044\) and \(-0.040\) on the two Coveo outcomes and \(0.062\) on YooChoose. On Coveo, no model gains and five of the ten intervals lie below zero; on YooChoose, GPT, Gemini and Sonnet gain, GPT by \(0.143\) and Sonnet by \(0.109\), and Qwen and GLM do not. Compared directly, the gain from comparison is larger for summaries than for sessions in 14 of the 15 settings, by \(0.30\), \(0.34\) and \(0.19\) on average, and indistinguishable for GPT on YooChoose (Appendix~\ref{app:did}).

\noindent{\bf Batching a product's own sessions into one call lowers accuracy.}
\arm{Session} is expensive, at 80 calls per product. The obvious way to save calls is to place a product's 80 sessions in one call and ask for 80 probabilities (\arm{Same-product batch}), which keeps the requested judgments and uses one-eightieth of the calls. However, it ranks worse. On the same sampled sessions and under the same system message, it trails \arm{Session} by \(0.144\), \(0.161\) and \(0.104\) on the three outcomes, with the paired interval above zero in 14 of the 15 settings; the exception is GPT on YooChoose (Table~\ref{tab:batchedc}). Asking instead for a single product rate from the same call (\arm{Session bag}) does no better (Appendix~\ref{app:diagnostics}).

\noindent{\bf Comparative summaries lead.}
Among the four main methods, \arm{Batch summary} has the highest rank correlation in 12 of the 15 settings; in the other three, \arm{Session} is higher, but not by a margin our intervals distinguish from zero. \arm{Batch summary} leads \arm{Session} by \(0.066\), \(0.086\) and \(0.118\) on the three outcomes, with the paired interval above zero in 10 of the 15 settings. Even one standardized grouping of \arm{Batch summary}, which requests a single judgment per product, is ahead of \arm{Session}'s 80 judgments per product on average: the differences are \(0.040\), \(0.055\) and \(0.056\), the last two with intervals above zero (Appendix~\ref{app:normavg}).

\begin{figure}[tb]
\centering
\includegraphics[width=\linewidth]{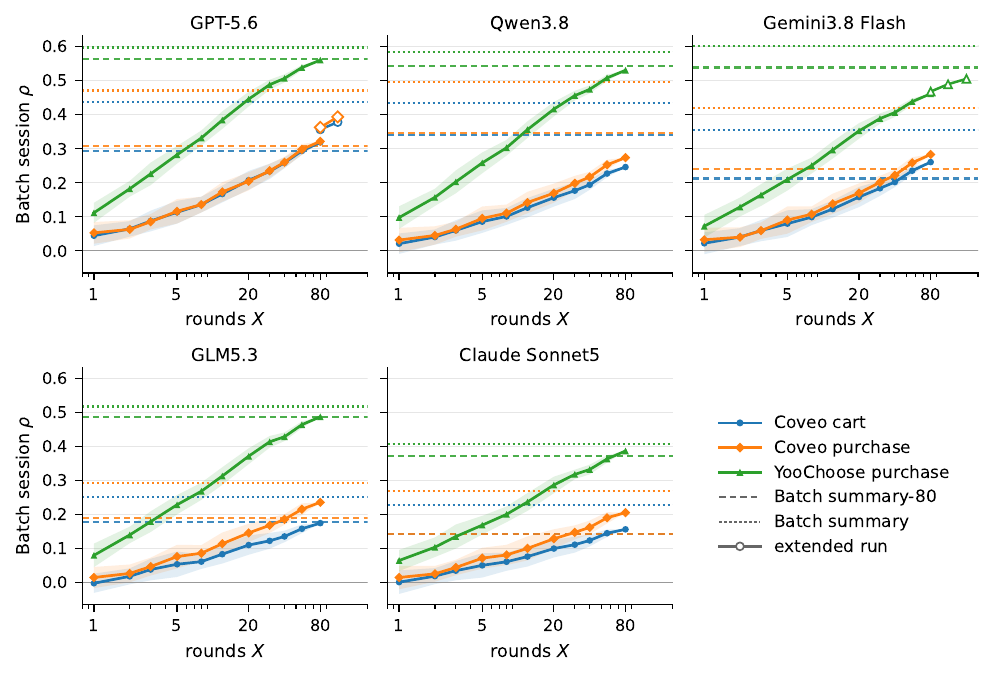}
\small\caption{\arm{Batch session} using \(X\) of its 80 rounds. Each point averages random draws of \(X\) stored judgments (band \(\pm1\) sd). Dashed: \arm{Batch summary-80}; dotted: \arm{Batch summary}. Hollow markers: extended runs (Appendix~\ref{app:depth}).}
\label{fig:batchrow}
\end{figure}

\arm{Batch summary} also leads \arm{Batch session}, by \(0.110\), \(0.126\) and \(0.056\), but the lead comes from the full history: \arm{Batch summary-80}, which applies the comparative procedure to summaries of the same 80 sampled sessions, differs from \arm{Batch session} by \(0.002\), \(-0.019\) and \(0.015\) on average, with significant model-level differences in both directions, and trails \arm{Session} on Coveo by \(0.042\) and \(0.059\) while leading on YooChoose by \(0.078\) (Appendix~\ref{app:contrasts}). Figure~\ref{fig:batchrow} recomputes \arm{Batch session} from \(X\) of its 80 rounds. Accuracy rises with \(X\), but \arm{Batch session} needs 30 to 80 rounds to reach \arm{Batch summary-80}, for the models that reach it at all, and no model reaches \arm{Batch summary} within 80 rounds; doubling the sample to 160 sessions, for GPT on Coveo and Gemini on YooChoose, does not close that gap either (Appendix~\ref{app:depth}).

\noindent{\bf Inference cost.}
Per model and dataset, \arm{Session} makes 80,000 calls, \arm{Batch session} 2,000 for the same judgments, and \arm{Batch summary} 125. \arm{Session} costs 1.6 to 3.2 times \arm{Batch session}, and \arm{Batch summary-80}, whose five-model mean is within \(0.02\) of \arm{Batch session}'s on the same 80 sessions, costs at most 12 to 18\% as much for every model: requesting fewer judgments saves far more than batching them (Appendix~\ref{app:usage}).

\noindent{\bf Supervised and single-feature references.}
We train ridge regression, a random forest, and gradient boosting on thousands of products outside the evaluation and discovery sets, using the traffic counts withheld from the LLM prompts. All of them lead every model's \arm{Batch summary} on Coveo. The same holds for the tree ensembles on YooChoose (Appendix~\ref{app:supervised}). Nevertheless, the best \arm{Batch summary} reaches 66\% of the forest's correlation on Coveo add-to-cart, 89\% on Coveo purchase, and 85\% on YooChoose. The forest matches these levels with 15--25, 200--400 and 100--200 labeled products (Table~\ref{tab:holdout-curve}). A single-feature rule fixed on the discovery pool, the average price of the most-considered competing product, beats every LLM method on Coveo add-to-cart but trails the best comparative summaries on both purchase outcomes (Table~\ref{tab:holdout}). The LLM methods thus fare best on purchase, the outcome that matters more in practice.

\section{Representation and in-context inference: a Bayesian perspective}
\label{sec:mechanism}

Comparison improves on \arm{Summary} in every setting, has mixed effects relative to \arm{Session}, and lowers accuracy when all sessions in a call belong to one product. To explain these observations, {we propose a model of the LLM as a Bayesian predictor whose pretrained knowledge provides prior beliefs about the store and its shoppers. In this model, the other summaries or sessions in the prompt provide evidence that updates these beliefs and thereby changes the prediction.}

{An {\it environment} $\theta$ describes a store's products and their characteristics, its shoppers, and how browsing behavior relates to cart additions and purchases. The online shop behind the Coveo data is one example. In our Bayesian interpretation, we represent the LLM's knowledge acquired during pretraining by a prior distribution $\pi$ over possible environments. This distribution describes which environments the model considers more or less likely before seeing the behavioral evidence in the prompt.}

{Each prompt contains one or more {\it entries}, where an entry is either a summary of one product's browsing sessions or a record of a single session. For the entry whose prediction we are considering, let $x_i$ denote its features and let $C$ denote the other entries in the prompt. These other entries may be summaries of other products, sessions of other products, or other sessions of the same product.}

{Let $p$ denote the conversion probability being predicted. We represent the prediction from $x_i$ alone as $\E_\pi[p\mid x_i]$. This prediction averages over possible environments, weighted by how likely each environment is after observing $x_i$:}
\[
 {\E_\pi[p\mid x_i]=\int \E[p\mid x_i,\theta]\,\pi(\theta\mid x_i)\,d\theta.}
\]

{Adding the context $C$ provides further evidence about the environment. We assume that, given $\theta$ and $x_i$, $C$ provides no additional information about $p$. Under this assumption, context changes the prediction by updating the distribution over environments:}
\begin{equation}
 \E_\pi[p\mid x_i,C]=\int \E[p\mid x_i,\theta]\;\pi(\theta\mid x_i,C)\,d\theta .
 \label{eq:lens}
\end{equation}
{In this model, the context changes the prediction by changing which environments the model considers more or less likely. We call this update} prior refinement.

{\bf Refinement decomposition.} {For a given entry, we can decompose how its prediction changes when other randomly selected entries are included in the prompt into (i) the average change across contexts and (ii) the deviation from that average for a particular context.} We write $P_\theta(\cdot\mid x_i)$ for the distribution of the context $C$ that our design assigns to entry $i$ when its companions are drawn at random in environment $\theta$. Then, for the environment $\theta$ that our data come from, we decompose the change from adding the context $C$ as
\begin{equation}
 \E_\pi[p\mid x_i,C]-\E_\pi[p\mid x_i]=m_\theta(x_i)+\eta_\theta(x_i,C),
 \label{eq:decomp}
\end{equation}
where $m_\theta(x_i)=\E_{C\sim P_\theta(\cdot\mid x_i)}\bigl[\E_\pi[p\mid x_i,C]-\E_\pi[p\mid x_i]\bigr]$ is the average change over these draws and $\eta_\theta(x_i,C)$, which has mean zero under $P_\theta(\cdot\mid x_i)$, is the deviation for the companions actually drawn. We call $m_\theta$ the mean shift and $\eta_\theta$ the context noise. The mean shift need not be zero: the prediction without context, $\E_\pi[p\mid x_i]$, averages over the environments the model considers plausible, whereas our contexts come from the single environment behind our data.\footnote{By the law of iterated expectations, the expectation of the change over the prior is zero: $\E_{\theta\sim\pi(\cdot\mid x_i)}\E_{C\sim P_\theta(\cdot\mid x_i)}\bigl[\E_\pi[p\mid x_i,C]-\E_\pi[p\mid x_i]\bigr]=0$. In other words, $m_\theta(x_i)$ averages to zero over $\theta$ drawn from $\pi(\theta\mid x_i)$, but not in general for the environment our data come from.} The shift helps when it moves the prediction toward $\E[p\mid x_i,\theta]$, offsetting its bias. For an ideal Bayesian predictor, one that computes these expectations exactly, conditioning on $C$ cannot increase the expected squared error on average over environments drawn from its prior. Because we score ranking, only differences in the shift across products matter.

{\bf Properties of the decomposition.} For an ideal Bayesian predictor, the decomposition has two properties. {\it First}, the mean shift is nonzero only if (a) the prior given $x_i$, $\pi(\theta\mid x_i)$, does not already concentrate on a single environment and (b) the context is informative about the environment given $x_i$. {How much context changes a prediction depends on how the entry's interpretation varies across environments and how much the other entries change the model's beliefs about the environment.} The first factor depends on the entry's representation. The interpretation of a summary statistic, such as a product's average viewing time, may strongly depend on the environment, whereas a repeated view or a search in a single session signals customer intent more or less similarly across environments. The second factor depends on the nature of the context. Because we evaluate rankings, what matters is how much the context informs how products compare. {Summaries of different products show differences in their typical browsing behavior. Single sessions from different products may instead differ mainly because shoppers behave differently, making differences between products harder to discern. If all entries describe sessions of the same product, there is no variation across products among the entries. We call the variation across products represented in a prompt} {\it task variation}, {and expect it to help the model judge how products compare} (Appendix~\ref{app:mechanism}). %

{\it Second}, the context noise comes only from which companions are drawn, so averaging over the $n$ independent contexts in which a product appears divides its variance by $n$ ($n=5$ for \arm{Batch summary}, 80 for \arm{Batch session} and 1 for \arm{Same-product batch}), while the mean shift $m_\theta$ does not change. Comparison therefore trades a possible reduction in bias, through $m_\theta$, against added variance, through $\eta_\theta$: a second bias--variance tradeoff, alongside the choice of where to aggregate. In this view, standardization and averaging reduce the context noise. Part of $\eta_\theta$ is shared by all entries in a call: {a common upward or downward shift in the predictions within a call.} Standardizing within the call (equation~\eqref{eq:z}) removes this shared {shift} when the call includes several products.%

{\bf The perspective in action.} The results of Section~\ref{sec:comparative} are consistent with the decomposition perspective in the following respects.
\begin{enumerate}[leftmargin=*,label=(\arabic*),itemsep=0pt,topsep=0pt,parsep=0pt,partopsep=0pt]
\item {\bf Gains follow task variation.} {We expect larger gains from comparison when task variation is informative about differences in product conversion rates. To assess this informativeness, we use a supervised model to score each product from a single entry and compute the rank correlation between these scores and observed conversion rates. We consider three types of entries (or representations): a single session record, a summary of 80 sessions, or a full-history summary. The supervised model used to score each product is trained on a subset of the other products excluding the product being scored. We use this correlation as a proxy for task-relevant variation: a higher correlation means the representation more clearly distinguishes products with different rates} (Table~\ref{tab:context} in Appendix~\ref{app:context}). For every outcome, the gain over the isolated counterpart rises with {this correlation}, from \(-0.04\) to \(+0.06\) for single sessions of different products, through \(+0.15\) to \(+0.21\) for summaries of 80 sessions, to \(+0.25\) to \(+0.30\) for full-history summaries. {In contrast, \arm{Same-product batch} places one product's sessions in a single call, providing no variation across products, and trails \arm{Session} by \(0.10\) to \(0.16\). This interpretation also fits the difference between datasets (Section~\ref{sec:comparative}): comparing single sessions helps on YooChoose but not on Coveo. On Coveo, only 13\% (cart) and 10\% (purchase) of the variance in supervised session scores reflects differences between products, compared with 41\% on YooChoose} (Appendix~\ref{app:context}). {Thus, a call of 40 YooChoose sessions more clearly represents differences between products.}
\item {\bf Precise summaries matter more after comparison.} If comparison refines the prior, more precise summaries should help more after that refinement, since precision matters more {once the model can interpret the input features in relation to the store.} Computing the comparative summaries from full histories rather than from the 80 sampled sessions raises the rank correlation by \(0.108\), \(0.145\) and \(0.041\) (\arm{Batch summary}$-$\arm{Batch summary-80}), above zero in 14 of the 15 settings, while the same change leaves the isolated summaries essentially unchanged (\arm{Summary}$-$\arm{Summary-80} is \(-0.004\), \(0.010\) and \(-0.002\)).
\item {\bf Same-product context adds little and shares its noise.} A call of one product's sessions has no variation across products, so we expect it to provide little refinement that helps ranking (Appendix~\ref{app:mechanism}). It does add noise: all 80 judgments share one context ($n=1$), so the noise does not average out, and standardization cannot remove it without also removing the product's level needed for ranking. We therefore expect a net loss, which is what we observe (Section~\ref{sec:comparative}). On Coveo the shared noise is directly visible: when a product's 80 sessions are judged in one call, their predictions become much more correlated with each other, and the common part is unrelated to the product's observed rate (Appendix~\ref{app:batchedc}).
\item {\bf {Comparing summaries helps even without standardization or averaging.}} Standardization and averaging bring largely overlapping gains to the comparative summaries (Section~\ref{sec:comparative}), consistent with both reducing common context noise.\footnote{Standardization also divides by the call's spread, so scores depend partly on the other products in the call.} Standardization changes \arm{Batch session}, which averages 80 calls, by less than \(0.01\) on average, with intervals containing zero (Appendix~\ref{app:context}). Neither device changes the sign of any comparison: a single unstandardized comparative call already improves summaries in all 15 settings, and without standardization, \arm{Batch session} still trails \arm{Session} on Coveo by \(0.038\) and \(0.035\) and leads on YooChoose by \(0.058\).
\end{enumerate}

\noindent{\bf Model-specific bias.} LLMs may not be ideal Bayesian predictors and may respond to context in ways the decomposition does not predict. A context that says little about how products compare offers a check: for an ideal predictor, it should produce little mean shift. {If the loss is mainly caused by random variation in context, we would expect it to diminish as more independently drawn contexts are averaged.} On Coveo, single sessions of different products are such a context (item (1)). There, \arm{Batch session} stays within 0.035 of \arm{Session} at every number of rounds \(X\leq 80\) for GPT, Gemini and Qwen, but trails it increasingly for Sonnet and GLM, by 0.122 and 0.083 on cart and 0.096 and 0.051 on purchase at 80 (although 80 rounds cannot rule out noise that averages away more slowly; Appendix~\ref{app:context}).

\section{Conclusion}
\label{sec:discussion}

This paper asks how to organize LLM inference when a product-level rate must be predicted from many individual browsing sessions. Our results point to two bias--variance tradeoffs. The first concerns where to aggregate. A summary lets a single judgment draw on a product's full history but discards detail; judging sessions one at a time keeps that detail but requires many noisy judgments, at a cost that grows with the amount of evidence used. The second concerns what context accompanies each judgment. Other entries in the same call can refine the model's prior about the environment and so reduce bias, but they also add noise that depends on which entries happen to be present.

In our interpretation, how these tradeoffs balance depends on how much the entries in a call vary across products. Summaries of different products vary mainly across products, and comparing them helps in every setting we examine. Single sessions vary mainly across shoppers, and comparing them helps only on YooChoose, where sessions vary more across products than on Coveo. The practical implication is to compare entries that differ mainly across the units being predicted, here products. In our experiments, comparative full-history summaries rank products best while requiring an order of magnitude fewer judgments per product than either session method. With the best model for each outcome, they reach 66--89\% of the rank correlation of the strongest supervised benchmark trained on thousands of labeled products.

These conclusions have limits (Appendix~\ref{app:protocol}). The prior-refinement account is our interpretation of the results: we check it only through its observable consequences, and the ordering by task variation rests on four kinds of context per outcome (Appendix~\ref{app:context}). We fixed the design choices on a discovery pool using GPT-5.6 alone and applied them unchanged to the other four models. The models also differ in how much they are affected by context that says little about how products compare, and five models are too few to identify why. The session methods were still improving at the largest number of sessions per product we could afford, so larger budgets could narrow their gap to the summaries. Finally, both datasets are e-commerce logs; in other domains, behavioral records may vary differently across items and individuals, which in our interpretation would change how much comparison helps.

Our task is also retrospective: the models predict rates observed under the existing offer. Whether the ordering of methods carries over to predicting responses to a changed offer, such as a new price, can only be tested with experimental data. We see this as the most useful next step toward using LLMs for counterfactual aggregate predictions.

\clearpage
\section*{Reproducibility statement}
Both datasets are public (Section~\ref{sec:design}). We reproduce every system message and a rendered user message for the four methods and \arm{Session bag} on both datasets in Appendix~\ref{app:prompts}, generated from the code that issued the calls rather than transcribed; the matched-evidence variants reuse the summary prompts. We give the pools and seeds, feature definitions, model endpoints and execution dates, decoding and retry settings, the coverage gate, the rank convention and the paired bootstrap in Appendix~\ref{app:protocol}, the supervised pipeline in Appendix~\ref{app:supervised}, and inference usage in Appendix~\ref{app:usage}. Every number in the results tables and figures is produced by a script from the saved predictions, including the doubled-sample runs; Figures~\ref{fig:depth} and~\ref{fig:batchrow} re-aggregate them without new inference and, at full depth, reproduce the reported arms. Two limits apply. The models are served through provider endpoints rather than fixed snapshots, and their outputs vary between calls even at temperature zero, so a rerun will differ in individual judgments. Coveo's pool and session-sample seeds came from a process-dependent Python string hash, so the sample cannot be redrawn; we instead record the identifiers of the sampled sessions, from which a script rebuilds it. The code, call designs, product pools and sampled-session identifiers, saved model predictions, and the outputs the paper was built from are available from the authors on request. These materials do not include the session samples or feature tables; the scripts rebuild them from the original logs, available from the dataset providers, and regenerate the reported tables, figures, and numbers.

\section*{Statement on the use of generative AI}
We used generative AI tools (Claude and ChatGPT), separately from the models evaluated in the study, to execute the methodologies and perform analyses that help interpret results. We did not use them to generate data or to formulate or prove mathematical claims. We also used them to produce initial drafts of sections of the appendices, to edit for readability, and to identify and format references. We reviewed all AI-assisted work: every reported number is regenerated by script from the saved predictions, and citations were checked against the original publications. We take responsibility for the final content of this work.

\bibliographystyle{iclr2027_conference}
\bibliography{references}

@misc{wolf2026partition,
  title = {Partition, Prompt, Aggregate: Statistical Self-Consistency in Language Models},
  author = {Wolf, Patrik and {Kleine Buening}, Thomas and Krause, Andreas and Mendler-D{\"u}nner, Celestine},
  year = {2026},
  eprint = {2607.15277},
  archivePrefix = {arXiv},
  primaryClass = {cs.CL},
  note = {Version 1, July 16, 2026},
  url = {https://arxiv.org/abs/2607.15277v1}
}

@misc{huang2026virtual,
  title = {{LLM}-Powered Virtual Population for Demand Simulation and Pricing},
  author = {Huang, Chengpiao and Wang, Kaizheng},
  year = {2026},
  eprint = {2606.16183},
  archivePrefix = {arXiv},
  note = {Version 1, June 15, 2026},
  url = {https://arxiv.org/abs/2606.16183v1}
}

@misc{li2026simgym,
  title = {{SimGym}: A Framework for {A/B} Test Simulation in E-Commerce with Traffic-Grounded {VLM} Agents},
  author = {Li, Han and Malik, Vibhor and Foumani, Zahra Zanjani and Castelo, Alberto and Xie, Shuang and Fan, Ailin and Koay, Keat Yang and Zhu, Yuanzheng and Feghhi, Meysam and Uliana, Ronie and Zhang, Zhaoyu and Martins, Angelo Ocana and Zhao, Mingyu and Pelland, Francis and Faerman, Jonathan and LeBlanc, Nikolas and Glazer, Aaron and McNamara, Andrew and Wu, Zhong and Wang, Lingyun},
  year = {2026},
  eprint = {2605.19219},
  archivePrefix = {arXiv},
  note = {Version 1, May 19, 2026},
  url = {https://arxiv.org/abs/2605.19219v1}
}

@misc{gui2025challenge,
  title = {The Challenge of Using {LLMs} to Simulate Human Behavior: A Causal Inference Perspective},
  author = {Gui, George and Toubia, Olivier},
  year = {2025},
  eprint = {2312.15524},
  archivePrefix = {arXiv},
  note = {Version 3, November 23, 2025; first circulated in 2023},
  url = {https://arxiv.org/abs/2312.15524v3}
}

@article{forman2008quantifying,
  title = {Quantifying Counts and Costs via Classification},
  author = {Forman, George},
  journal = {Data Mining and Knowledge Discovery},
  volume = {17},
  pages = {164--206},
  year = {2008},
  doi = {10.1007/s10618-008-0097-y},
  url = {https://link.springer.com/article/10.1007/s10618-008-0097-y}
}

@article{perezmon2025histnetq,
  title = {Quantification Using Permutation-Invariant Networks Based on Histograms},
  author = {P{\'e}rez-Mon, Olaya and Moreo, Alejandro and del Coz, Juan Jos{\'e} and Gonz{\'a}lez, Pablo},
  journal = {Neural Computing and Applications},
  volume = {37},
  pages = {3505--3520},
  year = {2025},
  doi = {10.1007/s00521-024-10721-1},
  note = {Published online December 14, 2024; February 2025 issue},
  url = {https://link.springer.com/article/10.1007/s00521-024-10721-1}
}

@article{szabo2016distribution,
  title = {Learning Theory for Distribution Regression},
  author = {Szab{\'o}, Zolt{\'a}n and Sriperumbudur, Bharath K. and P{\'o}czos, Barnab{\'a}s and Gretton, Arthur},
  journal = {Journal of Machine Learning Research},
  volume = {17},
  number = {152},
  pages = {1--40},
  year = {2016},
  url = {https://jmlr.org/papers/v17/14-510.html}
}

@inproceedings{lin2024batchprompt,
  title = {{BatchPrompt}: Accomplish More with Less},
  author = {Lin, Jianzhe and Diesendruck, Maurice and Du, Liang and Abraham, Robin},
  booktitle = {International Conference on Learning Representations},
  year = {2024},
  url = {https://proceedings.iclr.cc/paper_files/paper/2024/hash/5d8c01de2dc698c54201c1c7d0b86974-Abstract-Conference.html}
}

@inproceedings{aher2023,
 author={Aher, Gati V. and Arriaga, Rosa I. and Kalai, Adam Tauman},
 title={Using Large Language Models to Simulate Multiple Humans and Replicate Human Subject Studies},
 booktitle={Proceedings of the 40th International Conference on Machine Learning},
 volume={202}, series={Proceedings of Machine Learning Research}, pages={337--371}, year={2023},
 url={https://proceedings.mlr.press/v202/aher23a.html}
}

@inproceedings{brand2024,
 author={Brand, James and Israeli, Ayelet and Ngwe, Donald},
 title={Using {GPT} for Market Research},
 booktitle={Proceedings of the 25th ACM Conference on Economics and Computation},
 year={2024}, doi={10.1145/3670865.3673479}
}

@inproceedings{cheng2023,
  author={Cheng, Zhoujun and Kasai, Jungo and Yu, Tao},
  title={Batch Prompting: Efficient Inference with Large Language Model {APIs}},
  booktitle={Proceedings of the 2023 Conference on Empirical Methods in Natural Language Processing: Industry Track},
  pages={792--810},
  year={2023},
  doi={10.18653/v1/2023.emnlp-industry.74},
  url={https://aclanthology.org/2023.emnlp-industry.74/}
}

@misc{coveo2021,
 author={{Coveo Solutions Inc.}},
 title={{SIGIR} E-Commerce Data Challenge: Dataset and Documentation},
 year={2021},
 url={https://github.com/coveooss/SIGIR-ecom-data-challenge}
}

@inproceedings{hegselmann2023,
 author={Hegselmann, Stefan and Buendia, Alejandro and Lang, Hunter and Agrawal, Monica and Jiang, Xiaoyi and Sontag, David},
 title={{TabLLM}: Few-Shot Classification of Tabular Data with Large Language Models},
 booktitle={Proceedings of the 26th International Conference on Artificial Intelligence and Statistics},
 volume={206}, series={Proceedings of Machine Learning Research},pages={5549--5581},year={2023},
 url={https://proceedings.mlr.press/v206/hegselmann23a.html}
}

@inproceedings{lu2026,
 author={Lu, Yuxuan and Huang, Jing and Han, Yan and Yao, Bingsheng and Bei, Sisong and Xie, Yaochen and Sang, Yisi and He, Qi and Wang, Dakuo},
 title={Can {LLM} Agents Simulate Multi-Turn Human Behavior? Evidence from Real Online Customer Behavior Data},
 booktitle={Proceedings of the 64th Annual Meeting of the Association for Computational Linguistics (Volume 1: Long Papers)},
 pages={43961--43977},year={2026},doi={10.18653/v1/2026.acl-long.2034},
 url={https://aclanthology.org/2026.acl-long.2034/}
}

@article{tagliabue2021,
 author={Tagliabue, Jacopo and Greco, Ciro and Roy, Jean-Francis and Yu, Bingqing and Chia, Patrick John and Bianchi, Federico and Cassani, Giovanni},
 title={{SIGIR} 2021 E-Commerce Workshop Data Challenge},
 journal={arXiv preprint arXiv:2104.09423},year={2021},
 url={https://arxiv.org/abs/2104.09423}
}

@inproceedings{wang2026,
 author={Wang, Ziyi and Lu, Yuxuan and Li, Wenbo and Amini, Amirali and Sun, Bo and Bart, Yakov and Lyu, Weimin and Gesi, Jiri and Wang, Tian and Huang, Jing and Su, Yu and Ehsan, Upol and Alikhani, Malihe and Li, Toby Jia-Jun and Chilton, Lydia and Wang, Dakuo},
 title={{OPeRA}: A Dataset of Observation, Persona, Rationale, and Action for Evaluating {LLMs} on Human Online Shopping Behavior Simulation},
 booktitle={Proceedings of the 64th Annual Meeting of the Association for Computational Linguistics (Volume 1: Long Papers)},
 pages={43942--43960},year={2026},doi={10.18653/v1/2026.acl-long.2033},
 url={https://aclanthology.org/2026.acl-long.2033/}
}

@inproceedings{zhao2021,
 author={Zhao, Zihao and Wallace, Eric and Feng, Shi and Klein, Dan and Singh, Sameer},
 title={Calibrate Before Use: Improving Few-Shot Performance of Language Models},
 booktitle={Proceedings of the 38th International Conference on Machine Learning},
 volume={139},series={Proceedings of Machine Learning Research},pages={12697--12706},year={2021},
 url={https://proceedings.mlr.press/v139/zhao21c.html}
}

@inproceedings{zhou2024,
 author={Zhou, Han and Wan, Xingchen and Proleev, Lev and Mincu, Diana and Chen, Jilin and Heller, Katherine and Roy, Subhrajit},
 title={Batch Calibration: Rethinking Calibration for In-Context Learning and Prompt Engineering},
 booktitle={International Conference on Learning Representations},year={2024},
 url={https://arxiv.org/abs/2309.17249}
}

@inproceedings{Y,
 author={Ben-Shimon, David and Tsikinovsky, Alexander and Friedmann, Michael and Shapira, Bracha and Rokach, Lior and Hoerle, Johannes},
 title={{RecSys} Challenge 2015 and the {YOOCHOOSE} Dataset},
 booktitle={Proceedings of the 9th ACM Conference on Recommender Systems},
 pages={357--358}, year={2015}, doi={10.1145/2792838.2798723},
 url={https://dbs-home-page.s3.amazonaws.com/dbs/papers/recsyschallengeyc2p.pdf}
}

@inproceedings{sun2023,
 author={Sun, Weiwei and Yan, Lingyong and Ma, Xinyu and Wang, Shuaiqiang and Ren, Pengjie and Chen, Zhumin and Yin, Dawei and Ren, Zhaochun},
 title={Is {ChatGPT} Good at Search? Investigating Large Language Models as Re-Ranking Agents},
 booktitle={Proceedings of the 2023 Conference on Empirical Methods in Natural Language Processing},
 pages={14918--14937}, year={2023},
 url={https://aclanthology.org/2023.emnlp-main.923/}
}

@inproceedings{qin2024,
 author={Qin, Zhen and Jagerman, Rolf and Hui, Kai and Zhuang, Honglei and Wu, Junru and Yan, Le and Shen, Jiaming and Liu, Tianqi and Liu, Jialu and Metzler, Donald and Wang, Xuanhui and Bendersky, Michael},
 title={Large Language Models are Effective Text Rankers with Pairwise Ranking Prompting},
 booktitle={Findings of the Association for Computational Linguistics: NAACL 2024},
 pages={1504--1518}, year={2024},
 url={https://aclanthology.org/2024.findings-naacl.97/}
}

@inproceedings{long2025precise,
  author = {Long, Kehan and Li, Shasha and Xu, Chen and Tang, Jintao and Wang, Ting},
  title = {Precise Zero-Shot Pointwise Ranking with {LLMs} through Post-Aggregated Global Context Information},
  booktitle = {Proceedings of the 48th International ACM SIGIR Conference on Research and Development in Information Retrieval},
  year = {2025},
  url = {https://arxiv.org/abs/2506.10859}
}

@inproceedings{xie2022explanation,
  author = {Xie, Sang Michael and Raghunathan, Aditi and Liang, Percy and Ma, Tengyu},
  title = {An Explanation of In-Context Learning as Implicit {Bayesian} Inference},
  booktitle = {International Conference on Learning Representations},
  year = {2022},
  url = {https://arxiv.org/abs/2111.02080}
}

@inproceedings{min2022rethinking,
  author = {Min, Sewon and Lyu, Xinxi and Holtzman, Ari and Artetxe, Mikel and Lewis, Mike and Hajishirzi, Hannaneh and Zettlemoyer, Luke},
  title = {Rethinking the Role of Demonstrations: What Makes In-Context Learning Work?},
  booktitle = {Proceedings of the 2022 Conference on Empirical Methods in Natural Language Processing},
  year = {2022},
  pages = {11048--11064},
  publisher = {Association for Computational Linguistics},
  doi = {10.18653/v1/2022.emnlp-main.759},
  url = {https://aclanthology.org/2022.emnlp-main.759/}
}

@inproceedings{zhuang2024setwise,
  author={Zhuang, Shengyao and Zhuang, Honglei and Koopman, Bevan and Zuccon, Guido},
  title={A Setwise Approach for Effective and Highly Efficient Zero-shot Ranking with Large Language Models},
  booktitle={Proceedings of the 47th International ACM SIGIR Conference on Research and Development in Information Retrieval},
  year={2024},
  doi={10.1145/3626772.3657813},
  url={https://arxiv.org/abs/2310.09497}
}

@inproceedings{zaheer2017deepsets,
  title     = {{Deep Sets}},
  author    = {Zaheer, Manzil and
               Kottur, Satwik and
               Ravanbakhsh, Siamak and
               Poczos, Barnabas and
               Salakhutdinov, Ruslan and
               Smola, Alexander J.},
  booktitle = {Advances in Neural Information Processing Systems},
  volume    = {30},
  year      = {2017},
  url       = {https://papers.nips.cc/paper/6931-deep-sets}
}

@inproceedings{lee2019settransformer,
  title     = {{Set Transformer}: A Framework for Attention-based
               Permutation-Invariant Neural Networks},
  author    = {Lee, Juho and
               Lee, Yoonho and
               Kim, Jungtaek and
               Kosiorek, Adam and
               Choi, Seungjin and
               Teh, Yee Whye},
  booktitle = {Proceedings of the 36th International Conference
               on Machine Learning},
  series    = {Proceedings of Machine Learning Research},
  volume    = {97},
  pages     = {3744--3753},
  publisher = {PMLR},
  year      = {2019},
  url       = {https://proceedings.mlr.press/v97/lee19d.html}
}

@inproceedings{gupta2024context,
  title     = {Context is Environment},
  author    = {Gupta, Sharut and
               Jegelka, Stefanie and
               Lopez-Paz, David and
               Ahuja, Kartik},
  booktitle = {International Conference on Learning Representations},
  year      = {2024},
  url       = {https://openreview.net/forum?id=8VPWfqtQMX}
}

@inproceedings{gadetsky2025unsupervised,
  title     = {Large ({Vision}) Language Models are Unsupervised
               In-Context Learners},
  author    = {Gadetsky, Artyom and
               Atanov, Andrei and
               Jiang, Yulun and
               Gao, Zhitong and
               Hosseini Mighan, Ghazal and
               Zamir, Amir and
               Brbic, Maria},
  booktitle = {International Conference on Learning Representations},
  year      = {2025},
  url       = {https://openreview.net/forum?id=ohJxgRLlLt}
}
\clearpage
\appendix
\setcounter{topnumber}{3}\setcounter{bottomnumber}{2}\setcounter{totalnumber}{5}
\renewcommand{\topfraction}{0.92}\renewcommand{\bottomfraction}{0.75}
\renewcommand{\textfraction}{0.06}\renewcommand{\floatpagefraction}{0.80}
\section{Guide to the appendices}
\label{app:guide}

In Table~\ref{tab:guide}, we list each appendix with the part of the main text it supports. %

\begin{table}[htbp]
\centering\small
\caption{Appendix Guide.}
\label{tab:guide}
\begin{tabularx}{\linewidth}{@{}lXl@{}}
\toprule
Appendix & Contents & Supports\\
\midrule
\multicolumn{3}{@{}l}{\textit{Methods and inputs}}\\
\ref{app:diagnostics} & Diagnostic variants: \arm{Session bag} and the matched-evidence summaries; method key & Sections~\ref{sec:architectures}, \ref{sec:results}\\
\ref{app:counts} & Every method with the withheld traffic counts restored & Section~\ref{sec:architectures}\\
\ref{app:prompts} & System messages and rendered user messages & Section~\ref{sec:architectures}\\
\ref{app:ablations} & Discovery-pool ablations: summary statistics, wording, batch size, evidence depth, a mechanical summary & Section~\ref{sec:architectures}\\
\addlinespace[3pt]
\multicolumn{3}{@{}l}{\textit{Data and evaluation}}\\
\ref{app:protocol} & Pools, features, models, execution, coverage, rank convention, bootstrap, session sharing, observation window & Section~\ref{sec:design}\\
\ref{app:sensitivity} & Ordinal tie-breaking, Pearson correlation, unstandardized batch outputs & Sections~\ref{sec:introduction}, \ref{sec:design}\\
\addlinespace[3pt]
\multicolumn{3}{@{}l}{\textit{Results (Section~\ref{sec:comparative})}}\\
\ref{app:levels} & Rank correlation of every method, model and outcome & Figure~\ref{fig:models}\\
\ref{app:contrasts} & All 15 paired contrasts, by model and outcome, and \arm{Batch summary-80} against \arm{Session} & Section~\ref{sec:comparative}\\
\ref{app:batchedc} & Batching a product's own sessions into one call & Section~\ref{sec:comparative}\\
\ref{app:normavg} & Comparison, standardization and averaging in \arm{Batch summary}; one grouping against \arm{Session} & Section~\ref{sec:comparative}\\
\ref{app:did} & Difference in comparative gains between the representations & Section~\ref{sec:comparative}\\
\ref{app:depth} & \arm{Batch session} on 160 sessions per product & Figure~\ref{fig:batchrow}\\
\ref{app:supervised} & Supervised benchmarks and the single-feature rule: training pool, fitting, paired comparisons, training-size curve & Section~\ref{sec:comparative}\\
\ref{app:usage} & Calls, tokens, charges and latency by model and logged block & Section~\ref{sec:cost}\\
\addlinespace[3pt]
\multicolumn{3}{@{}l}{\textit{Interpretation (Section~\ref{sec:mechanism})}}\\
\ref{app:context} & Informativeness of single entries, effect of standardization, session methods at equal depth & Section~\ref{sec:mechanism}\\
\ref{app:mechanism} & Same-product and cross-product context & Section~\ref{sec:mechanism}\\
\bottomrule
\end{tabularx}
\end{table}

\section{Diagnostic methods and prediction granularity}
\label{app:diagnostics}

The diagnostic variants change one part of a main method at a time. \arm{Summary-80} and \arm{Batch summary-80} recompute the product statistics from the same 80 sessions used by \arm{Session} and \arm{Batch session}, so that summaries and sessions can be compared on the same evidence. \arm{Session bag} gives the model the same 80 records as \arm{Session} in one call and asks for a single product rate, so that the aggregation happens within the call rather than by averaging. \arm{Same-product batch} (Appendix~\ref{app:batchedc}) gives the model the same 80 records in one call but asks for a probability for each, which we average. Matching the evidence also removes information: with a fixed sample, \texttt{n\_sessions} equals 80 for every product and Coveo's \texttt{n\_search\_sessions} is rescaled to the same 80, so the matched summaries carry no traffic volume. This is one of the two reasons the reported summaries withhold the traffic counts (Appendix~\ref{app:counts}). Table~\ref{tab:arm-key} gives the stored result identifier for each method name and label.

\begin{table}[htbp]
\centering\small
\caption{Method names and source identifiers. C denotes the main session method throughout the paper; the source arm named C is the diagnostic R.}
\label{tab:arm-key}
\begin{tabularx}{\linewidth}{@{}ll lX@{}}
\toprule
Code & Method & Stored identifier & Role\\
\midrule
A & \arm{Summary} & \texttt{A} & Full-history summary; one product judgment\\
B & \arm{Batch summary} & \texttt{B} & Full-history summaries; comparative product judgments\\
C & \arm{Session} & \texttt{C\_sess} & One session per call; average the 80 probabilities\\
D & \arm{Batch session} & \texttt{D} & Cross-product sessions; average standardized judgments\\
\midrule
A80 & \arm{Summary-80} & \texttt{A80} & Isolated summary from the sampled 80 sessions\\
B80 & \arm{Batch summary-80} & \texttt{B80} & Comparative summaries from those 80 sessions\\
R & \arm{Session bag} & \texttt{C} & Same 80 records; one product judgment\\
--- & \arm{Same-product batch} & \texttt{solo\_vs\_batched} & Same 80 records in one call; average the 80 probabilities\\
\bottomrule
\end{tabularx}
\end{table}

\noindent{\bf Averaging session judgments or asking for one product rate.}
\arm{Session} and \arm{Session bag} receive the same sampled records of each product but differ in two
respects: what the model is asked to produce, 80 per-session probabilities that we average or a single
product rate, and how the records are batched, 80 calls or one. \arm{Session}
performs better, by \(0.131\;[0.0876,0.1704]\) for Coveo cart, \(0.167\;[0.1242,0.2054]\) for Coveo
purchase and \(0.173\;[0.1365,0.2084]\) for YooChoose purchase, above zero in 14 of the 15 settings,
with GPT on YooChoose the one null and none where \arm{Session bag} is higher. \arm{Same-product batch}, which batches the sessions as \arm{Session bag} does but asks for a probability for each, separates the two respects. Holding the batching fixed in this way, asking for per-session judgments makes no difference on Coveo, at
\(-0.013\;[-0.0475,0.0216]\) for cart and \(0.006\;[-0.0304,0.0406]\) for purchase, with individual
models split in both directions (per-session judgments are worse for Gemini on cart and better for GLM on purchase). On YooChoose it
helps, by \(0.070\;[0.0382,0.0997]\), above zero for Gemini, GLM and Qwen, the three models for which
\arm{Session bag} ranks lowest, and indistinguishable from zero for GPT and Sonnet. The remainder of the
\arm{Session} advantage comes from spreading the judgments across calls: \(0.144\) and \(0.161\) on
Coveo, essentially all of it, and \(0.104\) on YooChoose (Table~\ref{tab:batchedc}). Asking for one product rate instead of 80 averaged probabilities therefore lowers accuracy little beyond what batching the records into one call already does, except for three models on YooChoose.

\noindent{\bf Comparisons with matched evidence.}
\arm{Batch summary-80} improves on \arm{Summary-80} in all 15 model--outcome comparisons, with every paired interval above zero. Comparative processing therefore helps summaries even when their evidence is restricted to the sampled sessions. The mean \arm{Session}$-$\arm{Summary-80} differences are \(0.190\;[0.1467,0.2288]\) for Coveo cart, \(0.221\;[0.1784,0.2610]\) for Coveo purchase, and \(0.130\;[0.0864,0.1710]\) for YooChoose purchase. These comparisons hold sessions fixed, while the records and summaries still differ in fields, representation, and the number of judgments averaged.

\section{Summaries with the traffic counts}
\label{app:counts}

The reported summary block withholds a product's traffic counts: its number of sessions, and on
Coveo, its number of search sessions. They are withheld for two reasons. The session methods receive 80 sampled sessions per product and therefore cannot observe these counts, and the matched-evidence variants fix
these counts by construction (at 80 sessions for every product), so a full-history increment computed with
them present would mix the precision of the behavioral statistics with the appearance of a feature
that ranks products on its own. We ran both specifications on every model and dataset, changing
nothing else, and report them side by side in Table~\ref{tab:full-counts}. \arm{Session} and
\arm{Batch session} receive sessions rather than summary columns and are unaffected.

Restoring the counts lowers the isolated summary on Coveo, from \(0.081\) to \(0.048\) for cart and
from \(0.093\) to \(0.031\) for purchase, even though the counts (session count and search-session count) rank products at \(0.088\) and \(0.222\)
for cart and \(0.174\) and \(0.300\) for purchase on their own. It raises the isolated summary slightly
on YooChoose, from \(0.291\) to \(0.309\), where session count ranks at only \(0.034\). No ordering
among the methods changes on either dataset. Two YooChoose results are weaker with the counts present and lose their
separation from zero: the full-history increment \arm{Batch summary}$-$\arm{Batch summary-80}, at
\(0.020\) against \(0.041\), and \arm{Batch summary}$-$\arm{Batch session}, at \(0.027\) against
\(0.056\). Every other contrast keeps its sign and its significance.

\begin{table}[htbp]
\centering\small
\setlength{\tabcolsep}{4pt}
\caption{Summaries without and with the two traffic counts, mean of five models; the counts-free block is the reported specification. The session methods do not read the summary columns and are identical across blocks. A star marks a paired interval excluding zero.}
\label{tab:full-counts}
\begin{tabular}{@{}l cc cc cc@{}}
\toprule
& \multicolumn{2}{c}{Coveo cart} & \multicolumn{2}{c}{Coveo purchase} & \multicolumn{2}{c}{YooChoose purchase}\\
\cmidrule(lr){2-3}\cmidrule(lr){4-5}\cmidrule(lr){6-7}
& free & counts & free & counts & free & counts\\
\midrule
\multicolumn{7}{@{}l}{\emph{Arm levels}}\\
\arm{Summary} & 0.081 & 0.048 & 0.093 & 0.031 & 0.291 & 0.309\\
\arm{Batch summary} & 0.341 & 0.338 & 0.390 & 0.376 & 0.541 & 0.512\\
\arm{Summary-80} & 0.085 & 0.064 & 0.083 & 0.064 & 0.293 & 0.313\\
\arm{Batch summary-80} & 0.233 & 0.226 & 0.245 & 0.245 & 0.501 & 0.493\\
\addlinespace[3pt]
\multicolumn{7}{@{}l}{\emph{Paired contrasts}}\\
C$-$A & 0.194$^{*}$ & 0.227$^{*}$ & 0.211$^{*}$ & 0.273$^{*}$ & 0.132$^{*}$ & 0.114$^{*}$\\
C$-$A80 & 0.190$^{*}$ & 0.212$^{*}$ & 0.221$^{*}$ & 0.240$^{*}$ & 0.130$^{*}$ & 0.110$^{*}$\\
B$-$A & 0.260$^{*}$ & 0.289$^{*}$ & 0.297$^{*}$ & 0.345$^{*}$ & 0.250$^{*}$ & 0.203$^{*}$\\
B$-$C & 0.066$^{*}$ & 0.063$^{*}$ & 0.086$^{*}$ & 0.072$^{*}$ & 0.118$^{*}$ & 0.089$^{*}$\\
B$-$D & 0.110$^{*}$ & 0.107$^{*}$ & 0.126$^{*}$ & 0.112$^{*}$ & 0.056$^{*}$ & 0.027\\
B80$-$D & 0.002 & -0.005 & -0.019 & -0.019 & 0.015 & 0.007\\
B$-$B80 & 0.108$^{*}$ & 0.112$^{*}$ & 0.145$^{*}$ & 0.131$^{*}$ & 0.041$^{*}$ & 0.020\\
\bottomrule
\end{tabular}
\end{table}

\section{Prompts}
\label{app:prompts}
Every prompt is a system message that fixes the analyst role and the output schema, followed by a
user message that carries the evidence. Both are reproduced below, one method at a time, with the
system message verbatim and the user message as it was rendered for the first call of the reported
run. Long user messages are elided in the middle; the elision is marked and says what was removed.

The prompts are rendered by the same code that issued the calls, applied to the same pools, session samples, features and product groupings as the reported runs. Three details are visible in the renderings.
\arm{Batch session} (D) gives the model one session from each of 40 different products and does not
say which session belongs to which product, since each product contributes exactly one.
\arm{Session} (C) and \arm{Session bag} (R) read the same 80 sessions of one product and differ in how they are batched and in what they ask for: one session per call and a probability for it, against
all 80 in one call and a single product rate. Their system messages only differ in that one sentence. The summary methods list the same feature block in the same order, the 18 columns of
Appendix~\ref{app:protocol}, with absent values written \texttt{null}.

\noindent{\bf Summary (A).}
System message:
\begin{lstlisting}[style=prompt]
You are an e-commerce conversion analyst. You will be given aggregate behavioral statistics for one product. Estimate the fraction of browsing sessions containing this product that end in an add-to-cart, and the fraction that end in a purchase. Return ONLY a JSON object of the form {"cart_rate": <float>, "purchase_rate": <float>}. No prose.
\end{lstlisting}

User message, Coveo, from the reported run:
\begin{lstlisting}[style=prompt]
=== PRODUCT ===:
  price_bucket: 9
  avg_focal_dwell: 16.99
  median_focal_dwell: 0
  p25_focal_dwell: 0
  p75_focal_dwell: 16.9
  search_rate: 0
  avg_search_rank: null
  median_search_rank: null
  avg_comp_top1_dwell: 66.99
  avg_comp_top2_dwell: 30.59
  avg_comp_top3_dwell: 22.78
  median_comp_top1_dwell: 30
  avg_comp_top1_price: 8.16
  avg_comp_top2_price: 8.24
  avg_comp_top3_price: 8.16
  median_comp_top1_price: 9
  avg_n_competitors: 2.79
  median_n_competitors: 1

Predict cart_rate and purchase_rate for this product. JSON object only.
\end{lstlisting}

User message, YooChoose, from the reported run:
\begin{lstlisting}[style=prompt]
=== PRODUCT ===:
  price: 229
  avg_focal_dwell: 151.4
  median_focal_dwell: 43.1
  p25_focal_dwell: 0
  p75_focal_dwell: 120.2
  special_offer_rate: 0.6677
  category_click_rate: 0.3323
  brand_click_rate: 0
  avg_comp_top1_dwell: 383.6
  avg_comp_top2_dwell: 183
  avg_comp_top3_dwell: 146.3
  median_comp_top1_dwell: 179.9
  avg_comp_top1_price: 2976
  avg_comp_top2_price: 891.7
  avg_comp_top3_price: 1250
  median_comp_top1_price: 329
  avg_n_competitors: 4.486
  median_n_competitors: 2

Predict cart_rate and purchase_rate for this product. JSON object only.
\end{lstlisting}

\noindent{\bf Batch summary (B).}
System message:
\begin{lstlisting}[style=prompt]
You are an e-commerce conversion analyst. You will be given aggregate behavioral statistics for several products. For each, estimate the fraction of browsing sessions containing it that end in an add-to-cart, and the fraction that end in a purchase. Return ONLY a JSON array, one object per product, of the form {"product": <int>, "cart_rate": <float>, "purchase_rate": <float>}. No prose.
\end{lstlisting}

User message, Coveo, from the reported run:
\begin{lstlisting}[style=prompt]
=== PRODUCT 0 ===:
  price_bucket: 9
  avg_focal_dwell: 16.99
  median_focal_dwell: 0
  p25_focal_dwell: 0
  p75_focal_dwell: 16.9
  search_rate: 0
  avg_search_rank: null
  median_search_rank: null
  avg_comp_top1_dwell: 66.99
  avg_comp_top2_dwell: 30.59
  avg_comp_top3_dwell: 22.78
  median_comp_top1_dwell: 30
  avg_comp_top1_price: 8.16
  avg_comp_top2_price: 8.24
  avg_comp_top3_price: 8.16
  median_comp_top1_price: 9
  avg_n_competitors: 2.79
  median_n_competitors: 1

=== PRODUCT 1 ===:

[... the remaining 39 product blocks, in the same format ...]


Predict cart_rate and purchase_rate for each product. JSON array only.
\end{lstlisting}

User message, YooChoose, from the reported run:
\begin{lstlisting}[style=prompt]
=== PRODUCT 0 ===:
  price: 229
  avg_focal_dwell: 151.4
  median_focal_dwell: 43.1
  p25_focal_dwell: 0
  p75_focal_dwell: 120.2
  special_offer_rate: 0.6677
  category_click_rate: 0.3323
  brand_click_rate: 0
  avg_comp_top1_dwell: 383.6
  avg_comp_top2_dwell: 183
  avg_comp_top3_dwell: 146.3
  median_comp_top1_dwell: 179.9
  avg_comp_top1_price: 2976
  avg_comp_top2_price: 891.7
  avg_comp_top3_price: 1250
  median_comp_top1_price: 329
  avg_n_competitors: 4.486
  median_n_competitors: 2

=== PRODUCT 1 ===:

[... the remaining 39 product blocks, in the same format ...]


Predict cart_rate and purchase_rate for each product. JSON array only.
\end{lstlisting}

\noindent{\bf Session (C).}
System message:
\begin{lstlisting}[style=prompt]
You are an e-commerce conversion analyst. You will be given browsing sessions for a product. For each session, estimate the probability that the user adds that session's focal product to their cart and the probability that they purchase it. Return ONLY a JSON array, one object per session, of the form {"session": <int>, "p_add_cart": <float>, "p_purchase": <float>}. No prose.
\end{lstlisting}

User message, Coveo, from the reported run:
\begin{lstlisting}[style=prompt]
Session 0: focal_price=9.0, time_spent_viewing=0.0s, number_of_views=1, discovery_order=2, discovered_via_search=False, session_length=958s, total_searches_in_session=1, num_competing_products=7, top_competing_products=[(price=8.0, time_spent_viewing=107.4s, discovery_order=4), (price=8.0, time_spent_viewing=49.0s, discovery_order=5), (price=9.0, time_spent_viewing=46.9s, discovery_order=1)]

Predict p_add_cart and p_purchase for each session. JSON array only.
\end{lstlisting}

User message, YooChoose, from the reported run:
\begin{lstlisting}[style=prompt]
Session 0: focal_price=229, time_spent_viewing=0.0s, number_of_views=1, discovery_order=2, click_context=special_offer, session_length=36s, num_competing_products=1, top_competing_products=[(price=261, time_spent_viewing=36.6s, discovery_order=1)]

Predict p_add_cart and p_purchase for each session. JSON array only.
\end{lstlisting}

\noindent{\bf Batch session (D).}
System message:
\begin{lstlisting}[style=prompt]
You are an e-commerce conversion analyst. You will be given browsing sessions drawn from SEVERAL DIFFERENT products. For each session, estimate the probability that the user adds that session's focal product to their cart and the probability that they purchase it. Return ONLY a JSON array, one object per session, of the form {"session": <int>, "p_add_cart": <float>, "p_purchase": <float>}. No prose.
\end{lstlisting}

User message, Coveo, from the reported run:
\begin{lstlisting}[style=prompt]
Session 0: focal_price=4.0, time_spent_viewing=0.0s, number_of_views=1, discovery_order=1, discovered_via_search=False, session_length=0s, total_searches_in_session=0, num_competing_products=0, top_competing_products=[none]
Session 1: focal_price=9.0, time_spent_viewing=0.0s, number_of_views=2, discovery_order=1, discovered_via_search=False, session_length=365s, total_searches_in_session=0, num_competing_products=2, top_competing_products=[(price=9.0, time_spent_viewing=0.0s, discovery_order=2), (price=10.0, time_spent_viewing=0.0s, discovery_order=3)]
Session 2: focal_price=6.0, time_spent_viewing=0.0s, number_of_views=1, discovery_order=5, discovered_via_search=False, session_length=128s, total_searches_in_session=0, num_competing_products=5, top_competing_products=[(price=4.0, time_spent_viewing=22.0s, discovery_order=2), (price=4.0, time_spent_viewing=0.0s, discovery_order=1), (price=7.0, time_spent_viewing=0.0s, discovery_order=3)]

[... sessions 3 to 38, each the single session of a different product ...]

Session 39: focal_price=8.0, time_spent_viewing=0.0s, number_of_views=1, discovery_order=1, discovered_via_search=False, session_length=1028s, total_searches_in_session=1, num_competing_products=1, top_competing_products=[(price=5.0, time_spent_viewing=31.6s, discovery_order=2)]

Predict p_add_cart and p_purchase for each session. JSON array only.
\end{lstlisting}

User message, YooChoose, from the reported run:
\begin{lstlisting}[style=prompt]
Session 0: focal_price=3978, time_spent_viewing=49.4s, number_of_views=1, discovery_order=2, click_context=category_1, session_length=119s, num_competing_products=2, top_competing_products=[(price=3664, time_spent_viewing=70.1s, discovery_order=1), (price=null, time_spent_viewing=0.0s, discovery_order=3)]
Session 1: focal_price=22515, time_spent_viewing=114.7s, number_of_views=1, discovery_order=3, click_context=category_7, session_length=3481s, num_competing_products=6, top_competing_products=[(price=null, time_spent_viewing=2919.6s, discovery_order=6), (price=25028, time_spent_viewing=309.9s, discovery_order=2), (price=26075, time_spent_viewing=72.0s, discovery_order=1)]
Session 2: focal_price=2408, time_spent_viewing=35.3s, number_of_views=1, discovery_order=4, click_context=category_4, session_length=474s, num_competing_products=5, top_competing_products=[(price=2408, time_spent_viewing=289.3s, discovery_order=3), (price=null, time_spent_viewing=87.3s, discovery_order=2), (price=6282, time_spent_viewing=41.4s, discovery_order=1)]

[... sessions 3 to 38, each the single session of a different product ...]

Session 39: focal_price=1046, time_spent_viewing=12.2s, number_of_views=2, discovery_order=3, click_context=category_3, session_length=2981s, num_competing_products=7, top_competing_products=[(price=1105, time_spent_viewing=2188.9s, discovery_order=4), (price=null, time_spent_viewing=679.1s, discovery_order=2), (price=null, time_spent_viewing=53.1s, discovery_order=1)]

Predict p_add_cart and p_purchase for each session. JSON array only.
\end{lstlisting}

\noindent{\bf Session bag (R).}
System message:
\begin{lstlisting}[style=prompt]
You are an e-commerce conversion analyst. For each product below you are shown a sample of the browsing sessions in which shoppers viewed it. Using those sessions, estimate for each product the fraction of its sessions that end in an add-to-cart, and the fraction that end in a purchase. Return ONLY a JSON array, one object per product, of the form {"product": <int>, "cart_rate": <float>, "purchase_rate": <float>}. No prose.
\end{lstlisting}

User message, Coveo, from the reported run:
\begin{lstlisting}[style=prompt]
=== PRODUCT 0 ===  (80 sessions)
Session 0: focal_price=9.0, time_spent_viewing=0.0s, number_of_views=1, discovery_order=2, discovered_via_search=False, session_length=958s, total_searches_in_session=1, num_competing_products=7, top_competing_products=[(price=8.0, time_spent_viewing=107.4s, discovery_order=4), (price=8.0, time_spent_viewing=49.0s, discovery_order=5), (price=9.0, time_spent_viewing=46.9s, discovery_order=1)]
Session 1: focal_price=9.0, time_spent_viewing=5.8s, number_of_views=3, discovery_order=1, discovered_via_search=False, session_length=22s, total_searches_in_session=0, num_competing_products=0, top_competing_products=[none]
Session 2: focal_price=9.0, time_spent_viewing=68.6s, number_of_views=1, discovery_order=1, discovered_via_search=False, session_length=582s, total_searches_in_session=0, num_competing_products=1, top_competing_products=[(price=8.0, time_spent_viewing=0.0s, discovery_order=2)]

[... sessions 3 to 78 of the same product, in the same format ...]

Session 79: focal_price=9.0, time_spent_viewing=0.0s, number_of_views=1, discovery_order=4, discovered_via_search=False, session_length=269s, total_searches_in_session=0, num_competing_products=5, top_competing_products=[(price=9.0, time_spent_viewing=0.0s, discovery_order=1), (price=9.0, time_spent_viewing=0.0s, discovery_order=2), (price=9.0, time_spent_viewing=0.0s, discovery_order=3)]

Predict cart_rate and purchase_rate for each product. JSON array only.
\end{lstlisting}

User message, YooChoose, from the reported run:
\begin{lstlisting}[style=prompt]
=== PRODUCT 0 ===  (80 sessions)
Session 0: focal_price=229, time_spent_viewing=0.0s, number_of_views=1, discovery_order=2, click_context=special_offer, session_length=36s, num_competing_products=1, top_competing_products=[(price=261, time_spent_viewing=36.6s, discovery_order=1)]
Session 1: focal_price=229, time_spent_viewing=81.3s, number_of_views=1, discovery_order=5, click_context=special_offer, session_length=1124s, num_competing_products=12, top_competing_products=[(price=261, time_spent_viewing=166.4s, discovery_order=6), (price=554, time_spent_viewing=146.3s, discovery_order=7), (price=null, time_spent_viewing=135.9s, discovery_order=11)]
Session 2: focal_price=229, time_spent_viewing=0.0s, number_of_views=1, discovery_order=2, click_context=special_offer, session_length=391s, num_competing_products=1, top_competing_products=[(price=null, time_spent_viewing=391.7s, discovery_order=1)]

[... sessions 3 to 78 of the same product, in the same format ...]

Session 79: focal_price=229, time_spent_viewing=0.0s, number_of_views=1, discovery_order=2, click_context=category_10, session_length=18s, num_competing_products=1, top_competing_products=[(price=345, time_spent_viewing=18.5s, discovery_order=1)]

Predict cart_rate and purchase_rate for each product. JSON array only.
\end{lstlisting}

\clearpage
\section{Discovery-pool design ablations}
\label{app:ablations}

The design choices behind Figure~\ref{fig:designs} -- which statistics the summary contains, how
many sessions it is computed from, how many products share a prompt, and how the request is
worded -- were fixed on the discovery pool before any interaction with the evaluation pools. Here we vary each
choice in turn on that same discovery pool (280 Coveo
products) with GPT-5.6, so the evaluation pools remain untouched by anything exploratory.

Every cell is \arm{Batch summary} and differs from a reference cell in exactly one respect.
The reference is the curated block at batch 40 with the reported prompt, re-run through the
same program as the ablations rather than taken from the earlier discovery runs, whose
records use a different format; all contrasts below are therefore internal to one code path.
Every cell uses the reported 18-column summary, with the two traffic counts withheld as in the evaluation runs. Running the same grid with the counts present leaves the depth and wording results unchanged; the only difference is that the smallest batch sizes then fall below the reference, at \(-0.066\) and \(-0.060\) for batch 5, as does batch 280 on cart, at \(-0.044\). Reference correlations are \(0.307\) for cart and \(0.303\) for purchase. Intervals are
paired product bootstraps with a shared index set.

\noindent{\bf Summary statistics.}
The columns are not a selected subset. They are the complete set of non-outcome,
non-identifier columns produced by the feature builder: a price bucket, traffic counts, four
focal-dwell statistics, three search statistics, and the dwell, price and count statistics of
the three most-considered competing products, of which the reported summary renders all but
the traffic counts (Appendix~\ref{app:counts}). The construction crosses each behavioral
quantity with conventional descriptive statistics -- mean, median, quartiles, counts and
rates -- and keeps every result. Six columns are excluded, five because they are derived
from the outcome (\texttt{n\_add\_to\_cart}, \texttt{n\_purchased}, \texttt{n\_displaced},
\texttt{n\_competitor\_buy}, \texttt{competitor\_buy\_rate}) and would leak the target, and
one because it is a category identifier rather than a statistic. No column was added or
removed on the basis of predictive performance, and the feature file predates the
experiments reported here.

\noindent{\bf Prompt wording.}
Two paraphrases carrying identical information and an identical output contract change the
correlation by \(-0.009\) and \(+0.004\) on cart and \(-0.003\) and \(+0.000\) on purchase,
every interval containing zero. The reported results are not an artifact of one phrasing. All reported runs use the first wording, which is the one
reproduced in Appendix~\ref{app:prompts}; the two paraphrases are available from the authors on request.

\noindent{\bf Batch size.}
Holding the number of groupings at five, so each product receives five judgments at every
size, correlation is flat in batch size (Table~\ref{tab:abl-batch}). No size from 5 to 280 differs
from the reference by a margin the paired intervals distinguish from zero, on either outcome;
the largest movement is \(+0.024\) at batch 140 on cart. Batch 40 therefore
sits inside a flat region rather than at a favorable extreme, and the inference-cost
comparison in Section~\ref{sec:results} does not depend on the choice. Three features of this
curve deserve care. Batch size sets both the number of products compared and the number of
points entering the within-call standardization, so at batch 5 the score is divided by the
spread of five values. At the other end a single call holds the entire discovery pool, so the
five groupings at batch 280 repeat one comparison set instead of varying it, and averaging over
comparison contexts disappears. And batch 1 would be \arm{Summary} exactly, so the curve traces
a continuous path between the isolated and comparative methods rather than treating batching as
a switch.

\noindent{\bf Evidence depth.}
Recomputing the same statistics from the first \(K\) sampled sessions shows how accuracy depends
on the number of sessions a summary draws on (Table~\ref{tab:abl-depth}). Correlation rises with depth, apart from a small dip between 10 and 20 sessions on both outcomes,
at a constant 35 calls per cell, which is the property that makes aggregation attractive:
additional evidence enters the prediction without additional inference. Both outcomes are still rising at 240 sessions, the deepest sample we hold, and the cells
below 80 all sit below the reference. Products retain every session they
have, so the realized depth at a cap of 240 averages 202: 280 products reach 80, 250 reach
120, 203 reach 160 and 155 reach 240.

\noindent{\bf An exhaustive mechanical summary.}
Replacing the curated block with a mechanical one -- count, mean, standard deviation,
minimum, median and maximum of every numeric field in the session record, with no selection
of any kind -- changes the correlation by \(+0.023\;[-0.0823,+0.1332]\) on cart and
\(+0.007\;[-0.0987,+0.1157]\) on purchase. The mechanical block carries 144 columns against
the reported block's 18 and involves no domain judgment at all, and we detect no difference from
the curated block. The intervals are wide, about \(\pm0.1\), so this does not establish that the two
are equivalent; it shows only that the result does not depend on the curated statistics by a
margin this pool can detect.

\begin{table}[htbp]
\centering\small
\caption{Batch size, discovery pool, GPT-5.6. Every size uses five groupings, so each product receives five judgments; at 280, one call holds the whole pool. A star marks a paired interval excluding zero.}
\label{tab:abl-batch}
\begin{tabular}{@{}lccc@{}}
\toprule
Products per prompt & Calls & Cart & Purchase\\
\midrule
5 & 280 & 0.280 & 0.276\\
10 & 140 & 0.297 & 0.283\\
20 & 70 & 0.288 & 0.280\\
40 (reference) & 35 & 0.307 & 0.303\\
80 & 20 & 0.324 & 0.305\\
140 & 10 & 0.331 & 0.294\\
280 & 5 & 0.317 & 0.311\\
\bottomrule
\end{tabular}
\end{table}

\begin{table}[htbp]
\centering\small
\caption{Evidence depth, discovery pool, GPT-5.6: statistics recomputed from the first $K$ sampled sessions, 35 calls per cell. Products with fewer sessions keep all of them, so the realized mean depth is shown. A star marks a paired interval excluding zero.}
\label{tab:abl-depth}
\begin{tabular}{@{}lccc@{}}
\toprule
Sessions (cap / realized mean) & Products reaching cap & Cart & Purchase\\
\midrule
5 / 5 & 280 & 0.057$^{*}$ & 0.023$^{*}$\\
10 / 10 & 280 & 0.150$^{*}$ & 0.127$^{*}$\\
20 / 20 & 280 & 0.132$^{*}$ & 0.119$^{*}$\\
40 / 40 & 280 & 0.203$^{*}$ & 0.165$^{*}$\\
80 / 80 (reference) & 280 & 0.307 & 0.303\\
120 / 119 & 250 & 0.371 & 0.330\\
160 / 151 & 203 & 0.453$^{*}$ & 0.379\\
240 / 202 & 155 & 0.502$^{*}$ & 0.410$^{*}$\\
\bottomrule
\end{tabular}
\end{table}

\section{Protocol and reproducibility}
\label{app:protocol}
Here we record what a reader would need in order to rerun the evaluation: how the pools were drawn, what the models were shown, how the runs were executed, and how the reported statistics are computed.

\noindent{\bf Product pools and sampling.}
The Coveo pool contains 1,000 distinct products, selected with at least 100 sessions and stratified by depth decile and price-bucket tercile across 30 cells. It is disjoint from the discovery pools used to develop the procedures, and its behavioral inputs and outcome rates describe the same observation period. We sample 80 sessions per product by reservoir sampling during a pass through the trace. Each record includes up to three co-considered products chosen by dwell time. A session containing two evaluated products can contribute to both histories.

For YooChoose, the 1,000-product pool requires at least 300 evaluation-window sessions and a positive-price purchase before August 1, 2014. Historical purchases supply prices; evaluation-window sessions supply behavior and target rates. The summary and session methods use these same information sources. The selection excludes products with no priced sales history. Competitor price is missing when the competitor has no priced purchase in the earlier window, even though focal price is present for every selected product.

The realized pools and prompt designs define the evaluations. Coveo's pool artifact stores seed 2327118 and its session-sample artifact stores 1422810. Their original seed expression uses a process-dependent Python string hash, so rerunning that expression can change the selected products or sessions. Reproduction requires the stored artifacts. The YooChoose construction uses the explicit seed 20260911. Coveo discovery results inform the procedures subsequently evaluated on separate products and on YooChoose.

\noindent{\bf Behavioral features.}
Coveo's 20 columns comprise price bucket and traffic; four focal-dwell statistics; search count, search share, and two search-rank statistics; four competitor-dwell statistics; four competitor-price statistics; and two competitor-count statistics. YooChoose has 19 columns: price and traffic; the same four focal-dwell statistics; special-offer rate, category-click rate, and brand-click rate; and the same competitor-dwell, competitor-price, and competitor-count groups. The two blocks reflect the fields available in each dataset. The reported summary prompts render all of these except the traffic counts, 18 columns on each dataset; we report the arms with them restored in Appendix~\ref{app:counts}, and the supervised benchmark uses the full block.

Session inputs retain individual records rather than these marginal summaries. The Coveo renderer additionally exposes view counts and consideration order. Coveo summary features use four significant digits and render missing values as \texttt{null}. Outcome fields are masked in the prompted records. Recomputing A80 and B80 updates the summary features from the fixed 80-session sample, so comparisons with the full history change the behavioral statistics alone, the traffic counts being withheld from every summary prompt.

\noindent{\bf Models and execution.}
The prompts are reproduced in Appendix~\ref{app:prompts}. We use the endpoints in Table~\ref{tab:models}, with temperature zero and low reasoning effort. Qwen uses an eightfold token-budget scale. The execution records span September 9--20, 2026 for Coveo and September 11--20 for YooChoose. The endpoints are provider slugs rather than immutable model snapshots; we preserve endpoint identifiers and execution dates as the version record.

\begin{table}[htbp]
\centering\small
\caption{Model endpoints used across the two datasets.}
\label{tab:models}
\begin{tabular}{@{}ll@{}}
\toprule
Model & Endpoint\\
\midrule
GPT-5.6 & \texttt{openai/gpt-5.6}\\
Qwen3.8 & \texttt{qwen/qwen3.8-2.4t-a95b}\\
Gemini3.8 Flash & \texttt{google/gemini-3.8-flash}\\
GLM5.3 & \texttt{z-ai/glm-5.3}\\
Claude Sonnet5 & \texttt{anthropic/claude-sonnet-5}\\
\bottomrule
\end{tabular}
\end{table}

Each call is attempted up to six times, doubling the token limit from a per-cell base up to 32,000 tokens. An empty parse triggers a retry; a call exhausting retries contributes no checkpoint entry. Because the sampling temperature is zero, a reply that the parser rejects for its shape rather than its content tends to recur on retry, so such a call is usually lost rather than recovered. Two response parsers in the execution code were strict in this way: the product-level parser required a JSON array and rejected an otherwise complete single-object reply, and the session parser discarded a response whose enclosing array was truncated instead of keeping the complete records inside it. We repaired both parsers to accept any complete record, to log rejected replies, and to report a call that yields nothing, then re-ran the affected calls: 38 single-product session calls and 22 cross-product session calls across three models and both datasets. Five product-level predictions rejected only for their shape were recovered from the logged replies rather than re-requested, since the model had already answered them at temperature zero. Every method now covers the full 1,000-product pool for every model on both datasets (Table~\ref{tab:full-coverage}). Comparative methods use 40 products per call. Five summary groupings produce 125 planned calls; 80 cross-product session rounds produce 2,000. Isolated summaries and session bags each require 1,000 planned calls, and the per-session arm 80,000. The coverage gate is 97\%; in cross-product session inference, calls with fewer than 80\% of their 40 session predictions parsed are dropped.

\noindent{\bf Common support.}
We use one model-specific intersection of products across all seven methods for every arm level and paired contrast. We report those counts in Table~\ref{tab:full-coverage}. Differences between a model's methods therefore use the same products. After the refills described above, every model's support is the full 1,000-product pool on both datasets.
\begin{table}[htbp]
\centering\footnotesize
\setlength{\tabcolsep}{2.5pt}
\caption{Prediction coverage: product calls for A, A80 and R (expected 1,000), 40-product calls for B and B80 (125), and parsed session predictions for C and D (80,000). $n$ is the common support for all seven arms; Coveo coverage is the same for both outcomes.}
\label{tab:full-coverage}
\begin{tabularx}{\linewidth}{@{}l*{8}{>{\raggedleft\arraybackslash}X}@{}}
\toprule
Model & $n$ & A & C & D & B & A80 & B80 & R\\
\midrule
\multicolumn{9}{@{}l}{\textit{Coveo}}\\
\addlinespace[2pt]
GPT-5.6 & 1000 & 1,000 & 80,000 & 80,000 & 125 & 1,000 & 125 & 1,000\\
Qwen3.8 & 1000 & 1,000 & 80,000 & 80,000 & 125 & 1,000 & 125 & 1,000\\
Gemini3.8 & 1000 & 1,000 & 80,000 & 80,000 & 125 & 1,000 & 125 & 1,000\\
GLM5.3 & 1000 & 1,000 & 80,000 & 80,000 & 125 & 1,000 & 125 & 1,000\\
Sonnet5 & 1000 & 1,000 & 80,000 & 80,000 & 125 & 1,000 & 125 & 1,000\\
\addlinespace[5pt]
\multicolumn{9}{@{}l}{\textit{YooChoose}}\\
\addlinespace[2pt]
GPT-5.6 & 1000 & 1,000 & 80,000 & 80,000 & 125 & 1,000 & 125 & 1,000\\
Qwen3.8 & 1000 & 1,000 & 80,000 & 80,000 & 125 & 1,000 & 125 & 1,000\\
Gemini3.8 & 1000 & 1,000 & 80,000 & 80,000 & 125 & 1,000 & 125 & 1,000\\
GLM5.3 & 1000 & 1,000 & 80,000 & 80,000 & 125 & 1,000 & 125 & 1,000\\
Sonnet5 & 1000 & 1,000 & 80,000 & 79,999 & 125 & 1,000 & 125 & 1,000\\
\addlinespace[5pt]
\bottomrule
\end{tabularx}
\end{table}

\noindent{\bf Aggregation and rank convention.}
For B, B80, and D, we standardize within each call using the population standard deviation, replacing a divisor at or below $10^{-12}$ with one, and average by product across calls. A, A80, and R each produce one product prediction. C averages the raw probabilities returned by its 80 single-session calls; with one output per call there is nothing within a call to standardize against.

For a prediction vector $s$ and observed rates $\bar Y$, we report Spearman's correlation \(\rho=\Corr(R(s),R(\bar Y))\), where \(R\) assigns tied values the average of the ranks they span (\texttt{scipy.stats.rankdata} with \texttt{method="average"}). The same convention is used for every method, contrast, supervised benchmark and the single-feature reference. In Appendix~\ref{app:sensitivity}, we report the effect of breaking ties by sort order instead.

\noindent{\bf Paired bootstrap.}
Within each dataset, we use 2,000 product resamples from NumPy's random generator with seed seven. The index set is drawn once, before any model, method, or outcome is processed, and resamples product positions in the shared pool ordering, so every level, contrast, and difference of differences in the paper uses the same resampled products. A model's common-support mask is applied to its prediction vector before resampling; a resampled position outside that model's support enters as a missing value and is dropped pairwise inside the correlation. Replicates are therefore aligned by product identity across models; with every support at the full pool, no product is dropped. We compute paired differences within each replicate and report the 2.5th and 97.5th percentiles. For model means, we average the model-specific values within each replicate before taking percentiles. The intervals condition on saved predictions.

\noindent{\bf Session sharing between products.}
Resampling products treats them as independent, which shared shopping sessions could violate. The stored session samples record behavior without session identifiers, so we measure sharing in the released logs directly. Sharing is common at the population level: 11.5\% of Coveo sessions touching a pooled product touch more than one, giving 27,327 of the 499,500 product pairs at least one shared session; on YooChoose the figures are 24.4\% and 108,753 pairs.

For the session methods, that statistic overstates the dependence that reaches the estimates. Each pooled product appears in hundreds of logged sessions, a median of 279 on Coveo and 643 on YooChoose, but contributes only 80 to the experiment, and a sharing pair shares a median of one logged session. Scaling each pair's shared count by the probability that a shared session enters both 80-session samples gives the expected overlap within the evaluated evidence: a mean of \(0.131\) shared sessions per sharing pair on Coveo and \(0.039\) on YooChoose, with 518 and 750 of the 499,500 pairs, about one in a thousand, expected to share even one sampled session. Averaged over all pairs this is roughly \(0.01\%\) of a product's 80 sessions per pair; summed over its partners, a product's 80 sampled sessions are expected to overlap other products' samples about 7 times on Coveo and 8 on YooChoose, spread thinly across many pairs. The argument does not extend to the full histories behind the summary methods or to the observed rates, which draw on all logged sessions and share them more often; the product bootstrap does not account for that dependence, and we report it without a cluster correction.

\noindent{\bf Further limitations.}
The summary is one particular compression, though on the discovery pool we detect no difference from an exhaustive block of 144 mechanical summary columns, within wide intervals (Appendix~\ref{app:ablations}). We measure ranking alone and leave calibration and cardinal accuracy aside. Our intervals condition on the saved predictions and treat products as independent, although full histories share sessions across products, as discussed above, and they do not include variation from drawing new sessions or new comparative groupings.

\noindent{\bf Observation window of the session records.}
We remove the focal cart and purchase fields from every prompted record. On Coveo, the remaining fields describe completed sessions: session length runs from the first to the last logged event, including cart and purchase events, and a product's viewing time runs from a detail view to the next event of any kind, so both can include time after the focal outcome. Session length appears only in the session records, while viewing-time statistics enter both the records and the summaries, so the methods differ in some fields under this regime; we do not measure how that affects their ordering. On YooChoose, where purchases are logged separately from clicks, our preprocessing truncates each session at the time of its first purchase, so its records contain no activity after a purchase: session length runs to the last click before that time, and that last click contributes no viewing time. On both datasets the task is a retrospective reconstruction of observed rates, not a forecast made before the outcome. YooChoose prices come from purchases before the August 2014 split for both focal and competing products.

\section{Metric sensitivity}
\label{app:sensitivity}
\label{app:pearson}
\label{app:metrics}

We recompute the results under three alternatives to the reported estimator, holding the predictions, the common supports, and the bootstrap index set fixed. The alternatives are ordinal ranks that break ties by sort order rather than averaging them, Pearson correlation, and Spearman correlation applied to batch outputs that have not been standardized within the call. We report the two headline contrasts under each in Table~\ref{tab:sensitivity}.

\noindent{\bf Rank convention.}
We rebuilt both full reports with ties broken by sort order (\texttt{argsort} applied twice) and compared every arm level and all 15 paired contrasts for every model and outcome. The largest shift is \(0.022\) in an arm level and \(0.030\) in a paired contrast. Of the 225 paired intervals, 2 change significance, one in each direction, and no significant result reverses sign: \arm{Session bag}$-$\arm{Summary-80} for Gemini on Coveo purchase is significant under midranks but not under ordinal ranks, and \arm{Session}$-$\arm{Summary-80} for Sonnet on YooChoose, \(0.055\), is significant under ordinal ranks but not under midranks. For the two headline contrasts in Table~\ref{tab:sensitivity} the largest shift is \(0.010\), with no change in significance.

\noindent{\bf Pearson correlation.}
The observed rates are strongly right-skewed, with skewness \(+2.16\) and \(+2.46\) on the two Coveo outcomes and \(+1.85\) on YooChoose, and 30.2\% of Coveo purchase rates are exactly zero. Pearson correlation therefore weights a small number of high-conversion products heavily, and it measures a linear relationship the methods are not designed to deliver: the batch methods standardize within each call, which removes level information by construction. We report it as a sensitivity analysis rather than as a second primary metric.

Both findings replicate. All 15 \arm{Batch summary}$-$\arm{Summary} intervals remain above zero, and \arm{Batch session}$-$\arm{Session} remains weak, with two intervals above zero under Pearson against three under the reported estimator. 

\noindent{\bf Unstandardized batch outputs.}
Averaging raw batch outputs instead of within-call standardized scores preserves all 30 conclusions: every \arm{Batch summary}$-$\arm{Summary} interval is still above zero, and \arm{Batch session}$-$\arm{Session} is still above zero in three settings. Standardization sharpens the comparative gain rather than creating it; we separate the two in Appendix~\ref{app:normavg}.

\begin{table}[htbp]
\centering\small
\caption{The two headline contrasts under four estimators: the reported Spearman correlation (midranks) on standardized outputs, ordinal ranks with ties broken by sort order, Pearson correlation, and Spearman correlation on unstandardized outputs. A star marks a paired 95\% interval excluding zero.}
\label{tab:sensitivity}
\begin{tabular}{@{}lcccc@{}}
\toprule
Model & Spearman & Ordinal ranks & Pearson & Unstandardized\\
\midrule
\multicolumn{5}{@{}l}{\emph{\arm{Batch summary}$-$\arm{Summary}, Coveo cart}}\\
GPT-5.6 & 0.287$^{*}$ & 0.289$^{*}$ & 0.234$^{*}$ & 0.236$^{*}$ \\
Sonnet5 & 0.228$^{*}$ & 0.228$^{*}$ & 0.173$^{*}$ & 0.211$^{*}$ \\
Gemini3.8 & 0.181$^{*}$ & 0.186$^{*}$ & 0.152$^{*}$ & 0.141$^{*}$ \\
GLM5.3 & 0.258$^{*}$ & 0.255$^{*}$ & 0.160$^{*}$ & 0.208$^{*}$ \\
Qwen3.8 & 0.349$^{*}$ & 0.354$^{*}$ & 0.318$^{*}$ & 0.313$^{*}$ \\
\addlinespace[3pt]
\multicolumn{5}{@{}l}{\emph{\arm{Batch summary}$-$\arm{Summary}, Coveo purchase}}\\
GPT-5.6 & 0.310$^{*}$ & 0.305$^{*}$ & 0.235$^{*}$ & 0.278$^{*}$ \\
Sonnet5 & 0.273$^{*}$ & 0.264$^{*}$ & 0.214$^{*}$ & 0.258$^{*}$ \\
Gemini3.8 & 0.201$^{*}$ & 0.198$^{*}$ & 0.152$^{*}$ & 0.165$^{*}$ \\
GLM5.3 & 0.236$^{*}$ & 0.232$^{*}$ & 0.190$^{*}$ & 0.190$^{*}$ \\
Qwen3.8 & 0.465$^{*}$ & 0.459$^{*}$ & 0.405$^{*}$ & 0.414$^{*}$ \\
\addlinespace[3pt]
\multicolumn{5}{@{}l}{\emph{\arm{Batch summary}$-$\arm{Summary}, YooChoose purchase}}\\
GPT-5.6 & 0.183$^{*}$ & 0.186$^{*}$ & 0.201$^{*}$ & 0.179$^{*}$ \\
Sonnet5 & 0.215$^{*}$ & 0.219$^{*}$ & 0.175$^{*}$ & 0.192$^{*}$ \\
Gemini3.8 & 0.251$^{*}$ & 0.261$^{*}$ & 0.271$^{*}$ & 0.092$^{*}$ \\
GLM5.3 & 0.269$^{*}$ & 0.268$^{*}$ & 0.248$^{*}$ & 0.201$^{*}$ \\
Qwen3.8 & 0.333$^{*}$ & 0.339$^{*}$ & 0.270$^{*}$ & 0.320$^{*}$ \\
\midrule
\multicolumn{5}{@{}l}{\emph{\arm{Batch session}$-$\arm{Session}, Coveo cart}}\\
GPT-5.6 & 0.006 & 0.006 & 0.018 & 0.001 \\
Sonnet5 & -0.122$^{*}$ & -0.122$^{*}$ & -0.144$^{*}$ & -0.119$^{*}$ \\
Gemini3.8 & -0.021 & -0.020 & -0.027$^{*}$ & -0.020 \\
GLM5.3 & -0.083$^{*}$ & -0.083$^{*}$ & -0.092$^{*}$ & -0.066$^{*}$ \\
Qwen3.8 & 0.000 & 0.000 & -0.013 & 0.017 \\
\addlinespace[3pt]
\multicolumn{5}{@{}l}{\emph{\arm{Batch session}$-$\arm{Session}, Coveo purchase}}\\
GPT-5.6 & -0.007 & -0.006 & -0.008 & -0.023 \\
Sonnet5 & -0.096$^{*}$ & -0.095$^{*}$ & -0.083$^{*}$ & -0.087$^{*}$ \\
Gemini3.8 & -0.027$^{*}$ & -0.026$^{*}$ & -0.039$^{*}$ & -0.024$^{*}$ \\
GLM5.3 & -0.051$^{*}$ & -0.050$^{*}$ & -0.053$^{*}$ & -0.032 \\
Qwen3.8 & -0.020 & -0.020 & -0.033$^{*}$ & -0.008 \\
\addlinespace[3pt]
\multicolumn{5}{@{}l}{\emph{\arm{Batch session}$-$\arm{Session}, YooChoose purchase}}\\
GPT-5.6 & 0.143$^{*}$ & 0.143$^{*}$ & 0.119$^{*}$ & 0.124$^{*}$ \\
Sonnet5 & 0.109$^{*}$ & 0.109$^{*}$ & 0.091$^{*}$ & 0.113$^{*}$ \\
Gemini3.8 & 0.033$^{*}$ & 0.032$^{*}$ & 0.025 & 0.034$^{*}$ \\
GLM5.3 & -0.003 & -0.003 & -0.005 & -0.005 \\
Qwen3.8 & 0.029 & 0.029 & 0.031 & 0.023 \\
\bottomrule
\end{tabular}
\end{table}

\section{Detailed arm levels}
\label{app:levels}
Every table uses the method names and source mapping in Table~\ref{tab:arm-key}. We display point estimates to three decimal places and interval endpoints to four; figures use the full-precision machine-readable values. Coveo has two outcomes and YooChoose one. Each model uses its common support across all seven methods. The mean of five is an equal-weight average of model-specific correlations, not an ensemble of predictions or a correlation on pooled data.

\begin{table}[htbp]
\centering\footnotesize
\setlength{\tabcolsep}{2.5pt}
\caption{Coveo cart: rank correlations with 95\% product-bootstrap intervals, on each model's common support (Table~\ref{tab:full-coverage}); the mean of five has its own interval. R is the session-bag diagnostic.}
\label{tab:full-levels-coveo-cart}
\begin{tabularx}{\linewidth}{@{}l*{6}{>{\centering\arraybackslash}X}@{}}
\toprule
Arm & GPT-5.6 & Qwen3.8 & Gemini3.8 & GLM5.3 & Sonnet5 & Mean\\
\midrule
A & \(0.151\) & \(0.084\) & \(0.175\) & \(-0.006\) & \(0.000\) & \(0.081\)\\
 & {\scriptsize[0.0908,0.2115]} & {\scriptsize[0.0241,0.1470]} & {\scriptsize[0.1133,0.2353]} & {\scriptsize[-0.0690,0.0568]} & {\scriptsize[-0.0625,0.0653]} & {\scriptsize[0.0410,0.1213]}\\
\addlinespace[3pt]
C & \(0.313\) & \(0.246\) & \(0.281\) & \(0.258\) & \(0.278\) & \(0.275\)\\
 & {\scriptsize[0.2505,0.3675]} & {\scriptsize[0.1824,0.3029]} & {\scriptsize[0.2208,0.3357]} & {\scriptsize[0.1971,0.3143]} & {\scriptsize[0.2182,0.3338]} & {\scriptsize[0.2161,0.3275]}\\
\addlinespace[3pt]
D & \(0.319\) & \(0.246\) & \(0.261\) & \(0.175\) & \(0.156\) & \(0.231\)\\
 & {\scriptsize[0.2556,0.3739]} & {\scriptsize[0.1802,0.3054]} & {\scriptsize[0.1978,0.3176]} & {\scriptsize[0.1118,0.2347]} & {\scriptsize[0.0924,0.2166]} & {\scriptsize[0.1703,0.2866]}\\
\addlinespace[3pt]
B & \(0.438\) & \(0.433\) & \(0.356\) & \(0.252\) & \(0.228\) & \(0.341\)\\
 & {\scriptsize[0.3835,0.4912]} & {\scriptsize[0.3787,0.4851]} & {\scriptsize[0.2977,0.4138]} & {\scriptsize[0.1898,0.3123]} & {\scriptsize[0.1681,0.2921]} & {\scriptsize[0.2881,0.3959]}\\
\addlinespace[3pt]
\midrule
A80 & \(0.171\) & \(0.032\) & \(0.132\) & \(0.065\) & \(0.027\) & \(0.085\)\\
 & {\scriptsize[0.1086,0.2300]} & {\scriptsize[-0.0346,0.0913]} & {\scriptsize[0.0680,0.1919]} & {\scriptsize[0.0022,0.1246]} & {\scriptsize[-0.0326,0.0895]} & {\scriptsize[0.0424,0.1252]}\\
\addlinespace[3pt]
B80 & \(0.292\) & \(0.341\) & \(0.212\) & \(0.178\) & \(0.143\) & \(0.233\)\\
 & {\scriptsize[0.2321,0.3486]} & {\scriptsize[0.2808,0.3946]} & {\scriptsize[0.1477,0.2742]} & {\scriptsize[0.1133,0.2400]} & {\scriptsize[0.0785,0.2067]} & {\scriptsize[0.1722,0.2903]}\\
\addlinespace[3pt]
R & \(0.235\) & \(0.174\) & \(0.225\) & \(0.029\) & \(0.059\) & \(0.145\)\\
 & {\scriptsize[0.1741,0.2934]} & {\scriptsize[0.1129,0.2339]} & {\scriptsize[0.1624,0.2833]} & {\scriptsize[-0.0348,0.0930]} & {\scriptsize[-0.0049,0.1203]} & {\scriptsize[0.1054,0.1803]}\\
\addlinespace[3pt]
\bottomrule
\end{tabularx}
\end{table}

\begin{table}[htbp]
\centering\footnotesize
\setlength{\tabcolsep}{2.5pt}
\caption{Coveo purchase: rank correlations with 95\% product-bootstrap intervals, on each model's common support (Table~\ref{tab:full-coverage}); the mean of five has its own interval. R is the session-bag diagnostic.}
\label{tab:full-levels-coveo-purchase}
\begin{tabularx}{\linewidth}{@{}l*{6}{>{\centering\arraybackslash}X}@{}}
\toprule
Arm & GPT-5.6 & Qwen3.8 & Gemini3.8 & GLM5.3 & Sonnet5 & Mean\\
\midrule
A & \(0.161\) & \(0.031\) & \(0.220\) & \(0.056\) & \(-0.004\) & \(0.093\)\\
 & {\scriptsize[0.0984,0.2245]} & {\scriptsize[-0.0331,0.0940]} & {\scriptsize[0.1612,0.2793]} & {\scriptsize[-0.0059,0.1164]} & {\scriptsize[-0.0664,0.0572]} & {\scriptsize[0.0569,0.1298]}\\
\addlinespace[3pt]
C & \(0.328\) & \(0.293\) & \(0.311\) & \(0.287\) & \(0.301\) & \(0.304\)\\
 & {\scriptsize[0.2669,0.3825]} & {\scriptsize[0.2325,0.3477]} & {\scriptsize[0.2503,0.3657]} & {\scriptsize[0.2240,0.3418]} & {\scriptsize[0.2389,0.3561]} & {\scriptsize[0.2442,0.3552]}\\
\addlinespace[3pt]
D & \(0.322\) & \(0.274\) & \(0.283\) & \(0.236\) & \(0.205\) & \(0.264\)\\
 & {\scriptsize[0.2583,0.3773]} & {\scriptsize[0.2112,0.3274]} & {\scriptsize[0.2210,0.3411]} & {\scriptsize[0.1742,0.2930]} & {\scriptsize[0.1424,0.2649]} & {\scriptsize[0.2042,0.3181]}\\
\addlinespace[3pt]
B & \(0.471\) & \(0.496\) & \(0.420\) & \(0.292\) & \(0.269\) & \(0.390\)\\
 & {\scriptsize[0.4179,0.5202]} & {\scriptsize[0.4452,0.5450]} & {\scriptsize[0.3668,0.4742]} & {\scriptsize[0.2334,0.3510]} & {\scriptsize[0.2114,0.3272]} & {\scriptsize[0.3420,0.4382]}\\
\addlinespace[3pt]
\midrule
A80 & \(0.168\) & \(0.008\) & \(0.114\) & \(0.083\) & \(0.042\) & \(0.083\)\\
 & {\scriptsize[0.1036,0.2292]} & {\scriptsize[-0.0543,0.0676]} & {\scriptsize[0.0511,0.1738]} & {\scriptsize[0.0166,0.1440]} & {\scriptsize[-0.0223,0.1056]} & {\scriptsize[0.0422,0.1222]}\\
\addlinespace[3pt]
B80 & \(0.307\) & \(0.347\) & \(0.240\) & \(0.190\) & \(0.142\) & \(0.245\)\\
 & {\scriptsize[0.2426,0.3626]} & {\scriptsize[0.2875,0.4032]} & {\scriptsize[0.1747,0.3004]} & {\scriptsize[0.1263,0.2502]} & {\scriptsize[0.0792,0.2032]} & {\scriptsize[0.1856,0.2992]}\\
\addlinespace[3pt]
R & \(0.225\) & \(0.185\) & \(0.189\) & \(0.009\) & \(0.077\) & \(0.137\)\\
 & {\scriptsize[0.1643,0.2812]} & {\scriptsize[0.1247,0.2496]} & {\scriptsize[0.1289,0.2496]} & {\scriptsize[-0.0543,0.0722]} & {\scriptsize[0.0153,0.1378]} & {\scriptsize[0.0984,0.1746]}\\
\addlinespace[3pt]
\bottomrule
\end{tabularx}
\end{table}

\begin{table}[htbp]
\centering\footnotesize
\setlength{\tabcolsep}{2.5pt}
\caption{YooChoose purchase: rank correlations with 95\% product-bootstrap intervals, on each model's common support (Table~\ref{tab:full-coverage}); the mean of five has its own interval. R is the session-bag diagnostic.}
\label{tab:full-levels-yoochoose-purchase}
\begin{tabularx}{\linewidth}{@{}l*{6}{>{\centering\arraybackslash}X}@{}}
\toprule
Arm & GPT-5.6 & Qwen3.8 & Gemini3.8 & GLM5.3 & Sonnet5 & Mean\\
\midrule
A & \(0.415\) & \(0.251\) & \(0.350\) & \(0.249\) & \(0.192\) & \(0.291\)\\
 & {\scriptsize[0.3625,0.4649]} & {\scriptsize[0.1899,0.3103]} & {\scriptsize[0.2931,0.4044]} & {\scriptsize[0.1895,0.3057]} & {\scriptsize[0.1284,0.2508]} & {\scriptsize[0.2524,0.3276]}\\
\addlinespace[3pt]
C & \(0.417\) & \(0.502\) & \(0.429\) & \(0.490\) & \(0.278\) & \(0.423\)\\
 & {\scriptsize[0.3617,0.4663]} & {\scriptsize[0.4513,0.5480]} & {\scriptsize[0.3732,0.4806]} & {\scriptsize[0.4401,0.5341]} & {\scriptsize[0.2162,0.3359]} & {\scriptsize[0.3730,0.4684]}\\
\addlinespace[3pt]
D & \(0.560\) & \(0.530\) & \(0.461\) & \(0.487\) & \(0.387\) & \(0.485\)\\
 & {\scriptsize[0.5126,0.6039]} & {\scriptsize[0.4792,0.5762]} & {\scriptsize[0.4069,0.5124]} & {\scriptsize[0.4349,0.5345]} & {\scriptsize[0.3287,0.4395]} & {\scriptsize[0.4343,0.5314]}\\
\addlinespace[3pt]
B & \(0.597\) & \(0.584\) & \(0.601\) & \(0.517\) & \(0.407\) & \(0.541\)\\
 & {\scriptsize[0.5541,0.6375]} & {\scriptsize[0.5378,0.6258]} & {\scriptsize[0.5533,0.6426]} & {\scriptsize[0.4654,0.5642]} & {\scriptsize[0.3485,0.4597]} & {\scriptsize[0.4983,0.5796]}\\
\addlinespace[3pt]
\midrule
A80 & \(0.421\) & \(0.255\) & \(0.388\) & \(0.180\) & \(0.222\) & \(0.293\)\\
 & {\scriptsize[0.3670,0.4700]} & {\scriptsize[0.1934,0.3138]} & {\scriptsize[0.3307,0.4426]} & {\scriptsize[0.1194,0.2373]} & {\scriptsize[0.1615,0.2844]} & {\scriptsize[0.2528,0.3324]}\\
\addlinespace[3pt]
B80 & \(0.564\) & \(0.544\) & \(0.539\) & \(0.485\) & \(0.371\) & \(0.501\)\\
 & {\scriptsize[0.5160,0.6066]} & {\scriptsize[0.4970,0.5867]} & {\scriptsize[0.4855,0.5843]} & {\scriptsize[0.4330,0.5321]} & {\scriptsize[0.3138,0.4237]} & {\scriptsize[0.4557,0.5418]}\\
\addlinespace[3pt]
R & \(0.407\) & \(0.154\) & \(0.291\) & \(0.194\) & \(0.202\) & \(0.250\)\\
 & {\scriptsize[0.3545,0.4581]} & {\scriptsize[0.0912,0.2137]} & {\scriptsize[0.2327,0.3468]} & {\scriptsize[0.1327,0.2517]} & {\scriptsize[0.1431,0.2614]} & {\scriptsize[0.2121,0.2851]}\\
\addlinespace[3pt]
\bottomrule
\end{tabularx}
\end{table}

\FloatBarrier
\section{Detailed paired contrasts}
\label{app:contrasts}
The following tables report the 15 pairwise contrasts for every model and outcome. A star marks a reported paired 95\% interval excluding zero, based on its unrounded endpoints. For a model mean, the interval uses the distribution of the mean paired contrast across shared bootstrap draws; it cannot be constructed by averaging or subtracting interval endpoints. These are individual comparisons without a multiplicity adjustment.

The tables do not include \arm{Batch summary-80} against \arm{Session}, which we compute from the same predictions and bootstrap draws. The model-mean difference is \(-0.042\;[-0.0763,-0.0059]\) for Coveo cart, \(-0.059\;[-0.0921,-0.0246]\) for Coveo purchase and \(0.078\;[0.0471,0.1090]\) for YooChoose. On both Coveo outcomes, \arm{Session} is higher for Sonnet, Gemini and GLM and \arm{Batch summary-80} is higher for Qwen, with intervals excluding zero, and GPT is indistinguishable; on YooChoose \arm{Batch summary-80} is higher for four models and GLM is indistinguishable. Against \arm{Batch session}, \arm{Batch summary-80} is ahead with an interval excluding zero in 3 of the 15 settings and behind in 4.

\begin{table}[htbp]
\centering\small
\setlength{\tabcolsep}{2.5pt}
\caption{Coveo cart: paired differences in rank correlation, group 1 of 3, with 95\% product-bootstrap intervals. A star marks an unrounded interval excluding zero; no multiplicity adjustment.}
\label{tab:full-contrasts-coveo-cart-1}
\begin{tabularx}{\linewidth}{@{}l*{5}{>{\centering\arraybackslash}X}@{}}
\toprule
Model & \(B{-}A\) & \(D{-}C\) & \(B{-}C\) & \(D{-}A\) & \(C{-}A\)\\
\midrule
GPT-5.6 & \(\mathbf{+0.287}^{*}\) & \(+0.006\) & \(\mathbf{+0.126}^{*}\) & \(\mathbf{+0.167}^{*}\) & \(\mathbf{+0.161}^{*}\)\\
 & {\scriptsize[0.2239,0.3481]} & {\scriptsize[-0.0182,0.0293]} & {\scriptsize[0.0786,0.1744]} & {\scriptsize[0.0988,0.2359]} & {\scriptsize[0.0922,0.2308]}\\
\addlinespace[4pt]
Qwen3.8 & \(\mathbf{+0.349}^{*}\) & \(+0.000\) & \(\mathbf{+0.187}^{*}\) & \(\mathbf{+0.162}^{*}\) & \(\mathbf{+0.162}^{*}\)\\
 & {\scriptsize[0.2758,0.4227]} & {\scriptsize[-0.0273,0.0265]} & {\scriptsize[0.1357,0.2431]} & {\scriptsize[0.0781,0.2344]} & {\scriptsize[0.0772,0.2360]}\\
\addlinespace[4pt]
Gemini3.8 & \(\mathbf{+0.181}^{*}\) & \(-0.021\) & \(\mathbf{+0.075}^{*}\) & \(\mathbf{+0.085}^{*}\) & \(\mathbf{+0.106}^{*}\)\\
 & {\scriptsize[0.1141,0.2427]} & {\scriptsize[-0.0441,0.0023]} & {\scriptsize[0.0212,0.1300]} & {\scriptsize[0.0130,0.1546]} & {\scriptsize[0.0321,0.1739]}\\
\addlinespace[4pt]
GLM5.3 & \(\mathbf{+0.258}^{*}\) & \(\mathbf{-0.083}^{*}\) & \(-0.006\) & \(\mathbf{+0.181}^{*}\) & \(\mathbf{+0.264}^{*}\)\\
 & {\scriptsize[0.1799,0.3373]} & {\scriptsize[-0.1212,-0.0434]} & {\scriptsize[-0.0734,0.0618]} & {\scriptsize[0.0983,0.2571]} & {\scriptsize[0.1818,0.3418]}\\
\addlinespace[4pt]
Sonnet5 & \(\mathbf{+0.228}^{*}\) & \(\mathbf{-0.122}^{*}\) & \(-0.050\) & \(\mathbf{+0.156}^{*}\) & \(\mathbf{+0.278}^{*}\)\\
 & {\scriptsize[0.1525,0.2984]} & {\scriptsize[-0.1664,-0.0768]} & {\scriptsize[-0.1130,0.0117]} & {\scriptsize[0.0788,0.2328]} & {\scriptsize[0.1973,0.3527]}\\
\addlinespace[4pt]
\textit{Mean of five} & \(\mathbf{+0.260}^{*}\) & \(\mathbf{-0.044}^{*}\) & \(\mathbf{+0.066}^{*}\) & \(\mathbf{+0.150}^{*}\) & \(\mathbf{+0.194}^{*}\)\\
 & {\scriptsize[0.2104,0.3082]} & {\scriptsize[-0.0675,-0.0202]} & {\scriptsize[0.0174,0.1184]} & {\scriptsize[0.0918,0.2025]} & {\scriptsize[0.1341,0.2471]}\\
\addlinespace[4pt]
\bottomrule
\end{tabularx}
\end{table}

\begin{table}[htbp]
\centering\small
\setlength{\tabcolsep}{2.5pt}
\caption{Coveo cart: paired differences in rank correlation, group 2 of 3, with 95\% product-bootstrap intervals. A star marks an unrounded interval excluding zero; no multiplicity adjustment.}
\label{tab:full-contrasts-coveo-cart-2}
\begin{tabularx}{\linewidth}{@{}l*{5}{>{\centering\arraybackslash}X}@{}}
\toprule
Model & \(C{-}A80\) & \(R{-}A80\) & \(C{-}R\) & \(A{-}A80\) & \(B{-}B80\)\\
\midrule
GPT-5.6 & \(\mathbf{+0.142}^{*}\) & \(+0.065\) & \(\mathbf{+0.077}^{*}\) & \(-0.019\) & \(\mathbf{+0.146}^{*}\)\\
 & {\scriptsize[0.0848,0.1957]} & {\scriptsize[-0.0075,0.1295]} & {\scriptsize[0.0239,0.1289]} & {\scriptsize[-0.0920,0.0530]} & {\scriptsize[0.1001,0.1943]}\\
\addlinespace[4pt]
Qwen3.8 & \(\mathbf{+0.214}^{*}\) & \(\mathbf{+0.142}^{*}\) & \(\mathbf{+0.072}^{*}\) & \(+0.052\) & \(\mathbf{+0.092}^{*}\)\\
 & {\scriptsize[0.1477,0.2831]} & {\scriptsize[0.0645,0.2210]} & {\scriptsize[0.0016,0.1387]} & {\scriptsize[-0.0241,0.1322]} & {\scriptsize[0.0420,0.1440]}\\
\addlinespace[4pt]
Gemini3.8 & \(\mathbf{+0.149}^{*}\) & \(\mathbf{+0.093}^{*}\) & \(\mathbf{+0.057}^{*}\) & \(+0.044\) & \(\mathbf{+0.144}^{*}\)\\
 & {\scriptsize[0.0907,0.2060]} & {\scriptsize[0.0213,0.1632]} & {\scriptsize[0.0002,0.1131]} & {\scriptsize[-0.0331,0.1195]} & {\scriptsize[0.0908,0.1995]}\\
\addlinespace[4pt]
GLM5.3 & \(\mathbf{+0.192}^{*}\) & \(-0.036\) & \(\mathbf{+0.228}^{*}\) & \(-0.071\) & \(\mathbf{+0.074}^{*}\)\\
 & {\scriptsize[0.1179,0.2656]} & {\scriptsize[-0.1224,0.0568]} & {\scriptsize[0.1512,0.3067]} & {\scriptsize[-0.1601,0.0222]} & {\scriptsize[0.0245,0.1248]}\\
\addlinespace[4pt]
Sonnet5 & \(\mathbf{+0.251}^{*}\) & \(+0.033\) & \(\mathbf{+0.219}^{*}\) & \(-0.027\) & \(\mathbf{+0.085}^{*}\)\\
 & {\scriptsize[0.1919,0.3123]} & {\scriptsize[-0.0433,0.1122]} & {\scriptsize[0.1495,0.2870]} & {\scriptsize[-0.0915,0.0488]} & {\scriptsize[0.0398,0.1329]}\\
\addlinespace[4pt]
\textit{Mean of five} & \(\mathbf{+0.190}^{*}\) & \(\mathbf{+0.059}^{*}\) & \(\mathbf{+0.131}^{*}\) & \(-0.004\) & \(\mathbf{+0.108}^{*}\)\\
 & {\scriptsize[0.1467,0.2288]} & {\scriptsize[0.0185,0.0991]} & {\scriptsize[0.0876,0.1704]} & {\scriptsize[-0.0470,0.0408]} & {\scriptsize[0.0639,0.1548]}\\
\addlinespace[4pt]
\bottomrule
\end{tabularx}
\end{table}

\begin{table}[htbp]
\centering\small
\setlength{\tabcolsep}{2.5pt}
\caption{Coveo cart: paired differences in rank correlation, group 3 of 3, with 95\% product-bootstrap intervals. A star marks an unrounded interval excluding zero; no multiplicity adjustment.}
\label{tab:full-contrasts-coveo-cart-3}
\begin{tabularx}{\linewidth}{@{}l*{5}{>{\centering\arraybackslash}X}@{}}
\toprule
Model & \(B80{-}D\) & \(B{-}D\) & \(B{-}R\) & \(R{-}A\) & \(B80{-}A80\)\\
\midrule
GPT-5.6 & \(-0.027\) & \(\mathbf{+0.120}^{*}\) & \(\mathbf{+0.203}^{*}\) & \(\mathbf{+0.084}^{*}\) & \(\mathbf{+0.121}^{*}\)\\
 & {\scriptsize[-0.0644,0.0129]} & {\scriptsize[0.0697,0.1693]} & {\scriptsize[0.1388,0.2660]} & {\scriptsize[0.0066,0.1550]} & {\scriptsize[0.0649,0.1732]}\\
\addlinespace[4pt]
Qwen3.8 & \(\mathbf{+0.095}^{*}\) & \(\mathbf{+0.187}^{*}\) & \(\mathbf{+0.259}^{*}\) & \(\mathbf{+0.090}^{*}\) & \(\mathbf{+0.309}^{*}\)\\
 & {\scriptsize[0.0560,0.1371]} & {\scriptsize[0.1347,0.2449]} & {\scriptsize[0.1848,0.3333]} & {\scriptsize[0.0032,0.1690]} & {\scriptsize[0.2440,0.3765]}\\
\addlinespace[4pt]
Gemini3.8 & \(\mathbf{-0.048}^{*}\) & \(\mathbf{+0.095}^{*}\) & \(\mathbf{+0.131}^{*}\) & \(+0.049\) & \(\mathbf{+0.080}^{*}\)\\
 & {\scriptsize[-0.0825,-0.0139]} & {\scriptsize[0.0414,0.1532]} & {\scriptsize[0.0633,0.2041]} & {\scriptsize[-0.0320,0.1251]} & {\scriptsize[0.0190,0.1400]}\\
\addlinespace[4pt]
GLM5.3 & \(+0.004\) & \(\mathbf{+0.077}^{*}\) & \(\mathbf{+0.223}^{*}\) & \(+0.036\) & \(\mathbf{+0.113}^{*}\)\\
 & {\scriptsize[-0.0414,0.0496]} & {\scriptsize[0.0160,0.1403]} & {\scriptsize[0.1395,0.3083]} & {\scriptsize[-0.0470,0.1212]} & {\scriptsize[0.0370,0.1908]}\\
\addlinespace[4pt]
Sonnet5 & \(-0.013\) & \(\mathbf{+0.072}^{*}\) & \(\mathbf{+0.169}^{*}\) & \(+0.059\) & \(\mathbf{+0.116}^{*}\)\\
 & {\scriptsize[-0.0643,0.0387]} & {\scriptsize[0.0089,0.1331]} & {\scriptsize[0.0870,0.2489]} & {\scriptsize[-0.0312,0.1486]} & {\scriptsize[0.0612,0.1731]}\\
\addlinespace[4pt]
\textit{Mean of five} & \(+0.002\) & \(\mathbf{+0.110}^{*}\) & \(\mathbf{+0.197}^{*}\) & \(\mathbf{+0.064}^{*}\) & \(\mathbf{+0.148}^{*}\)\\
 & {\scriptsize[-0.0320,0.0366]} & {\scriptsize[0.0594,0.1626]} & {\scriptsize[0.1434,0.2517]} & {\scriptsize[0.0164,0.1079]} & {\scriptsize[0.1065,0.1879]}\\
\addlinespace[4pt]
\bottomrule
\end{tabularx}
\end{table}

\begin{table}[htbp]
\centering\small
\setlength{\tabcolsep}{2.5pt}
\caption{Coveo purchase: paired differences in rank correlation, group 1 of 3, with 95\% product-bootstrap intervals. A star marks an unrounded interval excluding zero; no multiplicity adjustment.}
\label{tab:full-contrasts-coveo-purchase-1}
\begin{tabularx}{\linewidth}{@{}l*{5}{>{\centering\arraybackslash}X}@{}}
\toprule
Model & \(B{-}A\) & \(D{-}C\) & \(B{-}C\) & \(D{-}A\) & \(C{-}A\)\\
\midrule
GPT-5.6 & \(\mathbf{+0.310}^{*}\) & \(-0.007\) & \(\mathbf{+0.143}^{*}\) & \(\mathbf{+0.161}^{*}\) & \(\mathbf{+0.168}^{*}\)\\
 & {\scriptsize[0.2475,0.3697]} & {\scriptsize[-0.0334,0.0190]} & {\scriptsize[0.0980,0.1880]} & {\scriptsize[0.0908,0.2260]} & {\scriptsize[0.0958,0.2345]}\\
\addlinespace[4pt]
Qwen3.8 & \(\mathbf{+0.465}^{*}\) & \(-0.020\) & \(\mathbf{+0.203}^{*}\) & \(\mathbf{+0.243}^{*}\) & \(\mathbf{+0.263}^{*}\)\\
 & {\scriptsize[0.3902,0.5357]} & {\scriptsize[-0.0475,0.0091]} & {\scriptsize[0.1525,0.2559]} & {\scriptsize[0.1637,0.3172]} & {\scriptsize[0.1850,0.3395]}\\
\addlinespace[4pt]
Gemini3.8 & \(\mathbf{+0.201}^{*}\) & \(\mathbf{-0.027}^{*}\) & \(\mathbf{+0.110}^{*}\) & \(+0.064\) & \(\mathbf{+0.091}^{*}\)\\
 & {\scriptsize[0.1345,0.2620]} & {\scriptsize[-0.0509,-0.0034]} & {\scriptsize[0.0615,0.1608]} & {\scriptsize[-0.0099,0.1323]} & {\scriptsize[0.0176,0.1582]}\\
\addlinespace[4pt]
GLM5.3 & \(\mathbf{+0.236}^{*}\) & \(\mathbf{-0.051}^{*}\) & \(+0.005\) & \(\mathbf{+0.180}^{*}\) & \(\mathbf{+0.231}^{*}\)\\
 & {\scriptsize[0.1594,0.3119]} & {\scriptsize[-0.0917,-0.0091]} & {\scriptsize[-0.0609,0.0737]} & {\scriptsize[0.0995,0.2607]} & {\scriptsize[0.1476,0.3100]}\\
\addlinespace[4pt]
Sonnet5 & \(\mathbf{+0.273}^{*}\) & \(\mathbf{-0.096}^{*}\) & \(-0.032\) & \(\mathbf{+0.209}^{*}\) & \(\mathbf{+0.305}^{*}\)\\
 & {\scriptsize[0.2012,0.3417]} & {\scriptsize[-0.1420,-0.0470]} & {\scriptsize[-0.0977,0.0316]} & {\scriptsize[0.1326,0.2896]} & {\scriptsize[0.2251,0.3825]}\\
\addlinespace[4pt]
\textit{Mean of five} & \(\mathbf{+0.297}^{*}\) & \(\mathbf{-0.040}^{*}\) & \(\mathbf{+0.086}^{*}\) & \(\mathbf{+0.171}^{*}\) & \(\mathbf{+0.211}^{*}\)\\
 & {\scriptsize[0.2520,0.3387]} & {\scriptsize[-0.0648,-0.0152]} & {\scriptsize[0.0391,0.1314]} & {\scriptsize[0.1151,0.2261]} & {\scriptsize[0.1548,0.2640]}\\
\addlinespace[4pt]
\bottomrule
\end{tabularx}
\end{table}

\begin{table}[htbp]
\centering\small
\setlength{\tabcolsep}{2.5pt}
\caption{Coveo purchase: paired differences in rank correlation, group 2 of 3, with 95\% product-bootstrap intervals. A star marks an unrounded interval excluding zero; no multiplicity adjustment.}
\label{tab:full-contrasts-coveo-purchase-2}
\begin{tabularx}{\linewidth}{@{}l*{5}{>{\centering\arraybackslash}X}@{}}
\toprule
Model & \(C{-}A80\) & \(R{-}A80\) & \(C{-}R\) & \(A{-}A80\) & \(B{-}B80\)\\
\midrule
GPT-5.6 & \(\mathbf{+0.160}^{*}\) & \(+0.057\) & \(\mathbf{+0.104}^{*}\) & \(-0.007\) & \(\mathbf{+0.164}^{*}\)\\
 & {\scriptsize[0.1033,0.2155]} & {\scriptsize[-0.0148,0.1266]} & {\scriptsize[0.0487,0.1564]} & {\scriptsize[-0.0759,0.0660]} & {\scriptsize[0.1205,0.2077]}\\
\addlinespace[4pt]
Qwen3.8 & \(\mathbf{+0.285}^{*}\) & \(\mathbf{+0.177}^{*}\) & \(\mathbf{+0.108}^{*}\) & \(+0.023\) & \(\mathbf{+0.149}^{*}\)\\
 & {\scriptsize[0.2115,0.3535]} & {\scriptsize[0.0966,0.2563]} & {\scriptsize[0.0432,0.1729]} & {\scriptsize[-0.0589,0.0983]} & {\scriptsize[0.1054,0.1959]}\\
\addlinespace[4pt]
Gemini3.8 & \(\mathbf{+0.196}^{*}\) & \(\mathbf{+0.074}^{*}\) & \(\mathbf{+0.122}^{*}\) & \(\mathbf{+0.105}^{*}\) & \(\mathbf{+0.181}^{*}\)\\
 & {\scriptsize[0.1327,0.2563]} & {\scriptsize[0.0011,0.1458]} & {\scriptsize[0.0641,0.1739]} & {\scriptsize[0.0292,0.1826]} & {\scriptsize[0.1303,0.2346]}\\
\addlinespace[4pt]
GLM5.3 & \(\mathbf{+0.204}^{*}\) & \(-0.074\) & \(\mathbf{+0.278}^{*}\) & \(-0.027\) & \(\mathbf{+0.102}^{*}\)\\
 & {\scriptsize[0.1297,0.2767]} & {\scriptsize[-0.1581,0.0197]} & {\scriptsize[0.1951,0.3534]} & {\scriptsize[-0.1150,0.0601]} & {\scriptsize[0.0535,0.1526]}\\
\addlinespace[4pt]
Sonnet5 & \(\mathbf{+0.259}^{*}\) & \(+0.036\) & \(\mathbf{+0.224}^{*}\) & \(-0.046\) & \(\mathbf{+0.127}^{*}\)\\
 & {\scriptsize[0.2011,0.3201]} & {\scriptsize[-0.0414,0.1098]} & {\scriptsize[0.1544,0.2925]} & {\scriptsize[-0.1209,0.0303]} & {\scriptsize[0.0775,0.1722]}\\
\addlinespace[4pt]
\textit{Mean of five} & \(\mathbf{+0.221}^{*}\) & \(\mathbf{+0.054}^{*}\) & \(\mathbf{+0.167}^{*}\) & \(+0.010\) & \(\mathbf{+0.145}^{*}\)\\
 & {\scriptsize[0.1784,0.2610]} & {\scriptsize[0.0168,0.0943]} & {\scriptsize[0.1242,0.2054]} & {\scriptsize[-0.0336,0.0531]} & {\scriptsize[0.1003,0.1882]}\\
\addlinespace[4pt]
\bottomrule
\end{tabularx}
\end{table}

\begin{table}[htbp]
\centering\small
\setlength{\tabcolsep}{2.5pt}
\caption{Coveo purchase: paired differences in rank correlation, group 3 of 3, with 95\% product-bootstrap intervals. A star marks an unrounded interval excluding zero; no multiplicity adjustment.}
\label{tab:full-contrasts-coveo-purchase-3}
\begin{tabularx}{\linewidth}{@{}l*{5}{>{\centering\arraybackslash}X}@{}}
\toprule
Model & \(B80{-}D\) & \(B{-}D\) & \(B{-}R\) & \(R{-}A\) & \(B80{-}A80\)\\
\midrule
GPT-5.6 & \(-0.015\) & \(\mathbf{+0.150}^{*}\) & \(\mathbf{+0.246}^{*}\) & \(+0.064\) & \(\mathbf{+0.139}^{*}\)\\
 & {\scriptsize[-0.0538,0.0242]} & {\scriptsize[0.1034,0.2014]} & {\scriptsize[0.1842,0.3070]} & {\scriptsize[-0.0120,0.1369]} & {\scriptsize[0.0867,0.1920]}\\
\addlinespace[4pt]
Qwen3.8 & \(\mathbf{+0.074}^{*}\) & \(\mathbf{+0.223}^{*}\) & \(\mathbf{+0.311}^{*}\) & \(\mathbf{+0.154}^{*}\) & \(\mathbf{+0.339}^{*}\)\\
 & {\scriptsize[0.0284,0.1175]} & {\scriptsize[0.1684,0.2779]} & {\scriptsize[0.2401,0.3762]} & {\scriptsize[0.0715,0.2403]} & {\scriptsize[0.2668,0.4103]}\\
\addlinespace[4pt]
Gemini3.8 & \(\mathbf{-0.044}^{*}\) & \(\mathbf{+0.137}^{*}\) & \(\mathbf{+0.232}^{*}\) & \(-0.031\) & \(\mathbf{+0.125}^{*}\)\\
 & {\scriptsize[-0.0773,-0.0086]} & {\scriptsize[0.0867,0.1895]} & {\scriptsize[0.1663,0.2959]} & {\scriptsize[-0.1078,0.0441]} & {\scriptsize[0.0645,0.1857]}\\
\addlinespace[4pt]
GLM5.3 & \(\mathbf{-0.046}^{*}\) & \(+0.056\) & \(\mathbf{+0.283}^{*}\) & \(-0.047\) & \(\mathbf{+0.107}^{*}\)\\
 & {\scriptsize[-0.0899,-0.0022]} & {\scriptsize[-0.0039,0.1129]} & {\scriptsize[0.2043,0.3639]} & {\scriptsize[-0.1316,0.0379]} & {\scriptsize[0.0326,0.1848]}\\
\addlinespace[4pt]
Sonnet5 & \(\mathbf{-0.063}^{*}\) & \(\mathbf{+0.064}^{*}\) & \(\mathbf{+0.192}^{*}\) & \(+0.081\) & \(\mathbf{+0.101}^{*}\)\\
 & {\scriptsize[-0.1106,-0.0154]} & {\scriptsize[0.0022,0.1245]} & {\scriptsize[0.1093,0.2672]} & {\scriptsize[-0.0063,0.1657]} & {\scriptsize[0.0445,0.1578]}\\
\addlinespace[4pt]
\textit{Mean of five} & \(-0.019\) & \(\mathbf{+0.126}^{*}\) & \(\mathbf{+0.253}^{*}\) & \(+0.044\) & \(\mathbf{+0.162}^{*}\)\\
 & {\scriptsize[-0.0522,0.0156]} & {\scriptsize[0.0787,0.1754]} & {\scriptsize[0.2023,0.3007]} & {\scriptsize[-0.0006,0.0907]} & {\scriptsize[0.1204,0.2024]}\\
\addlinespace[4pt]
\bottomrule
\end{tabularx}
\end{table}

\begin{table}[htbp]
\centering\small
\setlength{\tabcolsep}{2.5pt}
\caption{YooChoose purchase: paired differences in rank correlation, group 1 of 3, with 95\% product-bootstrap intervals. A star marks an unrounded interval excluding zero; no multiplicity adjustment.}
\label{tab:full-contrasts-yoochoose-purchase-1}
\begin{tabularx}{\linewidth}{@{}l*{5}{>{\centering\arraybackslash}X}@{}}
\toprule
Model & \(B{-}A\) & \(D{-}C\) & \(B{-}C\) & \(D{-}A\) & \(C{-}A\)\\
\midrule
GPT-5.6 & \(\mathbf{+0.183}^{*}\) & \(\mathbf{+0.143}^{*}\) & \(\mathbf{+0.180}^{*}\) & \(\mathbf{+0.145}^{*}\) & \(+0.002\)\\
 & {\scriptsize[0.1418,0.2202]} & {\scriptsize[0.1104,0.1760]} & {\scriptsize[0.1269,0.2366]} & {\scriptsize[0.0898,0.2021]} & {\scriptsize[-0.0621,0.0644]}\\
\addlinespace[4pt]
Qwen3.8 & \(\mathbf{+0.333}^{*}\) & \(+0.029\) & \(\mathbf{+0.082}^{*}\) & \(\mathbf{+0.280}^{*}\) & \(\mathbf{+0.251}^{*}\)\\
 & {\scriptsize[0.2751,0.3910]} & {\scriptsize[-0.0028,0.0601]} & {\scriptsize[0.0375,0.1291]} & {\scriptsize[0.2049,0.3534]} & {\scriptsize[0.1810,0.3197]}\\
\addlinespace[4pt]
Gemini3.8 & \(\mathbf{+0.251}^{*}\) & \(\mathbf{+0.033}^{*}\) & \(\mathbf{+0.172}^{*}\) & \(\mathbf{+0.111}^{*}\) & \(\mathbf{+0.079}^{*}\)\\
 & {\scriptsize[0.1969,0.3062]} & {\scriptsize[0.0041,0.0600]} & {\scriptsize[0.1283,0.2150]} & {\scriptsize[0.0471,0.1805]} & {\scriptsize[0.0112,0.1456]}\\
\addlinespace[4pt]
GLM5.3 & \(\mathbf{+0.269}^{*}\) & \(-0.003\) & \(+0.028\) & \(\mathbf{+0.238}^{*}\) & \(\mathbf{+0.241}^{*}\)\\
 & {\scriptsize[0.2007,0.3365]} & {\scriptsize[-0.0444,0.0401]} & {\scriptsize[-0.0112,0.0655]} & {\scriptsize[0.1621,0.3086]} & {\scriptsize[0.1759,0.3023]}\\
\addlinespace[4pt]
Sonnet5 & \(\mathbf{+0.215}^{*}\) & \(\mathbf{+0.109}^{*}\) & \(\mathbf{+0.130}^{*}\) & \(\mathbf{+0.195}^{*}\) & \(\mathbf{+0.086}^{*}\)\\
 & {\scriptsize[0.1522,0.2840]} & {\scriptsize[0.0665,0.1537]} & {\scriptsize[0.0862,0.1746]} & {\scriptsize[0.1305,0.2603]} & {\scriptsize[0.0209,0.1540]}\\
\addlinespace[4pt]
\textit{Mean of five} & \(\mathbf{+0.250}^{*}\) & \(\mathbf{+0.062}^{*}\) & \(\mathbf{+0.118}^{*}\) & \(\mathbf{+0.194}^{*}\) & \(\mathbf{+0.132}^{*}\)\\
 & {\scriptsize[0.2139,0.2843]} & {\scriptsize[0.0338,0.0900]} & {\scriptsize[0.0833,0.1550]} & {\scriptsize[0.1426,0.2436]} & {\scriptsize[0.0845,0.1782]}\\
\addlinespace[4pt]
\bottomrule
\end{tabularx}
\end{table}

\begin{table}[htbp]
\centering\small
\setlength{\tabcolsep}{2.5pt}
\caption{YooChoose purchase: paired differences in rank correlation, group 2 of 3, with 95\% product-bootstrap intervals. A star marks an unrounded interval excluding zero; no multiplicity adjustment.}
\label{tab:full-contrasts-yoochoose-purchase-2}
\begin{tabularx}{\linewidth}{@{}l*{5}{>{\centering\arraybackslash}X}@{}}
\toprule
Model & \(C{-}A80\) & \(R{-}A80\) & \(C{-}R\) & \(A{-}A80\) & \(B{-}B80\)\\
\midrule
GPT-5.6 & \(-0.004\) & \(-0.014\) & \(+0.010\) & \(-0.006\) & \(\mathbf{+0.033}^{*}\)\\
 & {\scriptsize[-0.0638,0.0524]} & {\scriptsize[-0.0766,0.0488]} & {\scriptsize[-0.0394,0.0573]} & {\scriptsize[-0.0523,0.0388]} & {\scriptsize[0.0085,0.0577]}\\
\addlinespace[4pt]
Qwen3.8 & \(\mathbf{+0.247}^{*}\) & \(\mathbf{-0.101}^{*}\) & \(\mathbf{+0.348}^{*}\) & \(-0.004\) & \(\mathbf{+0.040}^{*}\)\\
 & {\scriptsize[0.1837,0.3103]} & {\scriptsize[-0.1798,-0.0211]} & {\scriptsize[0.2809,0.4152]} & {\scriptsize[-0.0733,0.0629]} & {\scriptsize[0.0150,0.0655]}\\
\addlinespace[4pt]
Gemini3.8 & \(+0.041\) & \(\mathbf{-0.097}^{*}\) & \(\mathbf{+0.138}^{*}\) & \(-0.037\) & \(\mathbf{+0.062}^{*}\)\\
 & {\scriptsize[-0.0239,0.1015]} & {\scriptsize[-0.1719,-0.0240]} & {\scriptsize[0.0845,0.1900]} & {\scriptsize[-0.1002,0.0200]} & {\scriptsize[0.0304,0.0942]}\\
\addlinespace[4pt]
GLM5.3 & \(\mathbf{+0.309}^{*}\) & \(+0.014\) & \(\mathbf{+0.296}^{*}\) & \(+0.068\) & \(+0.032\)\\
 & {\scriptsize[0.2478,0.3705]} & {\scriptsize[-0.0644,0.0938]} & {\scriptsize[0.2244,0.3627]} & {\scriptsize[-0.0087,0.1442]} & {\scriptsize[-0.0022,0.0672]}\\
\addlinespace[4pt]
Sonnet5 & \(+0.055\) & \(-0.020\) & \(\mathbf{+0.075}^{*}\) & \(-0.031\) & \(\mathbf{+0.036}^{*}\)\\
 & {\scriptsize[-0.0073,0.1188]} & {\scriptsize[-0.0918,0.0504]} & {\scriptsize[0.0064,0.1395]} & {\scriptsize[-0.0988,0.0336]} & {\scriptsize[0.0016,0.0701]}\\
\addlinespace[4pt]
\textit{Mean of five} & \(\mathbf{+0.130}^{*}\) & \(\mathbf{-0.044}^{*}\) & \(\mathbf{+0.173}^{*}\) & \(-0.002\) & \(\mathbf{+0.041}^{*}\)\\
 & {\scriptsize[0.0864,0.1710]} & {\scriptsize[-0.0846,-0.0019]} & {\scriptsize[0.1365,0.2084]} & {\scriptsize[-0.0353,0.0293]} & {\scriptsize[0.0181,0.0634]}\\
\addlinespace[4pt]
\bottomrule
\end{tabularx}
\end{table}

\begin{table}[htbp]
\centering\small
\setlength{\tabcolsep}{2.5pt}
\caption{YooChoose purchase: paired differences in rank correlation, group 3 of 3, with 95\% product-bootstrap intervals. A star marks an unrounded interval excluding zero; no multiplicity adjustment.}
\label{tab:full-contrasts-yoochoose-purchase-3}
\begin{tabularx}{\linewidth}{@{}l*{5}{>{\centering\arraybackslash}X}@{}}
\toprule
Model & \(B80{-}D\) & \(B{-}D\) & \(B{-}R\) & \(R{-}A\) & \(B80{-}A80\)\\
\midrule
GPT-5.6 & \(+0.004\) & \(+0.037\) & \(\mathbf{+0.191}^{*}\) & \(-0.008\) & \(\mathbf{+0.143}^{*}\)\\
 & {\scriptsize[-0.0341,0.0437]} & {\scriptsize[-0.0093,0.0846]} & {\scriptsize[0.1330,0.2511]} & {\scriptsize[-0.0694,0.0532]} & {\scriptsize[0.1046,0.1796]}\\
\addlinespace[4pt]
Qwen3.8 & \(+0.013\) & \(\mathbf{+0.053}^{*}\) & \(\mathbf{+0.430}^{*}\) & \(\mathbf{-0.097}^{*}\) & \(\mathbf{+0.289}^{*}\)\\
 & {\scriptsize[-0.0294,0.0575]} & {\scriptsize[0.0024,0.1057]} & {\scriptsize[0.3614,0.5005]} & {\scriptsize[-0.1825,-0.0119]} & {\scriptsize[0.2314,0.3469]}\\
\addlinespace[4pt]
Gemini3.8 & \(\mathbf{+0.077}^{*}\) & \(\mathbf{+0.139}^{*}\) & \(\mathbf{+0.310}^{*}\) & \(-0.059\) & \(\mathbf{+0.151}^{*}\)\\
 & {\scriptsize[0.0405,0.1128]} & {\scriptsize[0.0966,0.1822]} & {\scriptsize[0.2491,0.3679]} & {\scriptsize[-0.1316,0.0111]} & {\scriptsize[0.0975,0.2018]}\\
\addlinespace[4pt]
GLM5.3 & \(-0.002\) & \(+0.030\) & \(\mathbf{+0.323}^{*}\) & \(-0.055\) & \(\mathbf{+0.305}^{*}\)\\
 & {\scriptsize[-0.0420,0.0403]} & {\scriptsize[-0.0182,0.0760]} & {\scriptsize[0.2513,0.3967]} & {\scriptsize[-0.1460,0.0323]} & {\scriptsize[0.2405,0.3688]}\\
\addlinespace[4pt]
Sonnet5 & \(-0.015\) & \(+0.020\) & \(\mathbf{+0.205}^{*}\) & \(+0.010\) & \(\mathbf{+0.149}^{*}\)\\
 & {\scriptsize[-0.0590,0.0284]} & {\scriptsize[-0.0350,0.0732]} & {\scriptsize[0.1326,0.2723]} & {\scriptsize[-0.0610,0.0848]} & {\scriptsize[0.0902,0.2087]}\\
\addlinespace[4pt]
\textit{Mean of five} & \(+0.015\) & \(\mathbf{+0.056}^{*}\) & \(\mathbf{+0.292}^{*}\) & \(-0.042\) & \(\mathbf{+0.207}^{*}\)\\
 & {\scriptsize[-0.0189,0.0508]} & {\scriptsize[0.0141,0.0984]} & {\scriptsize[0.2496,0.3337]} & {\scriptsize[-0.0871,0.0033]} & {\scriptsize[0.1728,0.2388]}\\
\addlinespace[4pt]
\bottomrule
\end{tabularx}
\end{table}

\section{Batching a product's own sessions}
\label{app:batchedc}

\arm{Session} uses 80 calls per product. The obvious way to save calls is to put all 80 of its sampled sessions
into one call and ask for 80 probabilities, which keeps the evidence and the requested judgments
and uses one call instead of 80. We ran that variant on every model and dataset, on the same
sampled sessions and under the same system message as \arm{Session} (Appendix~\ref{app:prompts}), and it ranks worse
(Table~\ref{tab:batchedc}): by \(0.144\;[0.1050,0.1795]\) on Coveo cart, \(0.161\;[0.1246,0.1965]\)
on Coveo purchase and \(0.104\;[0.0697,0.1375]\) on YooChoose, with the paired interval above zero
in 14 of the 15 settings, with GPT on YooChoose the one null.

The pattern is consistent with the shared context noise of Section~\ref{sec:mechanism}. When a product's 80
judgments share a call, any shift specific to that call is common to all of them and survives the
average, whereas across 80 separate calls such shifts average out. On the logit scale, \(\log\!\big(p/(1-p)\big)\) with probabilities clipped to \([10^{-3},1-10^{-3}]\), the intraclass correlation of a
product's per-session predictions is \(0.011\) to \(0.021\) across the five models on Coveo when the
sessions are judged separately, and \(0.035\) to \(0.114\) when they share a call, every increase
distinguishable from zero; \(11\) to \(30\%\) of the batched arm's departure from the per-session one
is a single shift common to the whole call, and that shift is unrelated to the product's observed
rate, at rank correlations between \(-0.093\) and \(0.008\). On YooChoose the per-session intraclass correlation is \(0.029\) to \(0.086\) when the sessions are judged separately, and sharing a call raises it by \(0.011\)
to \(0.035\) for three models and not at all for Sonnet and GLM, so the shared shift does not account
for the whole loss there. The transform matters only for the scale of these numbers: on raw probabilities the same quantities are \(0.012\) to \(0.014\) per-session against \(0.022\) to \(0.064\) in a shared call on Coveo cart, positive for every model. The session-level predictions of the two arms still agree closely, at correlations of \(0.73\) to \(0.93\), so batching lowers product-level accuracy while leaving the session-level predictions largely unchanged. With one call per product, we cannot tell whether this shared shift is random, so that a rerun of the same call would change it, or a systematic response to the particular set of sessions in the call.

\begin{table}[htbp]
\centering\small
\setlength{\tabcolsep}{4pt}
\caption{\arm{Session} (one session per call) against \arm{Same-product batch} (all 80 in one call) on the same sessions: rank correlations and their paired difference. A star marks an interval excluding zero.}
\label{tab:batchedc}
\begin{tabular}{@{}l ccc ccc ccc@{}}
\toprule
& \multicolumn{3}{c}{Coveo cart} & \multicolumn{3}{c}{Coveo purchase} & \multicolumn{3}{c}{YooChoose purchase}\\
\cmidrule(lr){2-4}\cmidrule(lr){5-7}\cmidrule(lr){8-10}
& one & all 80 & diff & one & all 80 & diff & one & all 80 & diff\\
\midrule
GPT-5.6 & 0.313 & 0.198 & +0.114$^{*}$ & 0.328 & 0.193 & +0.135$^{*}$ & 0.417 & 0.400 & +0.017\\
Sonnet5 & 0.278 & 0.106 & +0.172$^{*}$ & 0.301 & 0.101 & +0.200$^{*}$ & 0.278 & 0.144 & +0.134$^{*}$\\
Gemini3.8 & 0.281 & 0.134 & +0.147$^{*}$ & 0.311 & 0.148 & +0.163$^{*}$ & 0.429 & 0.372 & +0.057$^{*}$\\
GLM5.3 & 0.258 & 0.095 & +0.163$^{*}$ & 0.287 & 0.131 & +0.156$^{*}$ & 0.490 & 0.362 & +0.128$^{*}$\\
Qwen3.8 & 0.246 & 0.122 & +0.124$^{*}$ & 0.294 & 0.142 & +0.152$^{*}$ & 0.502 & 0.319 & +0.182$^{*}$\\
\addlinespace[3pt]
\textit{Mean of five} & --- & --- & +0.144$^{*}$ & --- & --- & +0.161$^{*}$ & --- & --- & +0.104$^{*}$\\
\bottomrule
\end{tabular}
\end{table}

\section{Standardization and averaging in the comparative summaries}
\label{app:normavg}

\arm{Batch summary} differs from \arm{Summary} in three ways at once: it compares 40 products in a
call, it standardizes the outputs within that call, and it averages five groupings. Removing the
last two in turn separates them (Table~\ref{tab:normavg}). A single comparative call with no
standardization already improves on \arm{Summary} in all 15 settings, by \(0.149\), \(0.179\) and
\(0.127\) on the three outcomes, so the comparison itself accounts for more than half of the gain. Standardizing that
single call adds \(0.086\), \(0.087\) and \(0.061\); averaging five unstandardized groupings instead
adds \(0.073\), \(0.082\) and \(0.070\); doing both adds only a little more than either. The overlap
is consistent with both devices removing the same call-level variation, one by
correcting each call and the other by averaging across calls, although standardization also depends on
real differences among the products sharing a call. Consistent with that, two groupings of
the standardized arm agree at rank correlations of \(0.72\) to \(0.89\) on Coveo, against \(0.28\) to
\(0.62\) unstandardized, which leaves averaging little to remove. Giving \arm{Summary} five judgments per product without comparison, by averaging five identical calls, adds \(0.012\), \(0.013\) and \(0.063\) (last row of Table~\ref{tab:normavg}).

A single standardized grouping of \arm{Batch summary} requests one judgment per product, as \arm{Summary} does. It is ahead of \arm{Session}, which averages 80 judgments per product, by \(0.040\;[-0.0071,0.0908]\) on Coveo cart, \(0.055\;[0.0090,0.0995]\) on Coveo purchase and \(0.056\;[0.0219,0.0923]\) on YooChoose, with model-level intervals above zero in 2, 3 and 3 settings and below zero in one, GLM on YooChoose.

\begin{table}[htbp]
\centering\small
\caption{Gain of \arm{Batch summary} over \arm{Summary} with and without within-call standardization, at one and five groupings: mean of five models, with the number of settings whose interval lies above zero. The fourth row is the reported arm; the last row averages five identical \arm{Summary} calls, without comparison.}
\label{tab:normavg}
\begin{tabular}{@{}lccc@{}}
\toprule
Variant & Coveo cart & Coveo purchase & YooChoose purchase\\
\midrule
\arm{Batch summary}, one grouping, raw & +0.149 (5/5) & +0.179 (5/5) & +0.127 (5/5)\\
\arm{Batch summary}, one grouping, standardized & +0.235 (5/5) & +0.266 (5/5) & +0.188 (5/5)\\
\arm{Batch summary}, five groupings, raw & +0.222 (5/5) & +0.261 (5/5) & +0.197 (5/5)\\
\arm{Batch summary}, five groupings, standardized & +0.260 (5/5) & +0.297 (5/5) & +0.250 (5/5)\\
\midrule
\arm{Summary}, five identical calls averaged & +0.012 (1/5) & +0.013 (1/5) & +0.063 (5/5)\\
\bottomrule
\end{tabular}
\end{table}

\section{Comparative gains across representations}
\label{app:did}

Comparison helps summaries more than sessions in 14 of the 15 settings (Section~\ref{sec:comparative}). In Table~\ref{tab:did}, we estimate that difference directly rather than inferring it from the relative size of two gains. For each model and outcome we form
\[
 \bigl(\widehat\rho_B-\widehat\rho_A\bigr)-\bigl(\widehat\rho_D-\widehat\rho_C\bigr)
\]
within each bootstrap replicate and report the percentile interval of that quantity. Because both gains are computed on the same replicate and the same common support, this is a paired estimate rather than a comparison of two separate intervals.

Fourteen of the 15 intervals are above zero and one contains zero, GPT on YooChoose; none is below. This estimates the difference between two complete procedures, each of which changes comparison context and standardization together, and for summary batching also repeats the product judgment. It does not isolate the contribution of any single prompting element.

\begin{table}[htbp]
\centering\small
\caption{Paired difference of differences \((\rho_B-\rho_A)-(\rho_D-\rho_C)\); positive values mean that comparison helps summaries more than sessions. A star marks an interval excluding zero.}
\label{tab:did}
\begin{tabular}{@{}lcc@{}}
\toprule
Model & Estimate & 95\% interval\\
\midrule
\multicolumn{3}{@{}l}{\emph{Coveo cart}}\\
GPT-5.6 & 0.281$^{*}$ & [0.2117, 0.3475] \\
Sonnet5 & 0.350$^{*}$ & [0.2669, 0.4321] \\
Gemini3.8 & 0.201$^{*}$ & [0.1311, 0.2670] \\
GLM5.3 & 0.341$^{*}$ & [0.2561, 0.4244] \\
Qwen3.8 & 0.349$^{*}$ & [0.2697, 0.4246] \\
\addlinespace[3pt]
\multicolumn{3}{@{}l}{\emph{Coveo purchase}}\\
GPT-5.6 & 0.317$^{*}$ & [0.2454, 0.3825] \\
Sonnet5 & 0.369$^{*}$ & [0.2870, 0.4472] \\
Gemini3.8 & 0.228$^{*}$ & [0.1583, 0.2933] \\
GLM5.3 & 0.287$^{*}$ & [0.2075, 0.3695] \\
Qwen3.8 & 0.485$^{*}$ & [0.4006, 0.5649] \\
\addlinespace[3pt]
\multicolumn{3}{@{}l}{\emph{YooChoose purchase}}\\
GPT-5.6 & 0.040 & [-0.0089, 0.0864] \\
Sonnet5 & 0.106$^{*}$ & [0.0221, 0.1916] \\
Gemini3.8 & 0.218$^{*}$ & [0.1603, 0.2803] \\
GLM5.3 & 0.271$^{*}$ & [0.1912, 0.3455] \\
Qwen3.8 & 0.304$^{*}$ & [0.2373, 0.3651] \\
\bottomrule
\end{tabular}
\end{table}

\section{Doubling the session sample}
\label{app:depth}

\arm{Batch session} samples 80 sessions per product and requests one judgment on each, while
\arm{Batch summary} is computed from the product's full history. Section~\ref{sec:comparative}
attributes part of the difference between them to that gap in the amount of evidence each method
uses. We test the attribution directly with \arm{Batch session-d}, which repeats
\arm{Batch session} on 160 distinct sampled sessions per product instead of 80. Every other
feature is retained: one session from each of 40 products per prompt, within-prompt
standardization, and averaging by product.

The deeper arm doubles the inference cost, so we ran it for one model per dataset, choosing, on each dataset, the
model for which a shortfall is most informative. On Coveo that is GPT-5.6, which has the highest
\arm{Batch session} rank correlation of the five models on both outcomes, so a shortfall there
cannot be attributed to a model that ranks sessions poorly. On YooChoose most models are close to
\arm{Batch summary}, so a reduction would not be measurable; Gemini3.8 Flash
is the exception, with the one sizeable gap. The Coveo comparison is restricted to the 743 products with at least
160 recorded sessions so that every arm in Table~\ref{tab:depth-arms} is scored on one set of
products; \arm{Batch summary} is recomputed on the same restriction and is therefore not
identical to the value in Figure~\ref{fig:models}. All 1,000 YooChoose products qualify. Calls lost in the Coveo run leave every product with at least 147 of its 160 judgments; the table uses all of them, while the extended curve in Figure~\ref{fig:batchrow} stops at 112 rounds, the deepest grid point every product supports.

Doubling the sample raises the session method by \(0.084\;[0.0271,0.1418]\) on Coveo cart,
\(0.094\;[0.0392,0.1502]\) on Coveo purchase, and \(0.044\;[0.0216,0.0663]\) on YooChoose, every
interval above zero. The number of sessions the method draws is therefore part of what separates
it from the summary methods. It is not the whole of it: the \arm{Batch summary} advantage remains
above zero at doubled depth in all three comparisons. Doubling the sample therefore does not
close the gap, which is what we can show; it does not establish that no larger sample would, and the
deeper arm is still improving at 160 sessions.

\begin{table}[htbp]
\centering\small
\caption{Doubled session sample: GPT-5.6 on the 743 Coveo products with at least 160 sessions, and Gemini3.8 Flash on all 1,000 YooChoose products. Each arm's rank correlation and paired difference from \arm{Batch summary} on the same products; a star marks an interval excluding zero.}
\label{tab:depth-arms}
\begin{tabular}{@{}llcccc@{}}
\toprule
& & \multicolumn{2}{c}{Coveo cart} & \multicolumn{2}{c}{Coveo purchase}\\
\cmidrule(lr){3-4}\cmidrule(lr){5-6}
Arm & Sessions / judgments & $\rho$ & B$-$arm & $\rho$ & B$-$arm\\
\midrule
\arm{Batch summary} & full history / 5 & 0.510 & --- & 0.521 & ---\\
\arm{Batch session} & 80 / 80 & 0.324 & 0.186$^{*}$ & 0.330 & 0.192$^{*}$\\
\arm{Batch session-d} & 160 / 160 & 0.408 & 0.103$^{*}$ & 0.424 & 0.098$^{*}$\\
\addlinespace[4pt]
\midrule
& & \multicolumn{2}{c}{YooChoose purchase} & & \\
\cmidrule(lr){3-4}
\arm{Batch summary} & full history / 5 & 0.601 & --- & & \\
\arm{Batch session} & 80 / 80 & 0.461 & 0.139$^{*}$ & & \\
\arm{Batch session-d} & 160 / 160 & 0.505 & 0.096$^{*}$ & & \\
\bottomrule
\end{tabular}
\end{table}

The deeper arm also costs twice what \arm{Batch session} costs, at 4,000 planned calls and 160,000
requested judgments per model and dataset, against 125 calls and 5,000 judgments for
\arm{Batch summary}. The additional evidence that closes part of the gap requires additional inference,
whereas the summaries incorporate more sessions at no additional inference cost (Section~\ref{sec:cost}).

\section{Held-out supervised benchmarks}
\label{app:supervised}
The benchmarks in Section~\ref{sec:results} are fit on products no LLM ever saw and scored on the same 1,000 evaluation products. Here we describe the training pool and the fitting, and give the comparisons model by model.

\noindent{\bf Training pool.}
The benchmarks are trained on products that no LLM ever saw, so the 1,000 evaluation products and their labels play no part in any fit. On Coveo we take every SKU in the feature file with at least 100 recorded sessions -- the eligibility bar the evaluation pool itself used -- and remove every product ever sent to a model in this project, including the discovery and evaluation pools. That leaves 8,410 training products. On YooChoose the evaluation pool required at least 300 sessions in the evaluation window and a price from the earlier window, and only 732 such products sit outside the YooChoose pools sent to a model, so we lower the bar to 100 sessions and train on 2,530 products; the 732-product fit at the pool's own bar is reported below. Features for the YooChoose training products are built by the same trace pass and the same block function as the evaluation pool, on each product's full evaluation-window history.

\noindent{\bf Fitting.}
Ridge selects its penalty from 30 logarithmically spaced values between $10^{-3}$ and $10^4$ using \texttt{RidgeCV}. The random forest uses 400 trees, a minimum leaf size of five, and seed zero. Gradient boosting is \texttt{HistGradientBoostingRegressor} at its defaults with seed zero. Missing features are replaced by zero, and columns are centered and scaled using the training rows only. The predictors use the full feature block, 20 columns for Coveo and 19 for YooChoose, which includes the two traffic counts the summary prompts withhold. Intervals resample the saved predictions on the shared product bootstrap rather than refitting inside each replicate, so every benchmark interval is paired with the LLM intervals it is compared against.

\noindent{\bf Paired comparisons with the LLM methods.}
The random forest and the gradient-boosted tree each lead every model's \arm{Batch summary} in all 15 model--outcome settings, with the paired interval above zero in each. The narrowest gaps are Qwen on Coveo purchase, \(0.064\;[0.0252,0.1033]\), and Gemini on YooChoose, \(0.107\;[0.0728,0.1422]\); the widest is Sonnet on Coveo cart, \(0.432\). Ridge leads in 12 of the 15, the exceptions being GPT, Gemini and Qwen on YooChoose, where the intervals contain zero.

\begin{table}[htbp]
\centering\small
\setlength{\tabcolsep}{2.5pt}
\caption{Supervised benchmarks trained on products no LLM saw (8,410 Coveo, 2,530 YooChoose) and scored on the 1,000 evaluation products. Features are the full product block, including the traffic counts withheld from the summary prompts. The single-feature rule ranks products by the negative average price of the most-considered competing product, fixed on the discovery pool. The last two rows repeat \arm{Batch summary}.}
\label{tab:holdout}
\begin{tabularx}{\linewidth}{@{}l*{3}{>{\centering\arraybackslash}X}@{}}
\toprule
Method & Coveo cart & Coveo purchase & YooChoose purchase\\
\midrule
Ridge & \(0.629\) & \(0.529\) & \(0.612\)\\
 & {\scriptsize[0.5864,0.6675]} & {\scriptsize[0.4795,0.5752]} & {\scriptsize[0.5683,0.6503]}\\
\addlinespace[4pt]
Random forest & \(0.660\) & \(0.560\) & \(0.708\)\\
 & {\scriptsize[0.6187,0.6963]} & {\scriptsize[0.5133,0.6042]} & {\scriptsize[0.6697,0.7394]}\\
\addlinespace[4pt]
Gradient boosting & \(0.659\) & \(0.560\) & \(0.702\)\\
 & {\scriptsize[0.6183,0.6988]} & {\scriptsize[0.5123,0.6031]} & {\scriptsize[0.6648,0.7342]}\\
\addlinespace[4pt]
\midrule
Single-feature rule & \(0.519\) & \(0.371\) & \(0.360\)\\
 & {\scriptsize[0.4688,0.5640]} & {\scriptsize[0.3124,0.4230]} & {\scriptsize[0.3052,0.4129]}\\
\addlinespace[4pt]
\midrule
Best \arm{Batch summary} & \(0.438\) & \(0.496\) & \(0.601\)\\
Mean \arm{Batch summary} & \(0.341\) & \(0.390\) & \(0.541\)\\
\bottomrule
\end{tabularx}
\end{table}

\noindent{\bf Where zero-shot comes closest.}
The two Coveo outcomes share products, features and prompts and differ only in the label, which makes them a clean comparison. Cart is the easier target for the supervised references, with the forest at \(0.660\) against \(0.560\) on purchase, yet the best \arm{Batch summary} reaches 66\% of the forest on cart and 89\% on purchase; the LLM methods do not share that ordering, with the best \arm{Batch summary} at \(0.438\) on cart and \(0.496\) on purchase. Purchase is the rarer and noisier label: a mean rate of \(0.0073\) against \(0.0400\) for cart, no purchase at all for 30\% of products, and about 35\% of the variance across products attributable to binomial sampling error, against 9\% for cart.

\noindent{\bf A single-feature reference.}
To measure what the LLM methods add over the strongest single behavioral signal, we rank products by one summary column, chosen without evaluation labels. On the Coveo discovery pool we take the counts-free column with the largest mean absolute Spearman correlation across the two outcomes and fix its sign there; this is the average price of the most-considered competing product, entered negatively (\(0.402\), narrowly ahead of its median and of the other competitor-price averages). Applied unchanged, it reaches \(0.519\;[0.4688,0.5640]\) on Coveo cart, \(0.371\;[0.3124,0.4230]\) on Coveo purchase and \(0.360\;[0.3052,0.4129]\) on YooChoose (Table~\ref{tab:holdout}). On Coveo cart it leads every model's \arm{Batch summary}, by \(0.081\) to \(0.291\), with every paired interval excluding zero. On Coveo purchase the model-mean difference is \(0.019\) and indistinguishable from zero: GPT and Qwen lead the rule, Sonnet trails it, and the rest cannot be separated. On YooChoose \arm{Batch summary} leads the rule by \(0.181\) on average, above zero for four models and indistinguishable for Sonnet. The LLM methods thus fare best on purchase, which neither this rule nor a small labeled sample captures.

\noindent{\bf The matched YooChoose pool.}
Restricting YooChoose training to the 732 products that clear the evaluation pool's own 300-session bar changes little: the forest reaches \(0.712\) against \(0.708\) on the larger pool, gradient boosting \(0.680\) against \(0.702\), and ridge \(0.600\) against \(0.612\). Training on products that look exactly like the test products, rather than on more of them, is not what produces the gap.

\noindent{\bf Training-size curve.}
In Table~\ref{tab:holdout-curve}, we refit the forest on random subsets of the training pool and score every fit on the same 1,000 evaluation products, so the curve and the LLM levels are on one scale. Each value averages six draws. The forest is still improving at the full pool in all three outcomes, so the label counts quoted in Section~\ref{sec:results} hold at the tested scale and are not a supervised limit. Each evaluation pool contains 1,000 products, and the millions of sessions in each dataset are inputs to those product rows rather than additional labeled observations, so no benchmark at a larger labeled sample is available here and we claim none.

\begin{table}[htbp]
\centering\small
\setlength{\tabcolsep}{2.5pt}
\caption{Random forest by number of labeled training products drawn from the held-out pool, scored on the 1,000 evaluation products; mean and standard deviation over six draws. The full-pool row is a single fit and can differ from Table~\ref{tab:holdout} in the third decimal. The last row repeats the best \arm{Batch summary}, which uses no labels.}
\label{tab:holdout-curve}
\begin{tabularx}{\linewidth}{@{}l*{3}{>{\centering\arraybackslash}X}@{}}
\toprule
Training products & Coveo cart & Coveo purchase & YooChoose purchase\\
\midrule
15 & \(0.348\) {\scriptsize(0.156)} & \(0.279\) {\scriptsize(0.170)} & ---\\
25 & \(0.461\) {\scriptsize(0.121)} & \(0.276\) {\scriptsize(0.143)} & \(0.400\) {\scriptsize(0.107)}\\
50 & \(0.554\) {\scriptsize(0.017)} & \(0.389\) {\scriptsize(0.122)} & \(0.474\) {\scriptsize(0.026)}\\
100 & \(0.565\) {\scriptsize(0.018)} & \(0.457\) {\scriptsize(0.038)} & \(0.535\) {\scriptsize(0.030)}\\
200 & \(0.600\) {\scriptsize(0.015)} & \(0.464\) {\scriptsize(0.030)} & \(0.628\) {\scriptsize(0.015)}\\
400 & \(0.620\) {\scriptsize(0.011)} & \(0.506\) {\scriptsize(0.026)} & \(0.650\) {\scriptsize(0.014)}\\
800 & \(0.634\) {\scriptsize(0.007)} & \(0.527\) {\scriptsize(0.009)} & \(0.679\) {\scriptsize(0.004)}\\
1,600 & \(0.648\) {\scriptsize(0.003)} & \(0.542\) {\scriptsize(0.007)} & \(0.694\) {\scriptsize(0.001)}\\
3,200 & \(0.654\) {\scriptsize(0.004)} & \(0.552\) {\scriptsize(0.009)} & ---\\
Full pool & \(0.660\) {\scriptsize(8,410)} & \(0.562\) {\scriptsize(8,410)} & \(0.709\) {\scriptsize(2,530)}\\
\midrule
Best \arm{Batch summary} & \(0.438\) & \(0.496\) & \(0.601\)\\
\bottomrule
\end{tabularx}
\end{table}

\section{Inference usage}
\label{app:usage}
Tables~\ref{tab:full-usage-total-coveo} and~\ref{tab:full-usage-total-yoochoose} contain the totals across the six arms the logs identify; \arm{Session bag} was run before the reported summary runs and its usage is not separable, so it is excluded. Tables~\ref{tab:full-usage-blocks-coveo} and~\ref{tab:full-usage-blocks-yoochoose} give the recorded blocks. Charges use list prices at execution time; output tokens include reasoning tokens, which are billed as output. The latency entries describe the observed run under its concurrency and retry settings. They are not intrinsic model-speed estimates or sequential end-to-end pipeline runtimes.

The two full-history summary methods share a usage record for A+B, and the two matched summary methods share A80+B80. We preserve these blocks rather than allocate their costs to methods without supporting logs. Calls include retries and smoke tests where present. Offline feature computation, session sampling, and evaluation are not included.

\noindent{\bf Basis of the main-text cost comparison.}
At the same 80,000 judgments, \arm{Session} costs \$103.88 against \$38.96 for \arm{Batch session} for GPT on Coveo, \$36.99 against \$17.97 for Gemini, \$38.02 against \$21.96 for GLM, \$77.54 against \$47.80 for Sonnet, and \$234.93 against \$74.47 for Qwen. The difference is instruction overhead: for GPT on Coveo, \arm{Session} uses 17.3M input tokens against 8.8M. On exactly the same 80 sessions, the two matched-evidence summary methods together used 1.35M input tokens and 0.20M output tokens for GPT on Coveo, against 8.79M and 1.78M for \arm{Batch session}, and cost \$5.09 against \$38.96. Because that block is pooled, the combined figure bounds either summary method from above. Reasoning tokens account for essentially none of the output for Sonnet (at most 0.1\%) and 70--90\% for Qwen everywhere, which produces the fourfold spread across models under \arm{Batch session} and the sixfold spread under \arm{Session}. Across the six arms the logs identify, the recorded charges are \$751.27 on Coveo and \$761.51 on YooChoose.

\begin{table}[htbp]
\centering\footnotesize
\setlength{\tabcolsep}{2.5pt}
\caption{Coveo: total inference usage for A, B, A80, B80, C and D, including retries. Tokens are in millions, and output includes reasoning; prices are list prices at run time. Seconds per call are specific to this run; offline data preparation is excluded.}
\label{tab:full-usage-total-coveo}
\begin{tabularx}{\linewidth}{@{}l*{8}{>{\raggedleft\arraybackslash}X}@{}}
\toprule
Model & Calls & Input M & Output M & Reason M & Reason \% & Out/call & s/call & USD\\
\midrule
GPT-5.6 & 84,266 & 28.774 & 7.955 & 3.875 & 48.7 & 94 & 3.7 & 153.01\\
Qwen3.8 & 84,271 & 32.857 & 45.164 & 38.792 & 85.9 & 536 & 8.2 & 336.70\\
Gemini3.8 & 84,303 & 33.111 & 9.253 & 3.695 & 39.9 & 110 & 2.0 & 59.53\\
GLM5.3 & 84,264 & 29.644 & 5.449 & 1.136 & 20.9 & 65 & 1.4 & 65.48\\
Sonnet5 & 84,259 & 39.400 & 5.775 & 0.001 & 0.0 & 69 & 2.3 & 136.55\\
\midrule
\multicolumn{8}{@{}l}{Total cost} & 751.27\\
\bottomrule
\end{tabularx}
\end{table}

\begin{table}[htbp]
\centering\footnotesize
\setlength{\tabcolsep}{2.5pt}
\caption{YooChoose: total inference usage for A, B, A80, B80, C and D, including retries. Tokens are in millions, and output includes reasoning; prices are list prices at run time. Seconds per call are specific to this run; offline data preparation is excluded.}
\label{tab:full-usage-total-yoochoose}
\begin{tabularx}{\linewidth}{@{}l*{8}{>{\raggedleft\arraybackslash}X}@{}}
\toprule
Model & Calls & Input M & Output M & Reason M & Reason \% & Out/call & s/call & USD\\
\midrule
GPT-5.6 & 84,259 & 27.618 & 8.294 & 4.191 & 50.5 & 98 & 3.8 & 154.76\\
Qwen3.8 & 84,273 & 33.069 & 47.559 & 40.952 & 86.1 & 564 & 8.7 & 351.49\\
Gemini3.8 & 84,332 & 32.724 & 8.411 & 2.857 & 34.0 & 100 & 2.0 & 56.08\\
GLM5.3 & 84,319 & 29.294 & 5.660 & 1.436 & 25.4 & 67 & 1.5 & 65.91\\
Sonnet5 & 84,258 & 37.717 & 5.783 & 0.006 & 0.1 & 69 & 2.1 & 133.26\\
\midrule
\multicolumn{8}{@{}l}{Total cost} & 761.51\\
\bottomrule
\end{tabularx}
\end{table}

\begin{table}[htbp]
\centering\footnotesize
\setlength{\tabcolsep}{2.5pt}
\caption{Coveo: inference usage by recorded arm block; the pooled A+B and A80+B80 logs cannot be split by arm. Tokens are in millions, output includes reasoning, and counts include retries.}
\label{tab:full-usage-blocks-coveo}
\begin{tabularx}{\linewidth}{@{}l*{8}{>{\raggedleft\arraybackslash}X}@{}}
\toprule
Block & Calls & Input M & Output M & Reason M & Reason \% & Out/call & s/call & USD\\
\midrule
\multicolumn{9}{@{}l}{\textit{GPT-5.6}}\\
A+B & 1,133 & 1.360 & 0.197 & 0.077 & 39.1 & 174 & 3.9 & 5.09\\
A80+B80 & 1,133 & 1.345 & 0.200 & 0.076 & 38.1 & 176 & 3.8 & 5.09\\
C & 80,000 & 17.280 & 5.776 & 3.622 & 62.7 & 72 & 3.5 & 103.88\\
D & 2,000 & 8.788 & 1.782 & 0.099 & 5.6 & 891 & 9.9 & 38.96\\
\addlinespace[5pt]
\multicolumn{9}{@{}l}{\textit{Qwen3.8}}\\
A+B & 1,133 & 1.531 & 1.952 & 1.695 & 86.8 & 1,723 & 24.8 & 14.78\\
A80+B80 & 1,131 & 1.508 & 1.585 & 1.358 & 85.7 & 1,401 & 19.4 & 12.52\\
C & 80,002 & 20.382 & 32.361 & 29.248 & 90.4 & 404 & 6.5 & 234.93\\
D & 2,005 & 9.435 & 9.267 & 6.491 & 70.1 & 4,622 & 59.6 & 74.47\\
\addlinespace[5pt]
\multicolumn{9}{@{}l}{\textit{Gemini3.8}}\\
A+B & 1,131 & 1.649 & 0.289 & 0.099 & 34.4 & 255 & 2.9 & 2.32\\
A80+B80 & 1,130 & 1.648 & 0.272 & 0.088 & 32.2 & 240 & 3.0 & 2.26\\
C & 80,037 & 19.047 & 6.054 & 3.508 & 57.9 & 76 & 1.9 & 36.99\\
D & 2,005 & 10.766 & 2.638 & 0.000 & 0.0 & 1,316 & 7.9 & 17.97\\
\addlinespace[5pt]
\multicolumn{9}{@{}l}{\textit{GLM5.3}}\\
A+B & 1,130 & 1.368 & 0.195 & 0.049 & 25.3 & 173 & 3.7 & 2.77\\
A80+B80 & 1,129 & 1.357 & 0.188 & 0.045 & 23.7 & 166 & 4.4 & 2.73\\
C & 80,001 & 17.988 & 2.918 & 0.714 & 24.5 & 36 & 1.1 & 38.02\\
D & 2,004 & 8.931 & 2.148 & 0.328 & 15.3 & 1,072 & 11.0 & 21.96\\
\addlinespace[5pt]
\multicolumn{9}{@{}l}{\textit{Sonnet5}}\\
A+B & 1,129 & 1.899 & 0.182 & 0.000 & 0.0 & 162 & 3.0 & 5.62\\
A80+B80 & 1,129 & 1.884 & 0.183 & 0.000 & 0.1 & 162 & 2.8 & 5.59\\
C & 80,000 & 23.757 & 3.002 & 0.000 & 0.0 & 38 & 2.1 & 77.54\\
D & 2,001 & 11.859 & 2.408 & 0.001 & 0.0 & 1,203 & 9.8 & 47.80\\
\addlinespace[5pt]
\bottomrule
\end{tabularx}
\end{table}

\begin{table}[htbp]
\centering\footnotesize
\setlength{\tabcolsep}{2.5pt}
\caption{YooChoose: inference usage by recorded arm block; the pooled A+B and A80+B80 logs cannot be split by arm. Tokens are in millions, output includes reasoning, and counts include retries.}
\label{tab:full-usage-blocks-yoochoose}
\begin{tabularx}{\linewidth}{@{}l*{8}{>{\raggedleft\arraybackslash}X}@{}}
\toprule
Block & Calls & Input M & Output M & Reason M & Reason \% & Out/call & s/call & USD\\
\midrule
\multicolumn{9}{@{}l}{\textit{GPT-5.6}}\\
A+B & 1,129 & 1.384 & 0.198 & 0.079 & 40.1 & 175 & 3.9 & 5.14\\
A80+B80 & 1,129 & 1.381 & 0.204 & 0.082 & 40.3 & 180 & 3.9 & 5.21\\
C & 80,001 & 16.672 & 6.103 & 3.927 & 64.3 & 76 & 3.6 & 106.58\\
D & 2,000 & 8.180 & 1.789 & 0.102 & 5.7 & 895 & 11.7 & 37.83\\
\addlinespace[5pt]
\multicolumn{9}{@{}l}{\textit{Qwen3.8}}\\
A+B & 1,135 & 1.674 & 1.840 & 1.613 & 87.6 & 1,621 & 21.6 & 14.39\\
A80+B80 & 1,133 & 1.650 & 1.733 & 1.515 & 87.4 & 1,530 & 21.5 & 13.70\\
C & 80,003 & 20.356 & 34.203 & 30.860 & 90.2 & 428 & 7.2 & 245.93\\
D & 2,002 & 9.389 & 9.783 & 6.964 & 71.2 & 4,886 & 52.6 & 77.47\\
\addlinespace[5pt]
\multicolumn{9}{@{}l}{\textit{Gemini3.8}}\\
A+B & 1,131 & 1.770 & 0.317 & 0.129 & 40.6 & 280 & 3.5 & 2.52\\
A80+B80 & 1,130 & 1.765 & 0.312 & 0.131 & 42.0 & 276 & 3.1 & 2.49\\
C & 80,069 & 18.735 & 5.137 & 2.597 & 50.6 & 64 & 1.8 & 33.31\\
D & 2,002 & 10.454 & 2.645 & 0.000 & 0.0 & 1,321 & 8.0 & 17.76\\
\addlinespace[5pt]
\multicolumn{9}{@{}l}{\textit{GLM5.3}}\\
A+B & 1,131 & 1.451 & 0.188 & 0.043 & 23.0 & 166 & 4.4 & 2.86\\
A80+B80 & 1,129 & 1.429 & 0.196 & 0.053 & 26.9 & 174 & 3.8 & 2.86\\
C & 80,000 & 17.638 & 3.068 & 0.873 & 28.4 & 38 & 1.1 & 38.19\\
D & 2,059 & 8.776 & 2.207 & 0.467 & 21.2 & 1,072 & 13.3 & 22.00\\
\addlinespace[5pt]
\multicolumn{9}{@{}l}{\textit{Sonnet5}}\\
A+B & 1,129 & 1.934 & 0.183 & 0.000 & 0.0 & 162 & 2.7 & 5.70\\
A80+B80 & 1,129 & 1.931 & 0.182 & 0.000 & 0.0 & 161 & 2.7 & 5.68\\
C & 80,000 & 22.878 & 3.007 & 0.003 & 0.1 & 38 & 1.9 & 75.82\\
D & 2,000 & 10.974 & 2.412 & 0.003 & 0.1 & 1,206 & 8.5 & 46.06\\
\addlinespace[5pt]
\bottomrule
\end{tabularx}
\end{table}

\section{Informativeness of context, standardization, and session methods at equal depth}
\label{app:context}
The analyses behind Section~\ref{sec:mechanism} use only the saved inputs and predictions; no model was called again. In Table~\ref{tab:context}, we set the gain from each kind of comparative context against the informativeness of a single entry in the call.

\begin{table}[htbp]
\centering
\small
\caption{Gain from comparison against the informativeness of one entry: the rank correlation between observed rates and a cross-fitted supervised score from a single session or summary of each product (Table~\ref{tab:informativeness}), not defined for calls whose entries all belong to one product. Gain: mean paired difference over five models from the isolated counterpart.}
\label{tab:context}
\begin{tabular}{@{}lcccccc@{}}
\toprule
 & \multicolumn{3}{c}{Informativeness of one entry} & \multicolumn{3}{c}{Gain from comparison} \\
\cmidrule(lr){2-4}\cmidrule(l){5-7}
Entries in one call & Cart & Purch. & YooCh. & Cart & Purch. & YooCh. \\
\midrule
Sessions of one product & \multicolumn{3}{c}{one product per call} & \(-0.144\) & \(-0.161\) & \(-0.104\) \\
One session from each of 40 products & 0.18 & 0.06 & 0.36 & \(-0.044\) & \(-0.040\) & \(+0.062\) \\
Summaries of 80 sessions, 40 products & 0.52 & 0.41 & 0.61 & \(+0.148\) & \(+0.162\) & \(+0.207\) \\
Summaries of full histories, 40 products & 0.63 & 0.54 & 0.70 & \(+0.260\) & \(+0.297\) & \(+0.250\) \\
\bottomrule
\end{tabular}
\end{table}

\noindent{\bf Informativeness of a single entry.}
For sessions, we fit a gradient-boosted classifier to predict a session's outcome from the fields of its record as rendered in the prompts (Appendix~\ref{app:prompts}), with five folds split by product so that no product's own sessions contribute to its score, and take the predicted log-odds as the session score. We report the share of the score's variance that lies between products (the intraclass correlation over each product's 80 sessions), the rank correlation between the product's observed rate and the score of one randomly drawn session per product (averaged over 50 draws), and the same for the average over all 80 sessions. For summaries, we fit a gradient-boosted regressor to the 18 summary columns given to the models, trained on the ranks of the observed rates with five-fold cross-fitting repeated five times; one entry is the product's summary. These scores are fit with labels and measure what the inputs contain, not what the LLMs extract. The LLMs' own session ratings show the same contrast between datasets: judged separately, 1.1 to 2.1\% of their variance lies between products on Coveo and 2.9 to 8.6\% on YooChoose (Appendix~\ref{app:batchedc}).

\begin{table}[h]
\centering
\small
\caption{Informativeness of a single entry. Sessions: between-product share of the cross-fitted session score's variance, and rank correlation with the observed rate of one session's score and of the 80-session average; ``No product-level fields'' drops fields constant within a product (Coveo: price bucket; YooChoose: price and click context). Summaries: rank correlation of the cross-fitted score and of the best single column.}
\label{tab:informativeness}
\setlength{\tabcolsep}{4pt}
\begin{tabular}{@{}llccc@{}}
\toprule
 & & Coveo cart & Coveo purchase & YooChoose \\
\midrule
Session, all fields & between-product share & 0.127 & 0.098 & 0.407 \\
 & one session & 0.183 & 0.056 & 0.355 \\
 & 80-session average & 0.261 & 0.222 & 0.582 \\
Session, no product-level fields & between-product share & 0.097 & 0.125 & 0.099 \\
 & one session & 0.124 & 0.048 & 0.150 \\
Summary of 80 sessions & cross-fitted score & 0.517 & 0.407 & 0.612 \\
Summary of full history & cross-fitted score & 0.629 & 0.544 & 0.699 \\
 & best single column & 0.519 & 0.371 & 0.415 \\
\bottomrule
\end{tabular}
\end{table}

Field by field, the behavioral fields of a Coveo session vary mostly within products: 1.2\% of the variance of viewing time and 0.5 to 1.7\% of the search fields lie between products, and at most 19\% for any competitor field. On YooChoose the exact price is constant across a product's sessions and the click context nearly so (79\% between products), and these two fields account for most of the difference in Table~\ref{tab:informativeness}.

\noindent{\bf Effect of standardization.}
In Table~\ref{tab:normeffect}, we compare each comparative arm with and without within-call standardization, on the same common supports and bootstrap draws as the main results. Standardization adds \(0.086\), \(0.087\) and \(0.061\) to a single comparative summary call (Table~\ref{tab:normavg}), less when five groupings are averaged, and changes \arm{Batch session}, which averages 80 calls, by \(-0.006\), \(-0.005\) and \(0.004\) on average, with every interval containing zero; for individual models, 2 of the 15 intervals exclude zero, one in each direction, at \(-0.017\) (GLM, Coveo cart) and \(0.019\) (GPT, YooChoose).

\begin{table}[h]
\centering
\small
\caption{Within-call standardization. Mean over five models of the standardized minus the unstandardized rank correlation, and the unstandardized \arm{Batch session}$-$\arm{Session} difference, with 95\% paired-bootstrap intervals.}
\label{tab:normeffect}
\setlength{\tabcolsep}{3pt}\resizebox{\linewidth}{!}{%
\begin{tabular}{@{}lccc@{}}
\toprule
 & Coveo cart & Coveo purchase & YooChoose \\
\midrule
\arm{Batch summary} (5 groupings) & \(0.039\;[0.025,0.052]\) & \(0.036\;[0.025,0.048]\) & \(0.053\;[0.040,0.066]\) \\
\arm{Batch summary-80} & \(0.011\;[0.001,0.022]\) & \(0.017\;[0.007,0.027]\) & \(0.062\;[0.048,0.075]\) \\
\arm{Batch session} (80 rounds) & \(-0.006\;[-0.020,0.008]\) & \(-0.005\;[-0.021,0.011]\) & \(0.004\;[-0.005,0.013]\) \\
\midrule
\arm{Batch session} (unstandardized)$-$\arm{Session} & \(-0.038\;[-0.058,-0.016]\) & \(-0.035\;[-0.058,-0.011]\) & \(0.058\;[0.029,0.086]\) \\
\bottomrule
\end{tabular}}
\end{table}

\noindent{\bf Batch session against Session at equal numbers of judgments.}
In Table~\ref{tab:depthgap}, we recompute both session methods from the same number \(X\) of stored judgments per product, as in Figures~\ref{fig:depth} and~\ref{fig:batchrow}, and report their difference by model, together with the loss from batching a product's own sessions into one call (Table~\ref{tab:batchedc}). On Coveo the difference stays within 0.035 of zero at every depth for GPT, Qwen and Gemini, and widens with \(X\) for GLM and Sonnet. With 80 rounds we cannot tell whether it would close with more. The loss from batching a product's own sessions is smallest for GPT and largest for Sonnet on both Coveo outcomes.

\begin{table}[h]
\centering
\small
\caption{\arm{Batch session}$-$\arm{Session} when both use \(X\) judgments per product (averaged over random draws), and the loss from batching a product's 80 sessions into one call (\arm{Session}$-$\arm{Same-product batch}).}
\label{tab:depthgap}
\setlength{\tabcolsep}{3pt}\resizebox{\linewidth}{!}{%
\begin{tabular}{@{}lccccccccccccc@{}}
\toprule
 & \multicolumn{3}{c}{Coveo cart, \(X=\)} & \multicolumn{3}{c}{Coveo purchase, \(X=\)} & \multicolumn{3}{c}{YooChoose, \(X=\)} & & \multicolumn{3}{c}{Same-product batch loss} \\
\cmidrule(lr){2-4}\cmidrule(lr){5-7}\cmidrule(lr){8-10}\cmidrule(l){12-14}
Model & 1 & 20 & 80 & 1 & 20 & 80 & 1 & 20 & 80 & & Cart & Purch. & YooCh. \\
\midrule
GPT-5.6 & \(+.002\) & \(+.019\) & \(+.006\) & \(+.006\) & \(+.004\) & \(-.007\) & \(+.025\) & \(+.128\) & \(+.143\) & & .114 & .135 & .017 \\
Qwen3.8 & \(+.013\) & \(+.011\) & \(+.000\) & \(+.006\) & \(-.009\) & \(-.020\) & \(-.003\) & \(+.029\) & \(+.029\) & & .124 & .152 & .182 \\
Gemini3.8 & \(-.014\) & \(-.010\) & \(-.021\) & \(-.009\) & \(-.018\) & \(-.027\) & \(-.006\) & \(+.041\) & \(+.033\) & & .147 & .163 & .057 \\
GLM5.3 & \(-.013\) & \(-.044\) & \(-.083\) & \(-.008\) & \(-.024\) & \(-.051\) & \(-.046\) & \(-.032\) & \(-.003\) & & .163 & .156 & .128 \\
Sonnet5 & \(-.037\) & \(-.070\) & \(-.122\) & \(-.033\) & \(-.054\) & \(-.096\) & \(-.005\) & \(+.070\) & \(+.109\) & & .172 & .200 & .134 \\
\bottomrule
\end{tabular}}
\end{table}

\section{Same-product and cross-product context}
\label{app:mechanism}

In Section~\ref{sec:mechanism}, we interpret task variation as the most direct way for context to inform how products compare. Here we state that interpretation and the evidence we have for it, which is consistent with it but does not establish it.

\noindent{\bf Interpretation.}
Ranking products correctly requires placing their characteristics relative to one another, for example knowing whether an average viewing time of 60 seconds or a given price is high or low for this store. Entries from different products show directly what is typical across products. A product's own sessions vary mainly across its shoppers, so we expect them to inform the product's own characteristics more than its place among other products. They are not uninformative about the environment, however: Coveo session records include the prices, viewing times and discovery order of competing products, and a model's prior may relate a product's records to the store as a whole. Sessions seen together can also reveal patterns no single session shows, such as the mix of shoppers a product attracts, and better inference about a product's own characteristics can improve its rank even when the model's view of the environment is unchanged. We therefore do not claim that same-product context cannot improve rankings, only that we expect it to contribute less than context that varies across products.

\noindent{\bf Evidence.}
\arm{Session bag} places a product's 80 sampled sessions in one call and asks for a single product rate, so the model can use any pattern across the sessions. It ranks at \(0.145\), \(0.137\) and \(0.250\) on the three outcomes, against \(0.085\), \(0.083\) and \(0.293\) for \arm{Summary-80}, the isolated summary of the same sessions, and \(0.275\), \(0.304\) and \(0.423\) for \arm{Session}. The paired \arm{Session bag}$-$\arm{Summary-80} differences are \(0.059\;[0.0185,0.0991]\), \(0.054\;[0.0168,0.0943]\) and \(-0.044\;[-0.0846,-0.0019]\). This comparison of procedures is in line with the interpretation: a single judgment from one product's sessions together ranks much nearer an isolated product-level judgment than the average of separate session judgments. It does not isolate the reason, since \arm{Session bag} also changes the requested output and the computation within the call. \arm{Same-product batch} also carries context noise shared by all 80 judgments (Appendix~\ref{app:batchedc}); with one call per product we cannot tell whether that shift is random or a systematic response to the particular sessions.

\end{document}